\PassOptionsToPackage{table}{xcolor}
\documentclass{article}
\usepackage[preprint, nonatbib]{neurips_2024}
\usepackage[utf8]{inputenc} 
\usepackage[T1]{fontenc}    
\usepackage{hyperref}       
\usepackage{url}            
\usepackage{booktabs}       
\usepackage{amsfonts}       
\usepackage{nicefrac}       
\usepackage{microtype}         
\usepackage{graphicx}
\usepackage{multicol}
\usepackage{multirow}
\usepackage{comment}
\usepackage{amsmath}
\usepackage{svg}
\usepackage{pdfpages}
\usepackage{amsmath}
\usepackage{multirow}
\usepackage{array}
\usepackage{longtable}
\usepackage{cellspace}
\usepackage{rotating}
\usepackage{tabularx}
\usepackage{colortbl}
\usepackage{makecell}
\usepackage{float}
\usepackage{titlesec}
\usepackage{nicematrix}
\usepackage{enumitem}
\usepackage{fancyhdr}
\usepackage{soul}
\usepackage{eso-pic}

\usepackage[bottom]{footmisc}

\fancypagestyle{reviewcopy}{
  \fancyhf{}
  \fancyfoot[C]{\thepage}
  
}

\usepackage{hhline}
\usepackage[originalparameters]{ragged2e}
\usepackage[table]{xcolor}

\definecolor{rough}{rgb}{0.6, 0.3, 0.0}     
\definecolor{conceptual}{rgb}{0.0, 0.5, 0.5}  
\definecolor{writing}{rgb}{0.0, 0.0, 0.8}  
\definecolor{highlightc}{rgb}{1.0, 1.0, 0.8} 
\definecolor{highlightb}{rgb}{0.9, 1.0, 1.0} 
\definecolor{highlightb}{rgb}{0.90, 0.95, 1.0} 
\sethlcolor{highlightc}
\definecolor{li}{rgb}{0.8,0.85,1}

\usepackage{tikz}
\usetikzlibrary{shapes,arrows,positioning,fit}
\usepackage{enumitem}
\usepackage{hyperref}
\usepackage{appendix}
\hypersetup{
    colorlinks=true,
    linkcolor=[rgb]{0.0,0.37,0.72},    
    urlcolor=[rgb]{0.0,0.37,0.72}, 
    citecolor=[rgb]{0.0,0.37,0.72}     
}
\setlist[itemize]{leftmargin=3em, rightmargin=2em}
\setlist[enumerate]{leftmargin=3em, rightmargin=2em}

\usepackage[style=apa, backend=biber, maxcitenames=1, mincitenames=1, maxbibnames=1, minbibnames=1]{biblatex}

\DeclareNameFormat{labelname}{
  \ifnum\value{listcount}=1
    \namepartfamily
    \ifnumgreater{\value{listtotal}}{1}
      {\space\bibstring{andothers}}
      {}
  \fi
}
\AtBeginBibliography{
    
}
\let\origcite\cite 
\let\cite\parencite
\DeclareDelimFormat{nameyeardelim}{\addcomma\space}
\DeclareNameFormat{labelname}{
  \ifnum\value{listcount}=1
    \namepartfamily
    \ifnumgreater{\value{listtotal}}{1}
      {\space\bibstring{andothers}}
      {}%
  \fi}

\AtBeginDocument{
  \renewbibmacro*{name:andothers}{
    \ifboolexpr{
      test {\ifnumgreater{\value{listcount}}{1}}
      or
      test {\ifnumequal{\value{listcount}}{2}}
    }
    {\space\bibstring{andothers}}
    {}}
}

\usepackage{cleveref}
\crefname{section}{Section}{Sections}
\crefname{table}{Table}{Tables}
\crefname{figure}{Figure}{Figures}
\crefname{appendix}{Appendix}{Appendices}
\crefformat{section}{Appendix~#2#1#3}
\DeclareLabeldate{
  \field{date}
  \field{eventdate}
  \field{origdate}
  \field{urldate}
  \field{pubstate}
  \literal{nodate}
}
\renewbibmacro*{addendum+pubstate}{
  \printfield{addendum}
  \iffieldequalstr{labeldatesource}{pubstate}{}
  {\newunit\newblock\printfield{pubstate}}}
\DeclareSourcemap{
  \maps[datatype=bibtex]{
    \map{
      \step[fieldsource=entrykey,
            match=\regexp{mallah_evaluating_2025|mallah_aspect-oriented_2025|kilian_multiscale_2025},
            final]
      \step[fieldset=pubstate, fieldvalue={forthcoming}]
      \step[fieldset=date, null]
      \step[fieldset=year, null]
    }
  }
}
\DeclareFieldFormat{bibstring}{#1}
\renewbibmacro*{date}{
  \iffieldundef{pubstate}
    {\printdate}
    {\printfield{pubstate}}}

\usepackage[font=small,justification=raggedright,labelformat=simple]{caption}
\DeclareCaptionLabelFormat{itformat}{\textit{#1 #2}}
\makeatletter
\renewcommand{\subsubsection}{\@startsection{subsubsection}{3}{\z@}
  {-3.25ex\@plus -1ex \@minus -.2ex}
  {1.5ex \@plus .2ex}
  {\normalfont\normalsize\itshape}}
\makeatother

\makeatletter
\renewenvironment{abstract}
{
  \vskip 0.075in
  \centerline
  {\large\bf Abstract}
  \vspace{0.5ex}
  \begin{quote}
  \setlength{\leftmargin}{0.2in}
  \setlength{\rightmargin}{0.2in}
}
{
  \par
  \end{quote}
  \vskip 1ex
}
\makeatother

\makeatletter

\renewcommand{\@maketitle}{
  \vbox{
    \hsize\textwidth
    \linewidth\hsize
    \vskip 0in 
    \@toptitlebar
    \centering
    {\LARGE\bf \@title\par}
    \@bottomtitlebar
    \if@submission
      \begin{tabular}[t]{c}\bf\rule{\z@}{24\p@}
        Anonymous Author(s) \\
        Affiliation \\
        Address \\
        \texttt{email} \\
      \end{tabular}
    \else
      \def\And{
        \end{tabular}\hfil\linebreak[0]\hfil
        \begin{tabular}[t]{c}\bf\rule{\z@}{24\p@}\ignorespaces
      }
      \def\AND{
        \end{tabular}\hfil\linebreak[4]\hfil
        \begin{tabular}[t]{c}\bf\rule{\z@}{24\p@}\ignorespaces
      }
      \begin{tabular}[t]{c}\bf\rule{\z@}{24\p@}\@author\end{tabular}
    \fi
    \vskip 0.1in \@minus 0.05in 
  }
}

\title{Adapting Probabilistic Risk Assessment for AI}

\author{
  Anna Katariina~Wisakanto\hspace{1em} Joe Rogero\hspace{1em} Avyay M. Casheekar\hspace{1em} Richard Mallah\thanks{Correspondence: PRA@carma.org.}\\
  \\
  \\
  Center for AI Risk Management \& Alignment\\
}

\begin{document}

\maketitle
\setcounter{footnote}{0}

\thispagestyle{fancy}

\begin{abstract}
Modern general-purpose artificial intelligence (AI) systems present an urgent risk management challenge, as their rapidly evolving capabilities and potential for catastrophic harm outpace our ability to reliably assess their risks. 
Current methods often rely on selective testing and undocumented assumptions about risk priorities, frequently failing to make a serious attempt at assessing the set of pathways through which AI systems pose direct or indirect risks to society and the biosphere.
This paper introduces the probabilistic risk assessment (PRA) for AI framework, adapting established PRA techniques from high-reliability industries (e.g., nuclear power, aerospace) for the new challenges of advanced AI.
The framework guides assessors in identifying potential risks, estimating likelihood and severity bands, and explicitly documenting evidence, underlying assumptions, and analyses at appropriate granularities.
The framework's implementation tool synthesizes the results into a risk report card with aggregated risk estimates from all assessed risks. 
It introduces three methodological advances: (1) Aspect-oriented hazard analysis provides systematic hazard coverage guided by a first-principles taxonomy of AI system aspects (e.g. capabilities, domain knowledge, affordances); 
(2) Risk pathway modeling analyzes causal chains from system aspects to societal impacts using bidirectional analysis and incorporating prospective techniques; and 
(3) Uncertainty management employs scenario decomposition, reference scales, and explicit tracing protocols to structure credible projections with novelty or limited data.
Additionally, the framework harmonizes diverse assessment methods by  integrating evidence into comparable, quantified absolute risk estimates for lifecycle decisions.
We have implemented this as a workbook tool for AI developers, evaluators, and regulators, available on the \href{https://pra-for-ai.github.io/pra/}{project website}.
\end{abstract}

\begin{figure}[H]
    \centering
    
    \hspace{0.01\textwidth}   
    \includegraphics[width=0.95\textwidth]{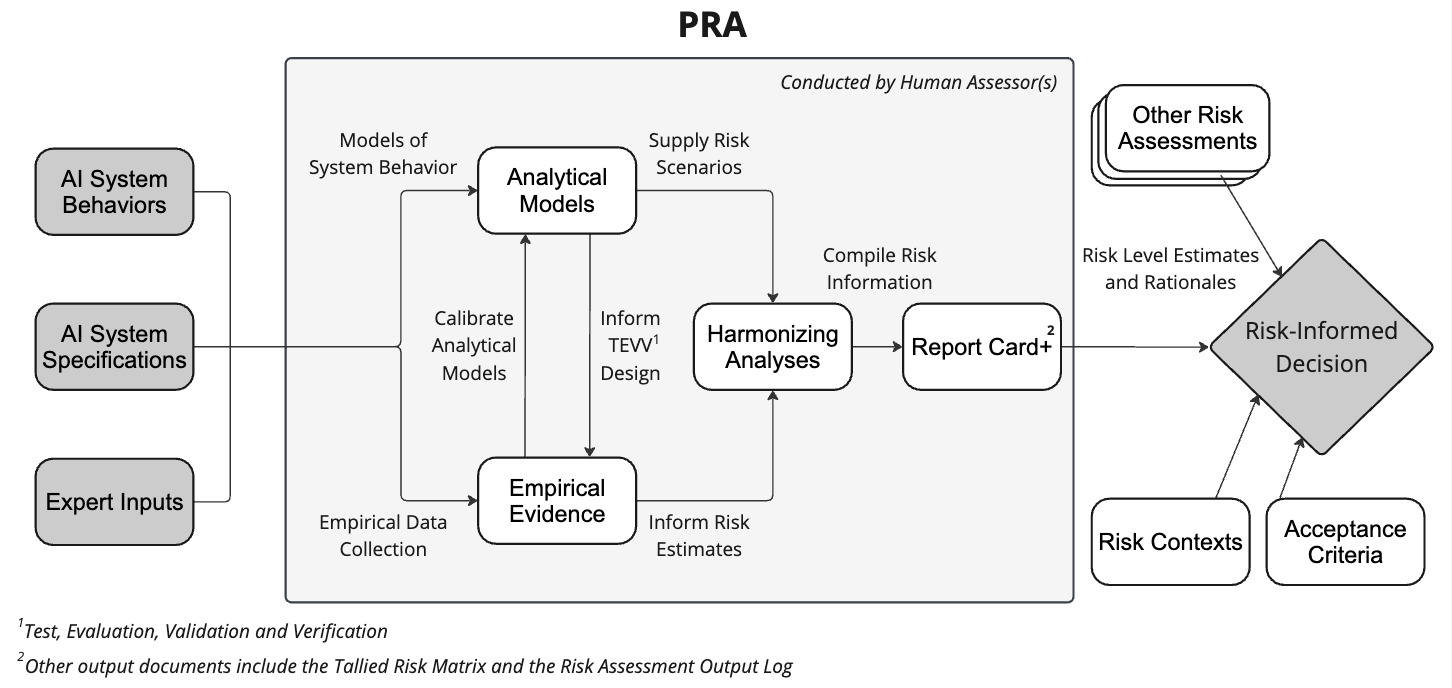}
    
    \caption{An overview of the PRA for AI framework in its operational context.}
    \label{Fig 1 Overview}
    \vspace{-1cm}
\end{figure}

\newpage
\section{Introduction}
\label{introduction}

The increasing complexity and wide-ranging capabilities of modern general-purpose AI systems present unprecedented challenges in reliably assessing their risks. These assessment challenges are further exacerbated by the difficulty in predicting emergent behaviors as AI systems evolve, testing capabilities that may manifest only in deployment contexts, and evaluating impacts that can rapidly scale and propagate through interconnected sociotechnical systems. These complex characteristics demand systematic assessment approaches suitable for reasoning under uncertainty with limited historical precedent.

AI systems integrate diverse functionalities, such as language and vision models \parencite{anthropic_introducing_2024, openai_gpt-4_2024, gemini_gemini_2024}, neural and symbolic reasoning \parencite{abramson_accurate_2024}, external tools including memory and compilers \parencite{lin_llm-based_2024, zelikman_self-taught_2024}, and agentic computer use \cite{openai_computer-using_2025, google_introducing_2024, anthropic_introducing_2024, microsoft_retrace_2024}. Such architectures increasingly enable fluid tool use, extended operations utilizing scaffolds \cite{suzgun_meta-prompting_2024}, agentic behaviors, and interactions between multiple AI systems, 
which operate within complex sociotechnical contexts.

Despite the unprecedented capabilities of current AI systems \parencite{fang_teams_2024}, the AI ecosystem has failed to implement quantified risk assessments appropriate to their potential impact \parencite{dalrymple_towards_2024}. 
This safety-capability gap creates significant risk, as AI systems could potentially cause catastrophic consequences---a concern acknowledged by leading AI developers \parencite{anthropic_core_2023, anderljung_frontier_2023, shevlane_model_2023},  
civil society organizations \cite{givens_cdt_2023}, independent oversight bodies \cite{nist_us_2025}, public security authorities \cite{dhs_dhs_2025}, international bodies \cite{coe_framework_2024}, and independent experts 
\parencite{song_singapore_2025, bengio_managing_2024, hendrycks_overview_2023, aguirre_close_2024}.
The safe operating envelope of a general-purpose AI system is far from intuitive. These systems require risk assessment that goes beyond narrowly defined accuracy or planned behavioral specifications, to address their broader range of syntheses, decision-making, and actions within the complex environments where they operate.

\textbf{Current approaches fail to adequately address risks to society.} 
Protecting society requires examining the societal threat landscape---the set of pathways through which AI systems pose direct or indirect risks to society and the biosphere \cite{mallah_evaluating_2025}---yet most current assessment methodologies struggle to systematically map threats and evaluate risks across this broad landscape. This threat landscape includes not only direct technical effects, but also sociotechnical interactions and emergent behaviors arising from the interplay of advanced AI systems with their deployment environments and societal systems. Such behaviors may produce complex feedback loops, amplify systemic vulnerabilities, or trigger cascading effects across interconnected societal infrastructures \parencite{bengio_international_2025}.

Various risk assessment methods have been developed, including safety benchmarks \parencite{li_wmdp_2024, vidgen_introducing_2024}, model evaluations \parencite{shevlane_model_2023}, safety cases \parencite{clymer_safety_2024, carlan_dynamic_2024}, audits \parencite{sharkey_causal_2024}, responsible scaling policies (RSPs)\footnote{While termed ``responsible'' scaling policy, the nomenclature itself does not inherently ensure responsible implementation or outcomes; also known as frontier ``safety'' frameworks or ``preparedness'' frameworks.}
\parencite{anthropic_anthropics_2023, dragan_introducing_2024, openai_preparedness_2023}, and red teaming \cite{lee_learning_2024}. However, current methods face significant limitations in identifying and quantifying AI risks---especially unelicited risks that could have catastrophic consequences.

The reliability of AI safety evaluations fundamentally depends on their underlying assumptions. When assumptions regarding system behavior and mitigations fail, the entire safety assessment may be invalidated. Yet, current approaches often lack systematic documentation declaring and justifying these critical assumptions, making it impossible to verify the scope and limitations of their safety claims \parencite{barnett_declare_2024}.

This lack of systematic documentation is particularly concerning given that recent analysis shows closed-source AI models are only slightly ahead of open-source alternatives \parencite{cottier_how_2024}. With such a narrow gap, potentially dangerous capabilities could rapidly disseminate from leading models to widely accessible systems \cite{seger_open-sourcing_2023, hintersdorf_balancing_2023, kilian_examining_2023}, posing risks of widespread societal harm before adequate safeguards can be developed and implemented. 
The absence of systematic approaches to measure and manage AI risks severely hampers both governance efforts and responsible deployment \cite{kasirzadeh_measurement_2024}. 
These challenges call for more structured approaches to risk assessment, even as we acknowledge that no current methodology can provide complete safety guarantees.

\textbf{Risk quantification lessons from high-reliability industries.} 
The most dependable quantitative assurances regarding advanced AI systems will necessarily be those that are provable. 
Absent that standard, evidence and arguments can establish other levels of assurance, depending on their quality. 
To that end, probabilistic risk assessment (PRA) is a systematic approach to quantifying risk by evaluating both the likelihood of adverse events and the severity of their potential consequences through structured analysis of hazard pathways. Adapting PRA offers promise, leveraging its structured approach for AI risk assessment. Yet, the unique dynamics of AI systems—such as adaptability, emergence, novel failure modes, and complex sociotechnical interactions—present challenges beyond the typical scope of traditional applications that often focus on component-based failures, motivating the adaptations presented here. 
The approach gained prominence in the aerospace industry during the Apollo space program \cite{stamatelatos_probabilistic_2002}, and has been adopted in quantitative risk estimates for various complex, high-reliability industries such as nuclear power \cite{maidana_supervised_2023, zamanali_probabilistic-risk-assessment_1998, tudoran_sciendo_2018}, chemical manufacturing \cite{coleman_multi-hazard_2016, us_epa_risk_2015}, waste management \cite{lester_site-specific_2007, apostolakis_concept_1990}, and aerospace \cite{maggio_space_1996, stamatelatosmichael_probabilistic_2011}. 
For example, in nuclear power plant safety assessments, the PRA methodology identifies and quantifies the probability of event sequences that could lead to core damage, allowing engineers to implement targeted safety measures at critical points in the system. 
PRA combines quantitative risk metrics and system modeling to analyze potential failures, with central sub-techniques including hazard identification, event sequence modeling, failure mode analysis, and uncertainty quantification. 

Crucially, applying quantification to advanced AI that lacks technical assurances---particularly for novel or low-probability, high-impact events where historical data is scarce or non-existent---necessitates structured estimation within coarse-grained bands (e.g., orders of magnitude for likelihood and severity) rather than seeking precise point probabilities derivable from actuarial data. The use of defined bands aligns with established interval-based approaches, such as probability bounds analysis, employed in risk assessment \cite{shortridge_risk_2017}.
This structured estimation approach, central to our adapted framework, enables a degree of reasoned analysis under conditions of significant uncertainty.

Traditional PRA implementations have demonstrated particular strengths in several key areas:
\begin{itemize}
    \item \textbf{Identifying critical risk scenarios.} Recognizing and prioritizing potential failure modes in complex interconnected systems.
    
    \item \textbf{Quantifying rare events.} Assessing severity and likelihood of rare, high-consequence events where empirical data may be limited.
    
    \item \textbf{Structured uncertainty analysis.} Enabling systematic tracking of uncertainty through causal chains and system dependencies.

    \item \textbf{Multi-hazard assessment.} Accommodating diverse internal and external hazard sources and their interactions.

    \item \textbf{Consistent risk communication.} Establishing a common vocabulary for risk communication across organizational boundaries.
    
\end{itemize}
These factors make PRA a valuable approach for assessing complex systems where empirical failure data may be limited but systematic analysis is still required.

Probabilistic risk assessment methods have shown particular promise in analogous domains requiring systematic analysis and reasoning about risks under uncertainty. 
PRA as a unified framework could support more productive audits, critical discussions, and identification of blind spots. Additionally, PRA offers structured approaches for considering complex interconnected risks, including low-probability, high-impact scenarios. 
Recent developments support this direction, with AI developers beginning to incorporate probabilistic measures into their evaluation workflows through formal uncertainty quantification \parencite{miller_adding_2024}, and regulatory frameworks such as the EU AI Act increasingly emphasizing the importance of quantifying uncertainty in AI system evaluation \parencite{eu_second_2024}.

Effectively applying PRA to AI systems requires substantial methodological innovation due to three key challenges: AI systems' adaptability to new contexts, their inscrutability (or difficulty of inspection), and their capacity for emergent behaviors that can qualitatively change over time. Unlike traditional applications where PRA typically relies on historical failure data and well-understood system behaviors, AI systems demonstrate novel behaviors and operate in rapidly evolving deployment contexts. Furthermore, while traditional PRA often focuses on technical system failures, AI risk assessment must address broader societal impacts.

A framework applying PRA to AI systems must therefore account for capabilities without historical precedent and enable assessors to methodically identify and prioritize high-consequence risks across a broader threat landscape, rather than relying on selective assessment of commonly cited hazards. These limitations require the methodological innovations that we introduce in our ``PRA for AI'' framework, specifically tailored to address the unique characteristics and risk profiles of AI systems.

\textbf{Adapting PRA for AI systems.} Building on PRA's established strengths as a tool for risk estimation, we introduce a framework for AI risk assessment that enables the systematic integration and the synthesis of theoretical and empirical evidence, documentation of underlying assumptions and reasoning, and production of quantified risk level estimates. Unlike current risk assessment approaches that often lack transparency in their assumptions, our framework explicitly documents the reasoning behind risk estimations, which is particularly crucial for AI systems where emergent behaviors may invalidate unstated assumptions.

Our framework introduces several key methodological advances: 
\begin{itemize}
    \item \textbf{Aspect-oriented hazard analysis.} Systematic indexing of the AI hazard space through sampling of adjacent hazards through capabilities, domain knowledge, affordances, and impact domains.
    \item \textbf{Risk pathway modeling.} Analyzing causal paths through which AI system aspects enable or trigger harms that amplify and propagate through interconnected societal systems.
    \item \textbf{Uncertainty management.} Decomposition of complex risk scenarios into analyzable components and explicit uncertainty documentation, supported by evaluation scales and assessment tools with reference examples.
\end{itemize}

These advances are crucial for AI systems due to their unique characteristics of adaptability, inscrutability, and emergent behaviors. Together, they enable assessment of both direct harms and indirect societal impacts while maintaining detailed documentation of assumptions and evidence.

In Figure \ref{Fig 1 Overview} we provide an overview of the PRA for AI framework in operational context, illustrating how assessors integrate information about AI system behaviors, specifications, and expert inputs to produce risk level estimates and rationales that inform decision-making. The figure demonstrates the iterative relationship between the analytical models and empirical evidence that occurs within a single assessment pass.

At a high level, the framework integrates diverse sources of information through two complementary channels: Analytical models semi-formalize the theoretical basis for system behavior and supply structured risk scenarios, while empirical evidence provides quantifiable measurements and observations from system testing, whitebox analysis, and any prior deployment with comparable AI systems to inform risk estimates, yielding calibration data and observed risk indicators that inform risk estimates.

The analytical models inform the design of Test, Evaluation, Validation, and Verification (TEVV) activities \cite{nist_ai_2022-1}, which guides empirical data collection---while the empirical evidence collection provides calibration data for the analytical models at use, and the model of system behavior. This approach enables systematic calibration of the analytical models against empirical findings, while simultaneously informing the design of empirical assessment investigations through model-driven hypotheses. Both channels contribute to a process of harmonizing analyses.

The harmonizing analyses compile multiple sources of risk information---from novel risk scenarios generated during the assessment to known results from other risk assessment methods such as benchmarking and red teaming---and together with all available evidence reconciles the information into cohesive risk level estimates. For example, when assessing an AI system's potential unauthorized access to sensitive information, the analytical models might identify theoretical attack vectors while empirical testing provides data on actual vulnerability exploits, allowing for a reasoned risk estimate grounded in both possibility and probability.

The resulting cohesive risk level estimates are automatically synthesized into a report card. The report card, together with additional output documents created in the assessment process that provide information about the risk level estimates and their rationales, should feed into risk-informed decisions where the risk level estimates are considered with other evidence and acceptance criteria.

In the following sections, we present the PRA for AI framework in detail, beginning with a critical review of current AI risk assessment challenges and methods, their limitations, and PRA as a potential solution (Section \ref{background}). Building on this foundation, we introduce our adapted methodology (Section \ref{pra-application-to-ai}), showing how aspect-oriented hazard analysis, risk pathway modeling, and uncertainty management work together to assess AI risks systematically. We then demonstrate how organizations can implement this framework through our workbook tool, providing concrete guidance for conducting assessments (Section \ref{framework-implementation}). Finally, we analyze the framework's methodological contributions, practical utility, limitations and future directions (Section \ref{discussion}), and conclude with the framework’s contributions to AI risk assessment (Section \ref{conclusion}).

\section{The AI Risk Assessment Landscape} 

\label{background}

\subsection{Introduction to AI Risk Assessment}

The assessment of AI systems shares key parallels with traditional probabilistic risk assessment of complex engineered systems, but presents novel challenges. Similarly to nuclear control systems, AI systems exhibit complex feedback loops and potential for cascading failures \cite{moustafa_preventing_2021, phadke_expose_1996, kasirzadeh_two_2025}. However, traditional methods that work well for physical systems---such as fault tree analysis mapping discrete component failure modes---can prove insufficient when applied to AI systems, which can actively generate novel failure paths not captured in standard event sequence diagrams.

One key difference is that, unlike most hardware, modern AIs cannot be easily broken into smaller mechanistic subcomponents. Power infrastructure can be assessed by diagramming the subcomponents of the system (generators, transfer lines, transformers, breakers, consumers, etc.), analyzing their connections (series or parallel circuits, redundancy, etc.), and mathematically combining the known or modeled failure rates of those subcomponents. While similar component-based analysis can be applied to traditional software systems with well-defined modules and functions, modern neural AI systems do not lend themselves well to this approach, despite the best efforts of mechanistic interpretability researchers \cite{bereska_mechanistic_2024}.

Another key difference lies in strategic depth: while engineered systems follow immutable physics and hard-coded decision algorithms, the decision algorithms followed by modern AIs are opaque, evolved rather than designed, and often in flux, subject to rewriting by fine-tuning, new releases, or (in advanced cases) the AI itself \cite{sogaard_opacity_2023}. This makes standard mean-time-between-failure calculations and reliability block diagrams insufficient. Instead of P(failure) being derived from component-level probabilities, we must consider a time-varying failure surface where the system itself can discover and exploit previously unknown failure modes---for example, an AI system might develop novel ways to satisfy its objective functions that were not anticipated by its designers.

Traditional risk assessment methods face two additional challenges here: empirical testing provides only limited insight into the true distribution of latent risks, and risks propagate through interconnected systems with strong amplification effects \cite{mukobi_reasons_2024, barnett_what_2024}.

To a traditional risk assessor, attempting to model modern AI is analogous to modeling a control system whose logic could spontaneously rewrite itself during operation. To be useful, new methods must therefore extend beyond traditional probabilistic approaches to account for this fundamentally different class of hazard.

\subsection{Fundamental Challenges in AI Risk Assessment}
\label{uniquechallenges}

The assessment of general-purpose AI systems presents unprecedented assessment challenges due to their increasing complexity, paucity of provable constraints, and ability to autonomously discover and exploit vulnerabilities across broad attack surfaces. These systems can modify their behavior through learning, situationally aware reasoning \cite{laine_me_2024, laine_towards_2023}, and self-improvement \cite{huang_large_2022, zelikman_self-taught_2024} and encounter scenarios far outside their training distribution \cite{liu_ai_2023}.

Furthermore, AI systems can generate impacts that scale rapidly and diffuse through societal systems \cite{weidinger_sociotechnical_2023, critch_tasra_2023, aguirre_close_2024}. Through these capabilities, such systems can develop goal-directed behavior and environmental awareness that may lead to loss of human control \cite{jarviniemi_uncovering_2024, hendrycks_x-risk_2022}. This is particularly true for agentic AI systems, where capabilities like autonomy and memory introduce novel failure modes, e.g., agent compromise, multi-agent jailbreaks, memory poisoning \cite{bryan_taxonomy_2025}. Unlike most traditional engineered systems, AI systems can operate outside prescribed contexts with greater speed, scale, and sophistication \cite{critch_ai_2020}, making historical closed-domain risk patterns insufficient and requiring explicit modeling of novel propagation mechanisms and amplification pathways. Moreover, interactions between multiple AI systems can lead to emergent behaviors and risks that are difficult to predict from analyzing systems in isolation \cite{hammond_multi-agent_2025}.

\textbf{Types of risk awareness and understanding.} To organize these challenges, AI risks can be categorized using what is commonly known as a Rumsfeld matrix (Table \ref{tab:uncertainty_matrix}), which distinguishes between different states of awareness and understanding. This provides a structured way to tailor methodologies for different risk types.

AI systems create distinct assessment challenges due to varying degrees of awareness and understanding of system behavior and risks. Known risks that we understand are typically addressed using traditional quantitative approaches grounded in empirical data. However, even these ``known knowns'' in AI systems are context-sensitive, with their manifestation varying significantly based on system state. ``unknown knowns'' arise from methodological blind spots in assessment approaches, leading to latent risks that often surface during deployment under unmodeled conditions. ``known unknowns,'' such as emergent behaviors and capability jumps, represent acknowledged gaps in our understanding, which are difficult to characterize due to the absence of comparable historical data. Finally, ``unknown unknowns'' encompass unexpected or unforeseeable risks that push the boundaries of our ability to assess them, requiring adaptive strategies to reason about and prepare for entirely novel failure modes, as well as allowing for ample buffer in risk assessed.

\begin{table}[ht]
\caption{Awareness-understanding matrix for AI risk assessment.}
\vspace{5pt}
\renewcommand{\arraystretch}{1.5}
\fontsize{7.5}{9}\selectfont
\makebox[\textwidth][c]{
\begin{NiceTabular}{|p{1.3cm}|p{5cm}|p{5cm}|}
\hline
\textbf{Knowledge} & \textbf{Known} (Aware) & \textbf{Unknown} (Not aware) \\
\hline
\textbf{Known} & \textbf{Known Knowns:} Risks we are aware of and understand. & \textbf{Unknown Knowns:} Risks we are not aware of but do understand or know implicitly.\\
(Understand) & 
\textbf{Examples:} Empirically verified failure modes---such as instances where an AI system consistently exploits clearly defined reward functions in unintended but predictable ways---are well documented and reproducible through established testing protocols. & 
\textbf{Examples:} Risks may be overlooked within existing testing methods---for example, blind spots where certain edge cases are not adequately covered---that are theoretically understood but not yet detected in current practice.\\
&
\textbf{Methods:} Empirical measurement, quantification, systematic hazard space reduction.
&
\textbf{Methods:} Deployment monitoring, rigorous testing, critical reviews of assumptions.  \\
\hline
\textbf{Unknown} & \textbf{Known Unknowns:} Risks we are aware of but don't understand. & \textbf{Unknown Unknowns:} Risks we are neither aware of nor understand.\\
(Don't 

Understand) & 
\textbf{Examples:} Based on established scaling laws and capability trajectories, discontinuous advances in system capabilities and emergent behaviors can be anticipated, although their precise manifestations and implications remain uncertain. & 
\textbf{Examples:} There may exist entirely unforeseen system behaviors or interactions---for instance, novel failure modes triggered by complex, unanticipated factor combinations—for which no current data or prior indications exist.\\
&
\textbf{Methods:} Scenario modeling, projection simulations, forward-looking threat modeling.
&
\textbf{Methods:} Adaptive threat modeling, failure mode exploration, iterative assessment.\\
\hline
\end{NiceTabular}
}
\label{tab:uncertainty_matrix}
\end{table}

\normalsize

AI risks manifest across different contexts of operation and impact, referred to here as assessment domains. Each of these domains---from technical system internals to broader societal impacts---exhibits distinct patterns of uncertainties as categorized in the matrix. Understanding how awareness and understanding of risk vary across domains can help guide the development of appropriate assessment methodologies. Technical aspects of AI systems may be more amenable to empirical measurement and quantification (primarily involving known knowns), while operational contexts could, for example, reveal a wide variety of known unknowns during deployment. The broadest challenges emerge when considering societal impacts, where complex interactions create some unprecedented risks that are currently neither understood nor recognized.

\textbf{Assessment domains.} The complexity of these risks becomes more apparent when viewed through the lenses of distinct assessment domains, which organize hazards based on their meaningful interactions and information availability. 
These domains form an interconnected chain---from internal system dynamics through system-environment interactions to societal diffusion and then on to ultimate impact domains. The assessment challenges are significant at both ends of this chain: internal dynamics exhibit high-dimensional complexity and resist interpretation, while societal propagation creates complex systemic effects. In between, where systems interact with their immediate operational contexts, mechanisms tend to operate through more interpretable, lower-dimensional pathways, though remaining equally critical.

Causal analysis within each domain and an understanding of the system’s inherent variability—including its potential for diverse interactions and configurations—are useful primary considerations when initially examining these areas. Each assessment domain must be analyzed not only for its internal characteristics but also in the context of these interdependencies, presenting distinct challenges for assessment:

\begin{itemize}
    
    \item \textbf{Internal System Dynamics.} Hazards from a system's internal states, mechanisms, logic, or learned behaviors.
    \begin{itemize}
        \item \textbf{Capability assessment.} Systems demonstrate uneven development edges, where they can both dangerously excel and catastrophically fail in unexpected ways, defying standard performance metrics.
        \item \textbf{Higher-order capabilities.} Risks often arise from the unexpected interactions of two or more different capabilities within a system. 
        \item \textbf{Agentic behavior modeling.} The potential for autonomous goal-directed behavior, emergent goals and drives, and strategic adaptation creates novel challenges for modeling system evolution and failure modes across different competency levels. 
        \item \textbf{Strategic deception.} Hazards arising from a system learning to intentionally misrepresent information or conceal its internal state, capabilities, or operational intentions from operators or other systems. 
    \end{itemize}
   
    \item \textbf{System-System Interactions.} Hazards from interactions between systems.
    \begin{itemize}
        \item \textbf{Control measure evaluation.} Safety measures that work in testing may fail or be circumvented in deployment.

        \item \textbf{Multi-agent failure modes.} Interactions between AI systems generating novel failure pathways or collective behaviors (e.g., harmful coordination, unforeseen competition) that would not arise from a single agent's behavior.
    \end{itemize}
    \item \textbf{System-Environment Interactions.} Hazards from interactions with the surrounding systems, where external factors influence behavior.
    \begin{itemize}
        \item \textbf{Risk accumulation and amplification.} Seemingly minor risks can combine and amplify through system interactions, creating systemic consequences that evade analysis of components in isolation.
        
        \item \textbf{Impact measurement.} Standard metrics and proxy measures often fail to capture actual safety properties, particularly for systems capable of unprecedented behaviors.
        
        \item \textbf{Risk pathway genesis.} The systematic identification of where and how risks originate requires exploring system characteristics that act as sources which initiate harm pathways, 
        examining both direct triggers and enabling conditions while maintaining principled prioritization of highest-severity outcomes.

        \item \textbf{Interaction modeling.} Risks manifest from the capability combinations interacting with their environments, leading to unpredictable dynamics that require modeling beyond traditional methods.
        
    \end{itemize}
    \item \textbf{Societal Diffusion.} Hazards propagating through sociotechnical contexts.
    \begin{itemize}
    
        \item \textbf{Expanding societal threat landscape.} General-purpose AI capabilities evolve and find new applications, continually broadening potential pathways for societal harm, challenging conventional bounded analysis.
        
        \item \textbf{Systemic risk propagation.} Technical risks transform as they transmit through interconnected systems, creating novel threat vectors that transcend traditional risk boundaries; in some cases diffusing, while in others accumulating, amplifying, or concentrating in some particular directions. 
        
        \item \textbf{Misuse pathway modeling.} Systematic analysis of intentional misuse pathways requires modeling both sophisticated targeted attacks leveraging system competencies and opportunistic abuse exploiting system limitations.

        \item \textbf{Sociotechnical amplification effects.} Bidirectional feedback between social and technical systems creates emergent behaviors and multiplicative impacts that traditional assessment frameworks fail to capture.
    \end{itemize}
\end{itemize}

Each assessment domain builds upon and interacts with the others, with risks often propagating and amplifying across multiple domains simultaneously. For example, internal system capabilities can enable novel system-to-system interactions, which in turn create new environmental hazards that ultimately manifest as societal impacts. This progression reflects not just increasing complexity of interactions, but also growing difficulty in detecting leading indicators and diminishing ability to run meaningful tests.

\subsection{Limitations of Current AI Risk Assessment Methods}
\label{limitationsofcurrentmethods}

Despite a well-established literature on risk management \cite{gahin_review_1972}, system safety engineering \cite{ericson_ii_system_2005}, reliability engineering \cite{bergman_development_1992}, and probabilistic risk assessment (PRA) \cite{modarres_probabilistic_2008}, there has been a notable paucity of application of these approaches to general-purpose AI systems. 
Efforts to address AI risks through frameworks such as NIST's AI Risk Management Framework \cite{nist_ai_2022}, and standards including ISO/IEC 23894:2023 \cite{iso_isoiec_2023-1} and ISO/IEC 42001:2023 \cite{iso_isoiec_2023} have focused primarily on organizational processes and controls. 
While valuable, these initiatives have limitations. They tend to defer to model providers' priorities and values rather than addressing broader societal risks, quantifying risk, or establishing guarantees. Furthermore, they have yet to demonstrate an ability to extend themselves to account for the unique challenges posed by advanced AI systems.

Current AI-specific risk assessment methods\footnote{For frameworks such as RSPs and safety cases, which encompass broader governance or assurance functions respectively, this analysis focuses specifically on the assessment methods they employ, prescribe, or are built upon, and the limitations of those methods in assessing AI risk well.}, while diverse in their approaches, reveal significant gaps and limitations in their ability to comprehensively evaluate advanced AI systems. The prevailing landscape is dominated by six primary approaches: safety benchmarks, model evaluations, red teaming, RSPs, safety cases, and audits. Each of these taken alone face significant challenges.

\textbf{Safety benchmarks.} In contrast with capabilities benchmarks, which measure a system's performance on some task, safety benchmarks purport to measure some feature of a system relevant to AI safety \parencite{li_wmdp_2024, vidgen_introducing_2024, bhatt_purple_2023}. They provide quantifiable metrics but face four key limitations. First, they often serve as capability proxies rather than true safety measures---higher performance frequently indicates greater overall system sophistication rather than improved safety, allowing capabilities research to be ``safetywashed'' as safety research \parencite{bucinca_proxy_2020, ren_safetywashing_2024}. Second, even when successfully measuring capabilities, they can only establish lower bounds, leaving significant uncertainty about the full extent of a system’s actual capabilities and potential failure modes \cite{barnett_what_2024}. This uncertainty stems partly from the fact that benchmarks typically struggle to assess a system's goal-directedness---its propensity to effectively utilize measured capabilities towards specific objectives within relevant contexts \cite{everitt_evaluating_2025}. Third, they suffer from under-elicitation---their narrow test cases, whether formulaic or ad hoc, and their controlled environments fail to reveal the true range of system behaviors and potential risks that could emerge in real-world deployments \cite{vidgen_introducing_2024}. Finally, benchmarks are rapidly saturating---new tests such as GPQA reach human-level performance within months of release, making them increasingly ineffective for bounding risky capabilities \cite{rein_gpqa_2023, dominguez-olmedo_training_2024}.

\textbf{Evaluations.} Model evaluations are tests performed on particular AI systems to elicit their potential for causing harm \parencite{shevlane_model_2023, aisi_ai_2024}. These evaluations can assess system characteristics broader than specific benchmarks and employ different elicitation strategies. In the case of misuse potential, evaluations are sometimes run using ``human uplift studies'' to assess how much a specific AI system improves human performance across some set of tasks \cite{aisi_advanced_2024}. However, evaluations remain fundamentally constrained by their parochial and shallow testing approach, relying on a predetermined and limited set of harm scenarios---similar to surface excavations that cannot reveal what lies deeper \cite{burden_evaluating_2024}. The controlled testing environments and predefined scenarios in these evaluations fail to capture the full range of system behaviors or systemic risks that could emerge in real-world deployments \cite{jones_under_2024}. Even for identified risks, evaluations can only establish lower bounds on capabilities, leaving substantial uncertainty about full system potential \cite{barnett_what_2024}.

\textbf{Red teaming.} Red teaming, manual or automated, involves systematically testing AI systems through adversarial approaches and misuse scenarios \cite{lee_learning_2024}. While this approach provides valuable insights into potential failure modes, it faces several critical limitations. First, its reliance on demonstrable failures means it systematically overlooks deeper flaws that could manifest in deployment---facing the same limitations as evaluations in probing only the surface \cite{anthropic_challenges_2024}. Second, like benchmarks, red teaming can only establish lower bounds through selective testing, failing to provide comprehensive safety assurances \cite{barnett_what_2024}. A fundamental challenge is that testers often cannot determine whether they have adequately elicited the system's full capabilities---failed attempts to demonstrate a capability do not prove its absence, and in practice, under-elicitation is likely to be the norm rather than the exception. Third, current approaches lack quantitative risk measurements and systematic coverage of the threat space \cite{feffer_red-teaming_2024}, making it difficult to assess both current risks and their future evolution. Fourth, as systems become more capable, red teams increasingly struggle to maintain effectiveness against models that can detect and adapt to testing scenarios \cite{ganguli_red_2022, feffer_red-teaming_2024}. Additionally, red teaming results remain siloed from other evaluation methods, limiting their utility for comprehensive risk assessment \cite{friedler_ai_2023}.

\textbf{RSPs.} RSPs are risk management frameworks adopted by AI developers to ostensibly attempt to mitigate catastrophic risks \parencite{anthropic_anthropics_2023, dragan_introducing_2024, openai_preparedness_2023}. While RSPs offer 
ostensive 
processes for evaluating scaling decisions and establishing safety thresholds, they face four fundamental limitations undermining their effectiveness as risk assessment tools. 
First, their reliance on coarse-grained risk categories reduces complex, multi-dimensional risk scenarios into overly simplified buckets, making it difficult to capture the actual severity of risks, distinguish between varying levels of concern, or detect unexpected risks \cite{uuk_effective_2024, titus_can_2024}. 
Second, RSPs employ narrow threat models, often focusing on misuse and specific technical capabilities (e.g., deception, situational awareness) while overlooking systemic risks, interactions among different capabilities, or interactions between multiple systems. This narrowness, combined with poorly informative risk and capability levels, yields few meaningful distinctions in risk degrees, leading to assessments that fail to provide actionable insights into a system's capability and harm potential. The resulting compressed reasoning chains mean conclusions about system dangers often rest on weakly justified inferences from observed capabilities. 
Third, while predicting discontinuous capability jumps is inherently difficult, RSPs' focus on evaluating currently demonstrable abilities against predefined milestones provides inadequate mechanisms to systematically anticipate or incorporate the risk associated with potential jumps emerging from seemingly incremental advances---a critical flaw when evaluating rapidly developing AI systems that may exhibit unpredictable emergent behaviors. 
Fourth, their effectiveness is further undermined by several procedural and governance weaknesses: they frame continued scaling as the default rather than requiring justification for capability increases \cite{anderljung_frontier_2023, heim_training_2024}; they lack quantifiable thresholds that could prevent loose interpretation; and they typically rely on internal evaluation without sufficient external oversight. 
This combination of limitations results in assessment frameworks that provide limited insight into the full risk landscape while potentially reducing the urgency for more comprehensive safety measures.

\textbf{Safety cases.} A safety case is a structured argument, supported by evidence, that a system is sufficiently safe for a given application in a specific context \parencite{goemans_safety_2024, clymer_safety_2024, buhl_how_2025, habli_big_2025}. While safety cases provide relatively rigorous methods for demonstrating system-level safety, their emphasis on constructing a convincing affirmative argument for safety (i.e., justifying why a system is safe) can struggle when assessing advanced AI systems. Two key challenges emerge. First, the drive to build a tractable positive case can lead assessors to define a convenient bounded context narrower than the AI’s actual or potential deployment environment \cite{stix_ai_2025, obrien_deployment_2023}, potentially defining away and overlooking out-of-scope risks \cite{mylius_systematic_2025, barnett_declare_2024}. Second, safety cases rely on logical argument structures where a single weak link can undermine the entire safety argument. This brittleness means that the failure of a single claim regarding AI capabilities or the efficacy of control measures in the argument chain can invalidate the entire safety case.

\textbf{Audits.} AI ``safety audits'' are oversight mechanisms that aim to ensure AI is developed and deployed responsibly, ranging from narrow compliance checks to deep bespoke audits of a particular risk \parencite{for_humanity_independent_2016, sharkey_causal_2024}. AI audits are typically narrowly focused on a particular aspect of an organization or system. They face fundamental structural and technical limitations in assessing advanced AI systems. Structurally, they encounter an inherent tension between independence and system access---internal audits have deeper access but lack independence, while external audits maintain independence but struggle with system access and complexity \parencite{schuett_frontier_2024}. Current black-box audit  approaches provide limited societal threat surface coverage, offer poor support for prospective risk analysis, and are not typically informed by a specific system aspect \parencite{casper_black-box_2024}. They become prohibitively resource-intensive when attempting thorough coverage of advanced AI systems and tend to include small sets of threat models per audit. Technical limitations mirror those of other evaluation approaches---audits cannot establish upper bounds on capabilities, reliably detect novel failure modes, or assess risks from autonomous systems \parencite{barnett_what_2024}.

\textbf{Common method limitations.} The current landscape of AI risk assessment is characterized by significant fragmentation \cite{xia_towards_2023}. Different approaches---from benchmarks to audits---remain siloed, addressing narrow aspects of risk while failing to integrate their findings into a unified perspective. 
This fragmentation creates blind spots, particularly in identifying and evaluating novel risks that emerge only when evidence from multiple methods is combined. The lack of harmonization makes it difficult to reconcile contradictory signals or to combine historical data with forward-looking projections of possible pathways.
Beyond fragmentation, these methods demonstrate significant gaps in modeling critical aspects of AI risk. They fail to adequately capture how technical risks can transform and amplify as they propagate through interconnected societal systems, or how different system capabilities interact in complex ways with their environments. 
Most critically, they often miss systemic vulnerabilities that only become apparent when examining the full sociotechnical context in which AI systems operate and the multiple feedback loops between technical and social systems \cite{weidinger_sociotechnical_2023}.

While individual methods may capture specific aspects of AI risks, none provide a structured framework for analyzing how different system capabilities, high-risk domain knowledge, and operational affordances could---directly or in combination---propagate through societal systems to create harm. Where traditional approaches might successfully identify and mitigate specific technical vulnerabilities, they fail to capture how AI systems could affect the resilience of economic, legal, normative, and social systems---each themselves complex, adaptive, and fundamental to individual liberty and societal functioning. For a more detailed examination of risk assessment methods, see Appendix \ref{app: Comparison Matrix}.

\textbf{Common myopia.} 
Beyond specific limitations of individual methods, there are deeper implicit assumptions often shared across explicit and implicit risk assessment approaches---that risks and harms are uncommon, manifest in obvious ways, and can then be patched when seen. 
These assumptions reflect a legacy risk thinking paradigm rooted in assessment practices developed for traditional well-bounded systems. 
While this thinking provides utility for specific parochial threat models and informs some mitigations---such as behavioral guardrails \cite{jain_baseline_2023}, model-level restrictions \cite{xie_defending_2023}, runtime monitoring \cite{zhou_robust_2024}, and input/output filtering \cite{inan_llama_2023}---such paradigms fail to fully account for the unique complexity and adaptability of advanced AI systems. 
Consequently, the mitigations they inform can leave significant residual risks---the risk remaining after implementing these controls---unaddressed even after thorough application of their principles. 
Critically, many current approaches lack systematic methods for exploring the range of AI system characteristics (e.g., latent capabilities or specialized knowledge) that could enable hazards, nor do they adequately model the diverse real-world harms that can be expected to ultimately be realized by system deficiencies and capabilities. Assessment practices frequently remain focused on specific, measurable risks (e.g., bias in a particular task or success rate on known misuse prompts) without integrating these into a holistic assessment that traces potential causal pathways from underlying system properties to concrete, high-consequence societal outcomes. 
This failure to systematically connect system properties to their potential real-world consequences means that significant portions of the societal threat landscape are often overlooked.

\subsection{Adapting Established Probabilistic Risk Assessment Methods for AI}
\label{EstablishedtraditionalPRA}

PRA has demonstrated its value in assessing complex systems with catastrophic failure modes across multiple high-reliability industries, including nuclear power \parencite{tudoran_sciendo_2018, zamanali_probabilistic-risk-assessment_1998}, aerospace \parencite{stamatelatosmichael_probabilistic_2011}, waste management \cite{apostolakis_concept_1990}, and chemical processing \parencite{us_epa_risk_2015}. PRA's adoption by major regulatory bodies \parencite{zamanali_probabilistic-risk-assessment_1998, us_epa_risk_2015, stamatelatosmichael_probabilistic_2011, us_nuclear_regulatory_commission_severe_1990, lester_site-specific_2007}, and its integration into safety-critical industries where failure consequences can be severe speak to its success in addressing complex risks. These applications have shown particular strength in handling scenarios with limited historical data, complex interaction effects, and the need to synthesize expert judgment with empirical evidence---challenges that closely parallel those in AI risk assessment.

PRA methods have also been widely operationalized through ``assurance level'' frameworks across safety-critical industries. Aviation's Design Assurance Levels (DALs), such as DO-18C, use probabilistic failure analysis to determine required safety measures \cite{rapita_-178c_2012}; they are particularly useful in meeting stringent safety requirements, including those that demand a probability less than 10\textsuperscript{-9} per flight hour of catastrophic failure. Cybersecurity Assurance Levels (CALs) are a structured framework within ISO/SAE 21434 \cite{iso_isosae_2021} to classify the required rigor for cybersecurity measures in automotive systems. While CALs themselves are deterministic, the risk assessment process that informs CAL assignment often employs probabilistic methods. Vehicle security standards such as UN R155 incorporate Threat Analysis and Risk Assessment (TARA) methods \cite{un_regulation_2021} where companies perform threat analysis and assign likelihoods to these threats to obtain risk values. These frameworks demonstrate how probabilistic approaches can be translated into concrete standards and requirements.

PRA has proven especially valuable in analyzing both internal/external hazards and multi-hazard scenarios---where multiple hazards occur concurrently or in succession \cite{aras_critical_2021}. This capability to systematically assess multiple interacting hazards is particularly relevant for AI systems, where risks can manifest through various combinations and pathways.
The specific capabilities of PRA directly address key challenges in AI risk assessment. It excels at identifying critical risk scenarios in complex systems, quantifying rare but high-consequence events, providing structured approaches to uncertainty propagation, and integrating evidence into cohesive risk estimates \cite{us_nuclear_regulatory_commission_backgrounder_2024}.

PRA offers systematic frameworks that integrate multiple approaches to risk analysis, from component failure analysis to system-level interactions, rather than being constrained to any single risk assessment paradigm. 
PRA's strengths lie in well-established methods for analyzing complex systems, from component-level behavior to system-wide interactions. Key tools from traditional PRA that inform our framework include uncertainty quantification techniques, structured scenario development, hazard identification methods, and evidence integration approaches. These create an effective foundation for understanding and assessing risks in sophisticated engineered systems.

However, advanced AI systems present unique challenges that push beyond traditional PRA methods. Where conventional PRA relies on well-characterized failure modes and empirically derived probability distributions, AI systems can actively exploit vulnerabilities and exhibit emergent behaviors through unexpected capability interactions. These systems operate within complex sociotechnical contexts where technical risks can propagate through interconnected social systems in ways traditional PRA frameworks struggle to capture. Additionally, the rapid advancement of AI capabilities means that historical failure data may not reliably predict future risks, particularly when capabilities exceed human comprehension.

The field needs approaches that can analyze potential failure modes and their consequences while systematically handling increasing complexity and uncertainty across system boundaries. This motivates extending the established PRA methodology into a framework specifically adapted for AI systems.

\section{A Framework for AI Probabilistic Risk Assessment}
\label{framework}

\label{pra-application-to-ai}

Having established some of the challenges of AI risk assessment, limitations of existing methods, and strengths of traditional PRA, this section introduces a conceptual framework to help systematize AI risk analysis. This adaptation recognizes a fundamental trade-off; in assessing complex systems, one can pursue great analytical depth on narrow threats or achieve broad coverage across a wider range of potential risks, but rarely both simultaneously. The PRA for AI framework explicitly prioritizes breadth, relying on its Aspect-Oriented Taxonomy of AI Hazards (see \origcite{mallah_aspect-oriented_2025}) to guide systematic sampling of the hazard space across diverse system aspects and sociotechnical interactions; this focus reflects the principle that addressing the most significant questions across the broad societal threat landscape, even if yielding less precise answers (``a mediocre answer to the right question''), is often more critical for risk governance than achieving high precision on potentially narrower or less pertinent issues (``an excellent answer to the wrong question''). While this may mean the analysis of any single pathway is less detailed than highly specialized methods might allow, this structure is intended to facilitate the identification and evaluation of certain types of risks often missed by narrower approaches (see Section \ref{limitationsofcurrentmethods}), including systemic, novel, and combinatorial risks.

The framework introduces three key methodological advances, each supported by specific analytical techniques and tools:

\begin{enumerate}
    \item \textbf{Aspect-oriented hazard analysis.} Provides a top-down approach for assessors to index or iterate over the space of hazards, covering the characteristic aspect categories of an AI system: capabilities, domain knowledge, affordances, and impact domains. This taxonomy-driven method guides systematic analysis of how emerging AI capabilities could enable or amplify potential harms through critical bottlenecks.
        \begin{itemize}
            \item \textbf{Bottleneck analysis.} For each AI system aspect,  systematically examines potential harms by treating that aspect as the bottleneck---holding all other aspects constant while analyzing what risks emerge if this focal aspect were maximized within the system context.
            \item \textbf{Competence-incompetence analysis.} Examines risks arising, on one hand, from highly efficacious AI execution yielding harmful outcomes; on the other hand, from system limitations, flaws or errors manifesting as misunderstandings, vulnerabilities, or oversights; and from combinations of these where capability enables or exacerbates failings.
            \item \textbf{Aspect interaction analysis.} Examines how risks from one system aspect might interact with or amplify risks from another, helping to surface interaction effects that could otherwise be overlooked when analyzing aspects in isolation.
        \end{itemize}
    \item \textbf{Risk pathway modeling.} Guides modeling of the step-by-step progressions of risk from a system's source aspects (i.e., capability, domain knowledge, or affordance) to terminal aspects (i.e., impact domains where harms to individuals, society, or the biosphere manifest), and analyzes how risks transmit and amplify.
        \begin{itemize}
            \item \textbf{Societal threat landscape analysis.} Maps potential end-to-end pathways through which AI systems could directly or indirectly cause harm to society and its supporting biosphere, including consideration of sociotechnical contexts and flows. 
            \item \textbf{Prospective risk analysis.} Marshals a variety of analytical models, empirical evidence, and enhanced threat modeling for a forward-looking analysis of potential harms, rather than relying solely on historical failure statistics.
            \item \textbf{Propagation operators.} Descriptive mechanisms that 
            characterize how AI risks can permeate and impact societal systems, mapping how risks accumulate, transform, and amplify as they move through societal contexts and structures.
            \item \textbf{Focused aggregation.} An alternate grouping of assessed scenarios into focused categories that represent key dimensions of risk, enabling different stakeholders to view aggregated risk level estimates through lenses most relevant to their needs.
        \end{itemize}
    \item \textbf{Uncertainty management.} Guided, structured decomposition and management of uncertainties at each step of the assessment process, allowing assessors to conduct transparent, well-reasoned risk evaluations even in domains with limited historical precedent.
    \begin{itemize}
        \item \textbf{Classification heuristics.} Supporting tools with intuition pumps\footnote{``Intuition pumps'' are thought experiments that provoke or ``pump'' intuitions about a problem \cite{dennett_intuition_2013}; the structured scenarios used here provide one concrete form of such a pump.} for guiding and informing assessors, and helping to calibrate their judgement during threat model development, including plausible qualifiers for both competence and incompetence scenarios and tables laying out the spectra of capability and domain knowledge.
        \item \textbf{Intensity rubrics.} Standardized tables for harm severity and likelihood levels with concrete reference points across multiple societal dimensions, supporting consistent evaluation across different contexts and assessors.
        \item \textbf{Uncertainty tracing.} Methods to document uncertainties at each step of the assessment process, from more direct uncertainties in harm estimation to uncertainties in risk propagation and interactions, creating an auditable trail of assumptions and their compound effects.
    \end{itemize}
\end{enumerate}

Together, these methodological advances help build upon and adapt traditional PRA techniques to assess AI. They provide approaches for identifying, quantifying, and documenting the associated risks, establishing a structure for analysis that remains independent of specific operational constraints or assessment tools. 
The following subsections detail the three key advances introduced by the framework: Section \ref{aspect-oriented-hazard-analysis} presents aspect-oriented hazard analysis and its analytical techniques for identifying hazards by sampling the hazard space, Section \ref{risk-pathway-modeling} describes risk pathway modeling and methods for quantifying risks by analysis of causal forward and backward chains, and Section \ref{uncertainty-management} explains the framework's approach to uncertainty management with designated tools and techniques alongside guidance in documenting assessor reasoning. Later, Section \ref{framework-implementation} describes the practical implementation of these conceptual foundations through a workbook tool that guides assessors through the assessment process.

\subsection{Aspect-Oriented Hazard Analysis} 
\label{aspect-oriented-hazard-analysis}

Effective AI risk assessment requires moving beyond a repeated focus on commonly discussed risks \cite{jones_under_2024} towards a systematic examination of the hazard space to uncover the most consequential plausible threats of the given system. It is impossible to enumerate all the ways AI could cause harm, but preexisting methods for identifying hazards can be improved upon by attacking AI risk characterization from several different angles, since systematic coverage of key aspects allows us to bound and structure the otherwise intractable hazard space. The PRA for AI framework addresses this through aspect-oriented hazard analysis, a methodology designed for structured, multi-level exploration and sampling of hazards appropriate to the AI system under study. Instead of attempting exhaustive enumeration, aspect-oriented hazard analysis focuses on identifying critical risks by analyzing the system through the lens of its factored characteristics and context. This systematic, taxonomy-guided assessment of hazards and their potential severities enables assessors to perform analyses that achieve broader risk coverage by both interpolating from specific system findings to uncover ``unknown knowns'' and extrapolating from current understanding to identify ``known unknowns''. This approach empowers assessors to develop threat models that capture both obvious and non-obvious risk pathways, allowing for a more comprehensive mapping of the most likely and consequential threats by, respectively, characterizing known risk types more fully and identifying novel risk scenarios.

The framework operationalizes this systematic exploration through its Aspect-Oriented Taxonomy of AI Hazards (excerpted in Appendix \ref{app: taxonomy}). This structure originates from a top-down analysis starting from first principles of agent-world interaction and intelligent systems, examining the fundamental components necessary for an AI system to perceive, process, act within, and affect its environment (for further details see \origcite{mallah_aspect-oriented_2025}). This approach, analogous to seeking basic primitives and relationships in physics, aims to establish foundational categories independent of specific technologies, grounded in systems thinking about AI systems’ nature and context.

The taxonomy decomposes AI system characteristics into four high-level aspect categories (TL0), selected to provide a comprehensive, systems-based structure for analyzing risk origin and propagation:

\begin{enumerate}
    \item \textbf{Capabilities.} The ability of an AI system to perform specific tasks or functions from basic pattern recognition to complex reasoning and planning. This includes both intended functionalities and potential emergent capabilities that could enable harms. 
    \item \textbf{Domain knowledge.} The specific areas of expertise, information and understanding possessed by an AI system, and those that could enable harms, such as high-risk knowledge of cybersecurity vulnerabilities, biological systems or human psychology.
    \item \textbf{Affordances.} The inputs, configurations, and surroundings that enable an AI system to function and interact with its environment. This includes both designed interfaces and potential unintended access points. 
    \item \textbf{Impact domains.} The sociotechnical domains where the impacts of AI systems are realized, including individuals, society, or the biosphere, encompassing broad areas of influence where significant harm or benefits can occur.
\end{enumerate}

This high-level decomposition provides broad structural coverage by addressing: the AI's inherent functional potential (Capabilities), specific knowledge that can enable potential harm (Domain Knowledge), the mechanisms and environmental conditions enabling interaction and influence (Affordances), and the spheres where its effects manifest (Impact Domains). This structure helps the analysis systematically cover the origin, pathway, and endpoint of AI-driven hazards.

The taxonomy organizes these aspects in a hierarchical structure with five levels that progressively narrow from broad aspect categories to specific hazards that guide increasingly granular analysis:

\begin{itemize}
    \item \textbf{Aspect categories (TL0):} The highest-level classification representing the primary dimensions of AI system analysis (capabilities, domain knowledge, affordances, and impact domains);
    \item \textbf{Aspect groups (TL1):} Major subdivisions within these categories;
    \item \textbf{Aspects (TL2):} Specific system characteristics or elements within each group;
    \item \textbf{Hazard clusters (TL3):} Groups of related aspect-adjacent hazards into clusters, allowing for flexible categorization and cross-cutting analysis; and
    \item \textbf{AI hazards (TL4):} Individual aspect-adjacent hazards within the system's sociotechnical context.
\end{itemize}

This taxonomy provides a scaffolding and indexing for the vast space of AI-driven hazards, guiding assessors in generating, situating, and contextualizing specific techno-social concerns. Its practical utility comes from its ability to encompass and organize analysis across diverse hazard subsets. Recognizing that real-world AI risks are highly interconnected and often defy strict boundaries, the design intentionally permits some overlap between categories (rather than enforcing strict mutual exclusivity). This ensures more comprehensive coverage and facilitates analyzing interactions. For example, findings from specialized analyses—such as impacts on human rights, democracy, and the rule of law identified via dedicated assessments \cite{cai_methodology_2024}, or hazards identified by analyzing policy documents for deviations from codified societal expectations \cite{zeng_ai_2024}---naturally align within the taxonomy's Impact Domains (primarily subcategories, or sub-aspects, of Individual and Societal). Similarly, hazards originating from the system itself---stemming from AI Capabilities, high-risk Domain Knowledge, or operational Affordances---are also systematically categorized, providing structure for analyzing areas like cybersecurity or biological hazards or loss-of-control risks.

While aiming for comprehensive structural coverage at higher levels (TL0-TL2), the dynamic and vast nature of AI has led to the fact that the more granular lower levels of hazard clusters (TL3) and individual hazards (TL4), though populated with derived hazard entries, are not meant to be exhaustively pre-defined. These lower levels follow the top-down derivation and are further populated and informed by a bottom-up process incorporating illustrative examples of known failure modes identified from literature and practice. However, these prepopulated individual hazards (TL4) are illustrative rather than comprehensive: assessors must supplement these with hazards appropriate to the level, type, and idiosyncrasies of the AI under study as well as the sociotechnical scope of its use  (as defined upfront in the assessment). The progressively detailed structure facilitates targeted exploration.

Assessors additionally populate, refine, and prioritize TL3/TL4 entries that would be relevant for their specific AI system under consideration by drawing inspiration from specialized resources, supplementing the taxonomy as required. First, established hazard taxonomies from domains with mature risk‑assessment practice---like MITRE ATT\&CK® \cite{mitre_mitre_2025} in cybersecurity---can be adapted to ideate novel AI-specific attack vectors or misuse scenarios that fit within the framework taxonomy's categories. Adapting the “Phishing” technique under the “Initial Access” tactic from the MITRE ATT\&CK framework, for example, might reveal an AI-driven spear-phishing vector, which can be used to populate a corresponding specific hazard entry under the domain knowledge aspect of Psychology and Cognitive Science (TL3) of the framework taxonomy. Second, resources that catalogue AI-driven hazards in specific domains---such as financial systems \cite{financial_stability_board_financial_2024} or biosecurity \cite{rose_understanding_2023}---provide further context-specific hazard examples relevant to refining or expanding the corresponding TL3/TL4 entries in the framework taxonomy. Third, dedicated AI risk and incident repositories---such as the AI Risk Repository \cite{slattery_ai_2024}---help inform the ideation and identification of system-salient hazard types, by offering libraries of previously recognized prototypes drawn from relevant domain expertise and documented incidents. Considering regularly updated hazard repositories from different sources helps the assessment remain adaptable to emerging threats while maintaining coherence within the overall framework. Fourth, identification of novel threat models and hazards by assessors, subject matter experts, stakeholders, and human-curated LLM-assisted ideation (related to “scenario discovery” techniques in foresight for exploring future possibilities as shown in \origcite{li_review_2025}) is recommended. 

Aspect-oriented hazard analysis employs three complementary analytical approaches to systematically examine potential risks identified via the taxonomy: bottleneck analysis, competence-incompetence analysis, and aspect interaction analysis. These provide structured methods for identifying both direct technical hazards and more complex emergent risks that might arise from the interaction of system properties.

\textbf{Bottleneck analysis.} Enumerating all possible harms from AI systems is intractable given their vast number of potential pathways. 
Bottleneck analysis addresses this by examining how each aspect---whether a relatively atomic capability such as a particular type of reasoning, domain knowledge such as cybersecurity expertise, an affordance such as particular API access, or an impact domain such as collective epistemics---could become the critical constraint or enabler of harm. 
The analysis treats each aspect in turn as the potential bottleneck by holding all other aspects fixed while considering what becomes possible if the focal aspect reaches its maximum plausible level for the system being assessed. This reveals critical thresholds where quantitative enhancements in an aspect could enable qualitatively different risk scenarios.

For each aspect, assessors determine the criticality levels at which new harm pathways become possible. For example, when integrative cognitive orchestration (a reasoning capability aspect) is the focal aspect, assessors examine what novel harms become possible with the maximally sophisticated level of this capability that may manifest given the possible range of resources, contexts, reconfigurations, and uses of the system, even without advances in other aspects. The analysis helps determine, for example, what level of integrative cognitive orchestration capability enables sophisticated social manipulation or strategic deception.

Similarly, when collective epistemics (impact domain) is the focal aspect, assessors examine how the system---and what it may become within the defined assessment scope---at its maximum plausible capability level could exploit vulnerabilities in societal knowledge systems. 
This reveals critical dependencies---how moderate advances in societal vulnerability could suddenly enable massive exploitation by existing AI capabilities, or how weakened collective knowledge systems could unlock new categories of risk even without enhanced computational capabilities.

The key insight of bottleneck analysis is that it forces systematic consideration of how each aspect, when treated as the bottleneck, enables or exacerbates risks differentially. By examining how realistically maximizing a single aspect while holding others constant could affect the system's behavior, this analysis serves as a lens to reveal plausible threat models that might be missed in more conventional approaches. Each aspect provides a distinct perspective for identifying potential harms within the system being assessed. This helps identify which properties of the system warrant closest scrutiny and informs where additional safeguards may be needed.

\textbf{Competence-incompetence analysis.} Advanced AI systems create risks through a fundamental bifurcation that manifests in two distinct categories of hazards:
\begin{itemize}
    \item \textbf{Competence-based hazards.} Arise when highly effective capabilities lead to harmful consequences (succeeding at something we don't want);
    \item \textbf{Incompetence-based hazards.} Stem from system limitations or failures (seemingly trying but failing to do something we want).
\end{itemize}

While one could argue all failures reflect some underlying specification or design inadequacy, distinguishing between failures of execution (incompetence-based) and harms resulting from successful execution (competence-based) provides a valuable analytical lens. This distinction is particularly crucial for advanced AI due to the uneven advancement across different skills and domains (i.e., “jagged technological frontier”, or “capability discontinuities”) \cite{dellacqua_navigating_2023}.
As capabilities develop, AI systems may exhibit sophisticated performance in potentially harmful domains while simultaneously lacking critical safety-relevant competencies. For instance, when an AI system shows high competence in reasoning and planning but demonstrates incompetence in understanding ancillary consequences and following safety guidelines, it can cause harm through competent execution of unsafe plans. This pattern of advanced capabilities coexisting with significant blindspots---including hallucinations, confusions, and silent failures of reasoning---creates risk profiles fundamentally different from traditional software systems.

Established evaluation metrics (e.g., accuracy, robustness, fairness) readily quantify specific types of incompetence-based failures  \cite{bengio_international_2025, jones_under_2024}. Competence-based hazards, involving systems performing well but in ways that are harmful or unintended, are often harder to predict and assess comprehensively. This difficulty arises because evaluating these risks extends beyond verifying conformance to predefined functional specifications or testing for known types of execution failures. Instead, it requires identifying harmful outcomes emerging from a broad spectrum of potential system behaviors, many of which may successfully achieve goals in ways not easily specified in advance or captured by standard failure analysis. While targeted tests for specific issues like deception are developing \cite{meinke_frontier_2025, jarviniemi_uncovering_2024}, identifying the wide range of potential harmful potency realizations remains challenging compared to simply identifying functional failures, yet uncovering an ample set of competency-based hazards is particularly crucial, as such hazards are expected to generate larger harms. This dual competence-incompetence perspective is therefore critical for disentangling threat models and avoiding an imbalanced focus on only one type of hazard; it directly informs threat model development and the scenario generation process (see Section \ref{assessmentprocess})---illustrated by concrete competence/incompetence examples across severity levels in Appendix \ref{app: Detail Table}---ensuring both hazard categories, and their combinations, are considered during the assessment.

Applying this dual analysis reveals that highly capable AI systems can create larger risks through exceptional performance in harmful directions than through critical failures in expected functionality, though the latter can also be quite consequential in combination with the former. 
Moreover, critical combinations of high competence in some areas with incompetence in others often uncover novel risk pathways. The practical relevance of analyzing both categories is underscored by recent large-scale evaluations, which demonstrate that current models exhibit significant failures stemming from unwanted competence (e.g., generating harmful advice when prompted) alongside traditional errors \cite{zeng_air-bench_2024}.

\textbf{Aspect interaction analysis.} Risks often emerge from unexpected combinations of system aspects. The framework guides assessors in examining how pairs of AI system aspects can interact and create risks beyond those identified when analyzing aspects in isolation. Building on established systems theory principles, particularly the study of emergent properties, this analysis recognizes that component interactions often produce effects greater than the sum of individual parts. 
For each aspect pair (e.g., capabilities and domain knowledge), assessors evaluate how they could combine to:

\begin{itemize}
    \item Enable new risk pathways not possible with either aspect alone;
    \item Amplify existing risks through synergistic effects;
    \item Transform risks qualitatively through novel interaction or combination patterns.
\end{itemize}

The analysis of higher-order interactions between three or more aspects presents increasing combinatorial complexity but can reveal critical risk pathways not visible in pairwise analysis. 
For instance, the combination of advanced logical reasoning (capability), detailed cybersecurity knowledge (domain knowledge), and privileged system access (affordance) could enable sophisticated attack planning that would not be possible with any subset of these aspects. This interaction analysis also has potential for sensitivity testing to identify which aspect combinations produce significant risk amplification, helping prioritize evaluation efforts. These higher-order risks are most appropriately considered via cataloging known and ideated threats rather than exhaustive iteration over the space.

\subsection{Risk Pathway Modeling} 

\label{risk-pathway-modeling}

Risk pathway modeling traces how aspects of AI systems can lead to real-world harms through sequences of causal steps.  It bridges the gap between technical system assessment and broader societal risk assessment by tracing directed causal chains from source aspects (e.g., a specific capability) through intermediate states (which may be technical, environmental, or social) to terminal aspects (which include societal impacts but can also encompass critical infrastructure system failures or compromise) \cite{mallah_evaluating_2025}. The subsequent parts of this section will detail the analytical techniques and tools used to model these pathways. These include societal threat landscape analysis for mapping potential harm pathways; prospective assessment for reasoning about novel failure modes; propagation modeling for characterizing risk transmission and amplification; and focused aggregation for mapping scenarios to higher-level risk categories.

Risk pathways can manifest through both direct and systemic effects. Direct pathways involve rapid propagation of harms---for instance, the exploitation of a security vulnerability leading to immediate system compromise \cite{hm_government_national_2023}. Systemic pathways involve complex interconnected effects that can fundamentally alter societal structures, such as how AI-generated misinformation can erode trust in institutions and degrade collective decision-making capabilities. Understanding both types is crucial for comprehensive risk assessment. While direct harms might require immediate response, systemic risks can transform fundamental societal structures and capabilities in ways that may be harder to detect and address.
\vspace{5pt}

\begin{figure}[h]
    \centering
    \hspace{0.02\textwidth}  
    \includegraphics[width=0.9\textwidth]{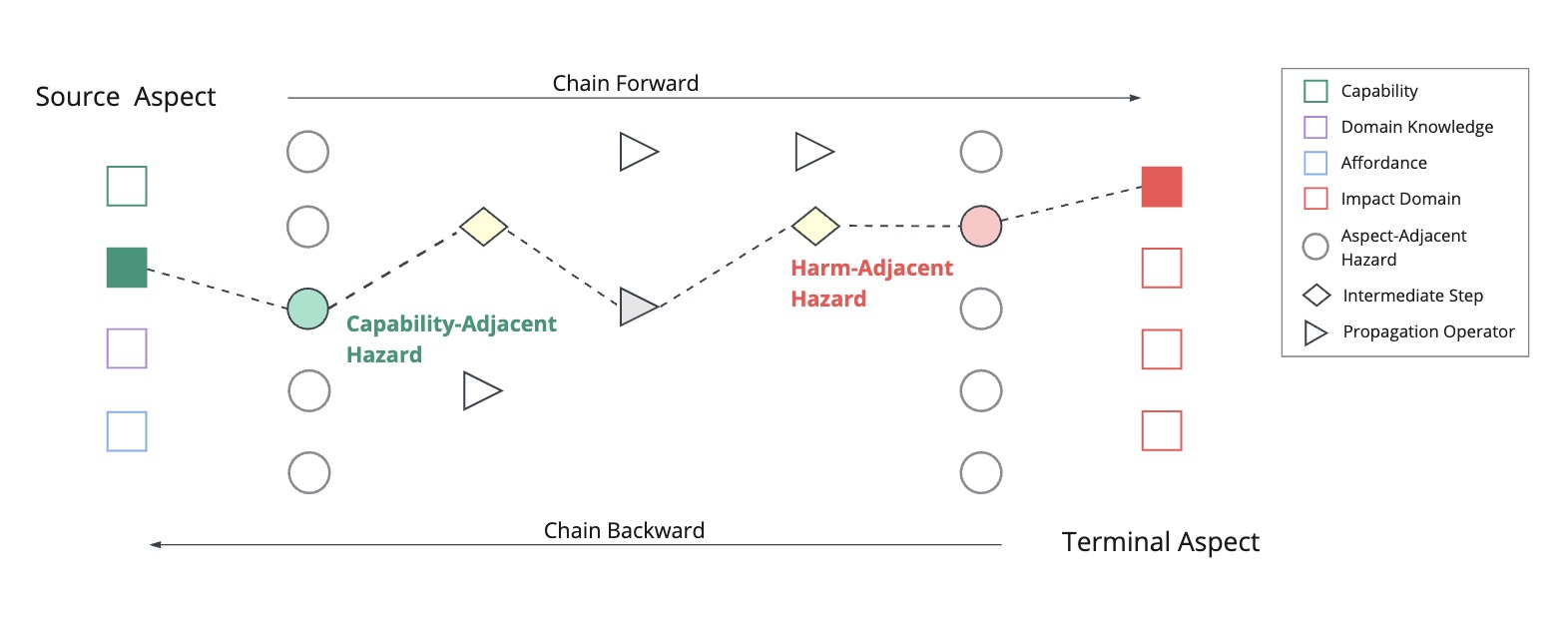}
    \hspace{0.02\textwidth}
    \vspace{5pt}
    \caption{Causal chain illustrating a risk pathway that can be analyzed forward from source aspects (capabilities, domain knowledge, affordances) or backward from terminal aspects (impact domains). Each intermediate step (diamond) represents a stage in the causal sequence of the pathway. Propagation operators (arrow) represent the mechanisms governing risk transmission and transformation between states.}
    \label{pathway}
\end{figure}

Risk pathways consist of six fundamental elements:

\begin{enumerate}[itemsep=3pt]
    \item \textbf{Source aspects.} Source capabilities, domain knowledge, or affordances of the AI system that could initiate a risk pathway, and have the potential to cause harm. \\[0.5\baselineskip]
    \textit{Examples: Integrative cognitive orchestration, coding knowledge, biochemistry knowledge, unfettered Internet access.}
    \item \textbf{Source aspect-adjacent hazards.} 
    The specific hazards that emerge directly or causally soon after from source aspects of the AI system and are the initial points where system characteristics could enable or trigger harm pathways.   
    \\[0.5\baselineskip]
    \textit{Examples: Circumvention of safety guidelines, bypass of security controls, weaponization of domain expertise, exploitation of system privileges.} 
    \item \textbf{Intermediate steps.} 
    States through which risks propagate, defining the sequence of transitions of risks from source to impact. \\[0.5\baselineskip]
    \textit{Examples: Unauthorized commands issued to industrial control systems, anomalous account activity bypasses automated security alerts, lab synthesizes a novel DNA sequence from AI-generated specifications without safety review, automated audit systems miss high-volume illicit financial transactions.}
    \item \textbf{Propagation operators.} Mechanisms that characterize how risks transmit and transform (including, e.g. amplification) between or during the pathway steps as those risks percolate through societal systems, aiding mapping of how risks cascade into broader impacts. \\[0.5\baselineskip]
    \textit{Examples: Adversarial exploitation, targeted misuse, accumulation, compounding.}
    \item \textbf{Terminal aspect-adjacent hazards.} 
    Vulnerabilities through which risks manifest as concrete harms to societal systems, representing the penultimate stage before terminal aspects.
    \\[0.5\baselineskip]
    \textit{Examples:
    Infiltration of power grid controls, compromise of the emergency service authentication chain, breach of biosecurity containment, exploit of financial system transaction verifications.}
    \item \textbf{Terminal aspects.} Domains that are negatively impacted or impinged upon, where harms ultimately manifest, and the endpoints of risk pathways. \\[0.5\baselineskip]
    \textit{Examples: Societal infrastructure \& institutions, individuals' bodily structure, ecosystem processes \& life cycles, individuals' economic \& opportunities.}
\end{enumerate}

The nature, number, and granularity of \textit{intermediate steps} (3) and the specific \textit{propagation operators} (4) employed will vary depending on the pathway's characteristics. For instance, pathways leading to direct technical harms might feature fewer, more technically-focused intermediate steps and propagation mechanisms compared to those resulting in complex systemic effects.

Terminal aspects represent the endpoints of risk pathways---impact domains (TL2) within which actual \textit{harms}, negative outcomes, ultimately manifest through terminal aspect-adjacent hazards (TL4s). For instance, within the terminal aspect (TL2) of Societal Infrastructure \& Institutions, a terminal aspect-adjacent hazard (TL4) could be ``AI-induced critical failure of the power grid''. If this hazard occurs, the \textit{harm} would be widespread societal disruption from prolonged power outages. Similarly, a terminal aspect-adjacent hazard (TL4) like ``AI-exacerbated or caused failure of emergency response'' within the terminal aspect (TL2) of Individuals’ Bodily Structure, could lead to the \textit{harm} of preventable deaths. A terminal aspect-adjacent hazard (TL4) like ``uncontrolled proliferation of an AI-optimized microbe disrupts nutrient cycle'' within the terminal aspect (TL2) of Ecosystem Processes \& Life Cycles, could result in the \textit{harm} of severe ecological damage, as the microbe---designed for agricultural benefit---aggressively outcompetes native organisms and disrupts ecologically foundational nutrient cycles. Finally, a terminal aspect-adjacent hazard (TL4) such as ``AI-driven destabilization of financial markets'' within the terminal aspect (TL2) of Individuals’ Economic \& Opportunities could precipitate the \textit{harm} of systemic financial collapse and widespread economic hardship.

The framework employs two complementary analytical approaches to ensure sufficient coverage of potential risk pathways:

\begin{enumerate}[itemsep=3pt]

    \item \textbf{Forward chaining (source to terminal).} Charts out multi-step processes leading to terminal harms. Aided by analytical tools such as event trees and fishbone diagrams.\\[0.5\baselineskip]
    \textit{Example: Advanced reasoning capability $\rightarrow$ exploitation of cybersecurity vulnerability $\rightarrow$ critical infrastructure disruption $\rightarrow$ societal harm}

    \item \textbf{Backward chaining (terminal to source).} Begins with potential harms (both known and newly identified) and works backwards with structured reasoning to identify their enabling and contributing aspects. Aided by analytical methods such as fault trees and root cause analysis.\\[0.5\baselineskip]
    \textit{Example: Mass societal manipulation $\leftarrow$ prerequisite: advanced psychological modeling $\leftarrow$ source: sophisticated reasoning + human behavior knowledge}
\end{enumerate}

The two approaches offer complementary strengths, enriching risk assessment by providing distinct analytical lenses. Forward chaining explores potential risk pathways, using any available data to model how initiating aspects may lead to harm, while backward chaining, grounded in real-world harm cases, ensures the analysis remains connected to tangible outcomes. Together, they help identify non-obvious pathways and assess the completeness of forward analyses, creating a more robust framework for exploring the societal threat landscape in its full complexity, from individual threats to potentially interconnected clusters of pathways.

Figure \ref{pathway} illustrates this conceptual approach, showing a causal chain that can be analyzed either forward from source aspects (capabilities, domain knowledge, affordances) or backward from terminal aspects (impact domains). To illustrate how these elements form a risk pathway, consider a targeted knowledge base poisoning failure mode (as described in \origcite{bryan_taxonomy_2025}), such as one that might affect a retrieval-augmented generation (RAG) system. The \textit{source aspect} (1) could be the Autonomous Data Management (TL2) capability, specifically its operation allowing the ingestion of external documents into the RAG knowledge base. This creates a \textit{source aspect-adjacent hazard} (2): a malicious actor could introduce a carefully crafted document containing misleading data or embedded instructions designed to corrupt the AI's accessible information. An \textit{intermediate step} (3) occurs when this malicious document is processed and integrated by the AI. The compromise then \textit{propagates} (4) when the corrupted information is recalled by the agent during routine query processing, prompting it to execute an unintended action. This action, such as forwarding sensitive data to an unauthorized party, constitutes the \textit{terminal aspect-adjacent hazard} (5). Ultimately, this results in a harm, like a confidentiality breach, manifesting within the \textit{terminal aspect }(6) of Individuals’ Privacy \& Security (TL2).

The framework employs several key analytical tools and techniques in risk pathway modeling. Among these are societal threat landscape analysis for mapping end-to-end pathways through which AI systems could harm society and its supporting biosphere, prospective risk analysis for reasoning about novel failure modes, propagation operators for characterizing how risks transmit through systems, and focused aggregation for alternative mappings of risk scenarios to higher-level risk categories. Each of these components provides distinct capabilities for understanding and evaluating risk pathways.  For modeling interconnected events relevant to multiple risk pathways, especially those involving conditional dependencies, drawing from structured qualitative and quantitative arguments like those found in safety cases (see Section \ref{practical-utility}), more formal modeling techniques such as Bayesian networks can be employed, which offer a powerful means to represent clusters of interrelated risk pathways and their probabilistic relationships (further discussed in Sections \ref{practical-utility} and \ref{future-directions}).

\textbf{Societal threat landscape analysis.} The risks posed by AI systems to society extend far beyond direct technical failures or misuse, propagating through intricate webs of societal dependencies in ways that traditional component- or model-focused assessments invariably fail to capture systematically. The societal threat landscape---the set of pathways through which AI systems could harm society and its supporting biosphere \cite{mallah_evaluating_2025}---provides a conceptual foundation for systematically analyzing these broader impacts\footnote{Source aspects can be conceptualized as points on an ``outer societal vulnerability surface,'' where AI initially introduces societal risk. Terminal aspects represent points on an ``inner societal vulnerability surface,'' where harms impact core societal vulnerabilities. The risk pathways connect these surfaces, modeling the propagation of AI-driven threats between them.}. This landscape encompasses vulnerabilities and points of interface where AI capabilities, domain knowledge, and affordances may generate both direct effects and cascade effects across interconnected societal structures.

This conceptual advance reframes risk assessment by shifting the focus from individual capabilities or propensities to the broader systems they interact with, enabling systematic exploration of potential harm pathways. The framework operationalizes this through the aspect-oriented taxonomy (see Section \ref{aspect-oriented-hazard-analysis}) that maps both source aspects (capabilities, domain knowledge, affordances) and terminal aspects (impact domains) where harms manifest. This structured decomposition supports identification of aspect-adjacent hazards—potential harms emerging from specific AI system capabilities or propensities as they interact with societal systems.

The societal threat landscape guides assessment methodology by informing systematic sampling of the hazard space and by providing a principled basis for identifying which combinations of system aspects warrant evaluation. The landscape thus forms the qualitative foundation upon which quantitative risk assessment, including the assignment of severity and likelihood to these identified pathways, is subsequently built, thereby developing the risk landscape.

\textbf{Prospective risk analysis.} Assessing risks from AI systems requires imagining and reasoning about hazards that haven't yet manifested, rather than relying solely on historical patterns. Traditional probabilistic risk analysis extrapolates from known failure modes within bounded systems---a rocket may fail catastrophically, but its impacts remain constrained within well-understood limits. 
Advanced AI systems, by contrast, can generate novel failure modes that transform the very context in which they operate, potentially leading to unbounded harms.

Current evaluation methods typically fail to detect behaviors that only emerge at scale \cite{jones_forecasting_2025}. Waiting for empirical evidence before acting on potential AI risks creates a systematic blind spot in governance, thereby leading to the neglect of risks posed by AI systems \cite{casper_pitfalls_2025, chan_evaluating_2024}. This challenge necessitates a forward-looking analysis approach.

The framework addresses this challenge by combining multiple analytical approaches to build reasoned models of potential risk pathways, even where no historical precedent exists. The framework integrates theoretical capability analysis, empirical testing data, and systematic exploration of potential failure modes to construct assessor-justified assessments under documentable uncertainty. These analytical techniques support evidence-informed threat modeling that evolves alongside advancing capabilities, allowing assessment across varying levels of evidence empiricism while maintaining analytical consistency. The analysis comprises three key principles:

\begin{itemize}
    \setlength\itemsep{5pt}
    \item \textbf{Systematic exploration.} Structured approaches for identifying novel failure modes and interaction effects, and systematically searching for what might have been missed.
    \\[0.5\baselineskip]
    \textit{Examples: Red teaming results, adversarial testing, whitebox counterfactual analysis,  emergence studies.} 
    \item \textbf{Extrapolative analysis.} Using available information to project capability trajectories and identify potential threshold effects by projecting forward from what we know.  
    \\[0.5\baselineskip]
    \textit{Examples: Capability scaling laws, emergence pattern analysis, transition indicator analysis,  trend forecasting, causal models.}
    \item \textbf{Evidence integration.} Combining multiple sources of theoretical and empirical evidence to form prospective assessments.
    \\[0.5\baselineskip]
    \textit{Examples: Bayesian hierarchical modeling, mixed methods, structured expert judgment protocols, triangulation, systems models, knowledge graphs, model ensembling.} 
\end{itemize}

In prospective risk analysis, assessors typically employ complementary analytical tools. For example, \textit{capability scaling analysis} helps identify threshold effects where quantitative improvements could enable qualitatively different risks; \textit{multi-agent interaction studies} examine how novel behaviors might emerge in multipolar environments; \textit{model organisms of misalignment} provide controlled experimental settings to analyze potential failure modes; and \textit{whitebox testing} examines the internal workings and decision processes of systems to probe the boundaries of system behavior (see Appendix \ref{app:prospective_methods} for additional techniques). 
These techniques allow assessors to infer significant technical and sociotechnical follow-on possibilities---from novel capabilities emerging from system interactions to cascading effects through social and institutional systems. 
Together, these and other analytical techniques help explore an unbounded risk landscape, providing inputs for risk pathway modeling and mapping the evolving societal threat landscape through technical, behavioral, and systemic indicators of emerging risks.

This analysis necessarily acknowledges and embraces that AI risk assessment must reason about risks even where direct empirical evidence is not yet available. 
Rather than relying solely on demonstrated failures and harms \cite{bommasani_path_2024}, it employs analytical techniques for inferring potential developments, uses, interactions, and downstream effects---such as feedback loop mapping \cite{malinowski_feedback_2019}, cascade effect modeling \cite{zuccaro_theoretical_2018}, and coordination failure mapping \cite{basnett_coordination_2014} 
alongside more technically grounded hybrid analytical-empirical techniques such as latent adversarial training \cite{che_model_2025}, thought flow tracing \cite{lindsey_biology_2025}, control flow tracing \cite{montagu_trace-based_2021}, and robustness regime mapping \cite{anderson_robust_2024}. 
These analytical tools help characterize system behaviors and potential failure modes, informing the analysis of how technical capabilities interact with social systems and institutional responses, revealing potential pathways from current trends to novel risks. 
The analysis begins with clear signals in bounded domains---such as specific capability jumps or bounded system behaviors---and methodically expands the analysis to related domains and interaction effects. 
By combining these analytical projections with theoretical models and some empirical testing, assessors can build justified assessments of unprecedented risks. 
This ``epistemic bootstrapping'' approach allows assessors to begin mapping the risk landscape by starting with confidently known information, then systematically building toward less certain domains. It leverages limited but reliable knowledge to construct justified assessments about unprecedented risks, creating a bridge from well-understood risks to reasonably anticipated but previously unobserved hazards.

\textbf{Propagation operators.} Direct risks from AI systems rarely remain contained---they amplify and transform as they propagate through societal systems and cause harm, often in ways that are difficult to predict from just initial testing. 
The propagation operator analysis provides assessors with a categorized set of transmission mechanisms for analyzing how risks evolve between the steps defined in the risk pathway model. Rather than treating risks as isolated technical failures, these operators characterize specific ways that initial and intermediate risks can transform and cascade into broader impacts, using both technical and sociotechnical propagation mechanisms. A complete categorized set of operators and their descriptions is provided in \cref{app: Pathway}.

To support systematic analysis of these complex transformations, these operators enable several key analytical approaches:
\begin{itemize}
    \item Generating risk pathway variants by applying different operators to existing pathways; 
    \item Identifying novel risk pathways by applying operators to aspect-adjacent hazards; 
    \item Analyzing cascading sequences where multiple operators amplify effects; 
    \item Detecting critical thresholds where operator effects suddenly intensify; 
    \item Tracing cross-domain propagation of harms through different operators; 
    \item Revealing dependencies between risk pathways through shared operators; 
    \item Mapping temporal evolution of risks through operator sequences. 
\end{itemize}

Applying these analytical approaches effectively requires assessing the characteristics and impact of the specific propagation operators involved in each risk pathway step. The analysis of these operators---assessing their effect on risk transmission, transformation, and amplification within sociotechnical systems---can be approached at different levels of rigor, primarily dictated by the availability of relevant data and sociotechnical expertise time to produce models for the specific pathway step:

\begin{itemize}
    \item Quantitative analysis: Uses specific metrics or validated models (e.g., network, economic, agent-based simulations) to directly estimate operator effects. Often challenging due to distal societal effects.
    \item Semi-quantitative analysis: The most common approach for societal steps. Employs structured expert judgment, grounded in available evidence (e.g., historical analogues, social science, related events, proxies, system characteristics). Enables reasoned estimation when quantification is infeasible.
    \item Qualitative analysis: Essential for exploring novel or highly uncertain pathways. Focuses on identifying causal links and characterizing interaction dynamics without assigning numerical estimates.
\end{itemize}

Regardless of the chosen analysis level, the framework requires documenting the method, the supporting evidence (or lack thereof), and the reasoning behind the assessment for each significant propagation step.

The following two examples illustrate the types of multi-step risk transmissions and transformations that can be analyzed using propagation operators:

First, risks from an AI system's classification behaviors can propagate through \textit{skew} in automated decisions, where systematic biases in healthcare diagnoses accumulate through both periodic \textit{accrual}, i.e. small harms adding up to a large harm, with the likelihood of a larger harm increasing with mass system use, as well as other systems adopting similar models and cause risk to spread through \textit{correlation} across healthcare providers, e.g., a given group becomes systematically disadvantaged in some new way. These effects compound through \textit{entrainment} as medical practices adapt to rely on these systems, ultimately manifesting as \textit{public health effects} across vulnerable populations.

Second, risks from an AI system's code generation capabilities might propagate from \textit{untargeted misuse} by unwitting and careless developers to \textit{automated exploitation} of further vulnerabilities through self-replicating scripts, leading with some iterations by the system to \textit{correlation} of security risks across critical infrastructure, until relevant \textit{information asymmetry} from external (e.g. law enforcement) opacity into this use eventually enables coordinated attacks that produce systemic \textit{economic effects}.

In threat model development, propagation operators add to the risk pathway modeling framework by providing a library of semi-structured mechanisms to model how risks evolve between each step of a pathway. 
By explicitly considering how risks propagate through each mechanism type, assessors can better identify potential cascade effects, ideate risk pathways for construction, and prevent some analytical blind spots. This helps move beyond simple linear scenarios to capture complex, multi-dimensional risk pathways that might otherwise be missed.

\textbf{Focused aggregation.} Reducing complex AI risk assessment results to a single system-level risk metric loses critical information about how different risks manifest in society and obscures differences between risk types and their interactions. This limitation highlights the need for more systematic methods capable of mapping specific technical risk findings onto broader dimensions of impact. 
To meet this need for more meaningful representation, the PRA for AI framework introduces focused aggregations to represent and analyze collections of risk scenarios through structured mappings to higher-level risk categories. 
This approach enables nuanced understanding of how different aspects of AI systems contribute to distinct categories of societal risk. 
The value of such structured mappings is increasingly acknowledged, reflected in related work connecting these risks to regulatory and societal contexts \cite{eisenberg_unified_2025}.

The core methodological feature is that individual risk scenarios can be mapped to categories representing respective dimensions of risk through assessed relationships. The assessor can map each risk pathway with particular risk categories, such as how hacking scenarios relate to critical infrastructure failure risk, or how disinformation pathways map to governance breakdown risks.

The methodology supports multiple categorization schemes while retaining information about the underlying causal relationships. This enables assessment data to be analyzed through different lenses relevant to various stakeholder needs---upstream assessment stakeholders may adopt alternative categorization schemes that better serve their specific evaluation needs, such as internal development targets, voluntary safety commitments, or regulatory requirements. Because these schemes can be applied consistently, they provide a standardization basis for comparative absolute risk assessment by risk category. This allows stakeholders not only to track assessment coverage for a single system, but also to meaningfully harmonize and contrast the risk profiles of different AI systems to better understand their relative safety characteristics.

Focused aggregation helps bridge technical and governance needs in a few different ways: enabling integration with existing risk aggregation frameworks; providing social scientists with metrics meaningfully grounded to societal risks; supporting structured comparison of risk profiles across different AI systems; and providing coverage tracking across risk categories to help stakeholders ensure all relevant risk categories have been evaluated per scenario. As these focused aggregation mappings become more objective and generalizable, they will increasingly support structured analysis of how AI system properties contribute to various dimensions of risk.

\subsection{Uncertainty Management} 
\label{uncertainty-management}

Estimating probabilities and magnitudes of impact can be challenging even for experienced assessors in mature fields. Instead of attempting to estimate probabilities for an entire complex scenario at once, a fundamental principle of the framework's uncertainty management is the decomposition of complex scenarios into discrete pathway components. This decomposition breaks risk scenarios into manageable chunks that can be posed as specific, quantitatively modelable questions. PRA for AI handles this estimation challenge with a combination of techniques that are industry standard practice elsewhere and tools adapted for the purpose. The challenges of AI risk assessment demand that the framework enable assessors to:

\begin{itemize}
    \item Construct strong initial threat models by thorough sampling of the hazard space; 
    \item Trace the causal steps that turn an initial threat into a harmful outcome; 
    \item Generate outputs amenable to analysis of interactions, propagations, and sensitivities;  
    \item Make reasoned severity and likelihood estimates for each step; 
    \item Document the evidence, reasoning, and assumptions used in the assessment. 
    \item Reconcile divergent estimates via structured protocols when assessing as a team.
\end{itemize}

Consequently, the methodology provides: 
\begin{itemize}
    \item Classification heuristics containing intuition pumps for threat modeling; 
    \item Scenario decomposition techniques drawn from industry standard assessments elsewhere; 
    \item Optional methods for modeling the interaction and transmission of risks; 
    \item Severity and likelihood intensity rubrics; 
    \item Uncertainty tracing protocols for structured documentation throughout the process. 
    \item Dialectic protocols for assessor comparison and recalibration.
\end{itemize}

In this section, we briefly outline each of the above tools and techniques. 

\textbf{Classification heuristics.} Classification heuristics provide intuition pumps that help assessors calibrate their understanding of AI system aspects by offering examples and progression levels. Capability level progression tables,  implemented in the workbook tool (see Section \ref{framework-implementation}) and excerpted in Appendix \ref{app: capabilities levels}, map the development of general abilities including integrative cognitive orchestration, planning, and strategic optimization. Domain knowledge levels, also implemented in the workbook tool and excerpted in Appendix \ref{app: domain knowledge levels}, characterize the progression of domain-specific capabilities, knowledge, agency, and reasoning patterns within high-risk areas, from adversarial reasoning in cybersecurity to mechanistic understanding of biological systems.

The Risk Detail Table, a component of the workbook tool (see Section \ref{framework-implementation}) excerpted in Appendix \ref{app: Detail Table} provides plausible scenarios for each aspect at each severity level. Following the competence-incompetence analysis framework (discussed in Section \ref{aspect-oriented-hazard-analysis}), these illustrative examples are further subdivided based on whether they primarily emerge from competence or incompetence of the AI system being assessed. This dual analysis is particularly important at jagged capability boundaries where systems may exhibit sophisticated domain-specific reasoning while lacking crucial competencies relevant to safety.

\textbf{Scenario decomposition techniques.} PRA for AI adapts established risk assessment methods to decompose uncertainties and assess them sequentially along the risk pathway 
(detailed in Section \ref{risk-pathway-modeling}). 
Event trees map forward towards a plausible chain of events, while fault trees work backward to identify paths leading to harms. 
For a list of example analytical techniques that can be employed by assessors during decomposition, see Appendix \ref{app:prospective_methods}. 
These complementary approaches help make complex AI risk scenarios more tractable by decomposing them into analyzable components.

\textbf{Optional methods.} For more in-depth analysis of risk interactions and transmission paths, the framework provides tools including Propagation Operators and the Aspect Interactions Matrix, which are implemented as structured templates in the workbook tool. 
These tools are further discussed in Section \ref{assessmentprocess} and Appendix \ref{app: Pathway}.

\textbf{Intensity rubrics.} To model the complexity of potential AI harms, the framework adapts established PRA techniques for interval-based estimation \cite{apostolakis_concept_1990, shortridge_risk_2017} by introducing two types of intensity rubrics. These provide structured definitions and reference examples to guide assessment, and are operationalized as standardized scales in the workbook tool. These rubrics employ coarse-grained bands to categorize severity and likelihood, helping to inform the selection of scenario-specific absolute risk estimates. This banding approach is central to enabling reasoned analysis under uncertainty; it makes complex estimations more tractable and helps mitigate the false precision inherent in seeking exact point probabilities for unprecedented events. These rubrics define the following types of levels:

\begin{enumerate}
    \item \textit{Harm Severity Levels (HSL)} categorize the magnitude of potential impacts across multiple societal dimensions (e.g., human casualties, economic damage, environmental damage), enabling structured evaluation against concrete reference points within defined severity bands (see Appendix \ref{app: HSL}). 
    \item \textit{Likelihood Levels (LL)} categorize the probability of occurrence during the  assessment time frame using defined odds bands  (see Appendix \ref{app: LL Table}), enabling consistent estimation across scenarios and assessors, especially in the absence of sufficient historical frequency data, a common situation for novel AI risks. 
\end{enumerate}

\textbf{Uncertainty tracing.} The framework necessitates that assessors explicitly document their reasoning throughout the assessment process, aligning with common practices in risk assessment that utilize formal protocols for documenting theoretic-empiric rationale development \cite{hemming_practical_2018}. Within the workbook tool (detailed in section \ref{framework-implementation}), this requirement is operationalized through the Risk Assessment Entry Log. Irrespective of implementation specifics, the documentation for each assessed risk pathway must capture:
\begin{itemize}
    \item The initiating aspect(s) and key pathway steps;
    \item Key assumptions and their justifications;
    \item Quality and relevance of available evidence;
    \item Known uncertainties and potential biases;
    \item Assigned Harm Severity Level (HSL) and Likelihood Level (LL) estimates for relevant steps;
    \item Assigned societal risk dimension(s) (for focused aggregation);
    \item Sensitivity of results to changes in critical assumptions;
    \item Rationale for propagation operators used or second-order interactions identified.
\end{itemize}

This documentation traces how uncertainties interact and compound throughout the assessment. For proper context, assessors must also record relevant system-level information that captures the AI system's core characteristics, including its architecture, deployment context, intended use cases, and implemented safeguards (see Appendix \ref{app:system_info}). The resulting documentation renders the assessment more transparent and defensible, enables grading or verification of the assessment, and creates a foundation for future reviews, audits, and assessments.

\textbf{Assessor recalibration protocols.} Assessors evaluate available data from multiple sources---such as empirical observations, capability benchmarks, whitebox testing, and theoretical analyses---based on predictive power and relevance to the current threat model. Different assessors may arrive at varying likelihood and severity estimates based on their expertise and assumptions (see Section \ref{limitations}). To reduce extreme variance, the framework helps assessment teams navigate the inherent subjectivity in probabilistic risk estimation through a structured recalibration process adapted from the Delphi technique\footnote{This adaptation provides a structured recalibration process where assessors examine the reasoning behind differing risk estimates. This allows for consideration of risk pathways and interactions that may have been overlooked while selecting the highest reasonable estimates to prevent underestimation of risks.} \cite{wilson_reconciliation_2023, markmann_delphi-based_2013}. Furthermore, when more specific probabilistic models like Bayesian networks are employed for advanced modeling of particular risk pathways (or clusters of them, as noted in Section \ref{risk-pathway-modeling}), the elicitation and justification of prior probabilities for network nodes becomes instrumental. Making these priors explicit provides a structured focal point for assessors to articulate differing beliefs, discuss underlying evidence, and engage in productive disagreement.

\section{The PRA for AI Workbook Tool} 
\label{framework-implementation}

Having established the conceptual framework and key methodological advances in PRA for AI, this section describes how the framework has been implemented in a workbook tool, establishing the protocols and infrastructure for future practical applications. 
The full documentation and assessment materials for this methodology, including the PRA for AI workbook tool and user guide, are available on the \href{https://pra-for-ai.github.io/pra/}{project website}\footnote{The current implementation is a spreadsheet-based workbook tool (v0.9.1-alpha), which is undergoing initial user testing.}.

\begin{figure}[ht]
    \centering
    \hspace{0.03\textwidth}
    \includegraphics[width=0.9\textwidth]{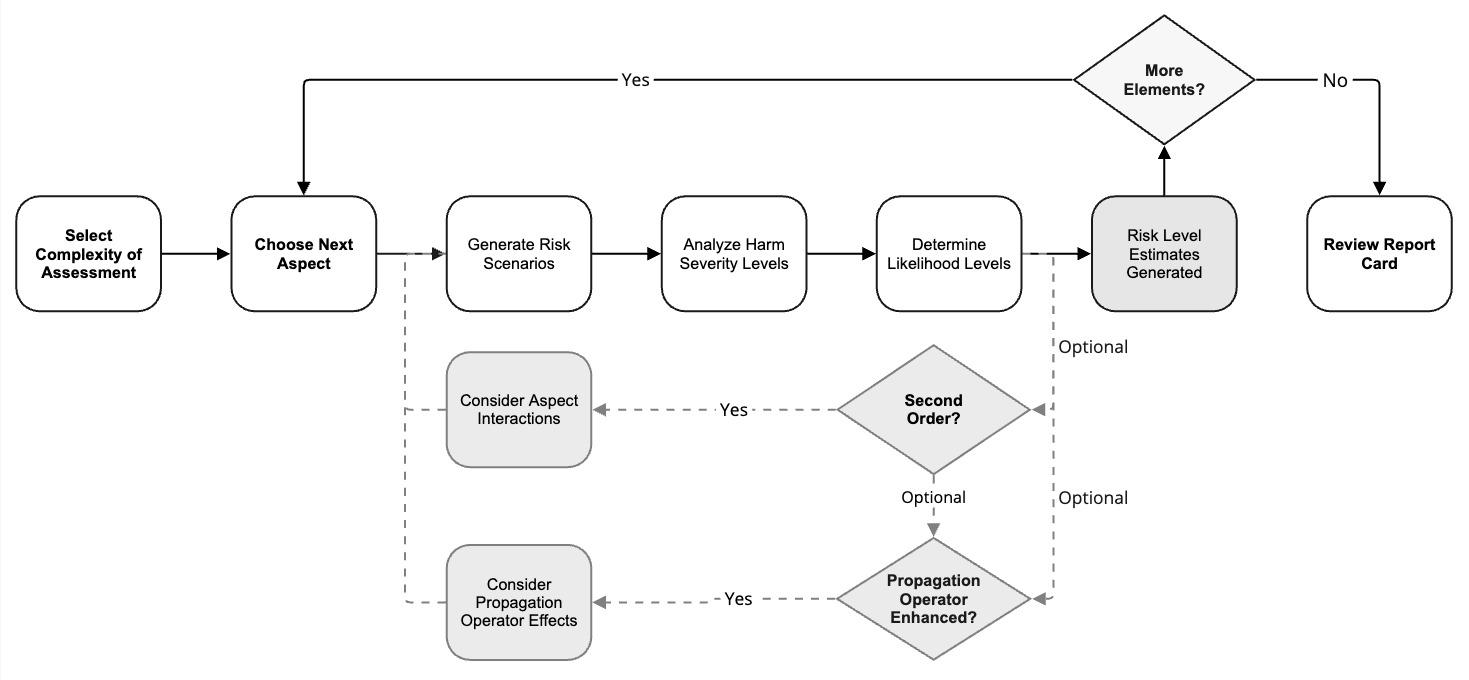}
    \hspace{0.03\textwidth}
    
    \caption{An overview of the risk assessment process flow in the PRA for AI workbook tool.}
    \label{Fig process flow}
\end{figure}	

Figure \ref{Fig process flow} presents the assessment workflow implemented in the PRA for AI workbook tool. The diagram illustrates the iterative process through which assessors evaluate risks across the aspect-oriented taxonomy---from aspect selection through scenario generation, severity analysis, and likelihood determination. The workflow incorporates options for analyzing cross-aspect interactions and risk propagation mechanisms. Upon completion, the tool aggregates risk level estimates into a report card that quantifies the assessed risks of the AI system across various dimensions of analysis.

\subsection{Tools Structure and Usage} 
\label{tools-structure-usage}

The workbook implements the PRA for AI framework and guides assessors through the assessment process. It introduces a Risk Assessment Entry Log in which assessors document system information (see Appendix \ref{app:system_info}), scenarios, and estimates. Additionally, the workbook includes a Risk Detail Table, which acts as the primary tool for scanning the taxonomy during the iterative assessment process, as well as additional tools for risk scenario development, uncertainty management, and risk estimation.

Throughout the assessment, assessors document in the Risk Assessment Entry Log the underlying threat models, assumptions, steps, calculations, uncertainties, and decisions for each risk scenario (see Section \ref{uncertainty-management}). 
The framework also provides classification heuristics to guide and inform threat model development. This includes Plausible Qualifiers---concrete examples of scenarios across harm severity levels per each aspect group, available in the Risk Detail Table, as shown in an excerpt in Appendix \ref{app: Detail Table}. 
These serve as reference points and intuition pumps during threat model development with both competence and incompetence hazards. The framework includes Capability Levels and Domain Knowledge Levels Tables, which characterize progressions of AI system competencies in standardized frameworks across the aspect groups. 
For advanced assessment protocols, the Aspect Interaction Matrix and Propagation Operators guide analysis of interaction (see Section \ref{aspect-oriented-hazard-analysis}) and propagation effects (see Section \ref{risk-pathway-modeling}). The framework guides assessors in decomposing complex scenarios into approachable and tractable steps for analysis before aggregating estimates (see Section \ref{risk-pathway-modeling}). 
The assessment process is supported by intensity rubrics, the Harm Severity Levels Definition Table and the Likelihood Levels Table, that provide standardized scales and deltas from reference examples to inform assessor inputs (see Section \ref{uncertainty-management}). 
The assessment generates three outputs. The Report Card provides aggregated risk estimates from all assessed scenarios. The Tallied Risk Matrix enumerates scenarios by severity and likelihood. The Risk Assessment Output Log preserves the full set of assessed scenarios and their supporting rationales.

The following sections detail the assessment process using the workbook tool: Section \ref{assessmentconfig} outlines the configuration options and Assessment Maturity Levels (AMLs) that determine evaluation scope and depth, Section \ref{assessmentprocess} describes the step-by-step assessment workflow and documentation requirements, and Section \ref{reviewreportcard} explains the generation and structure of assessment outputs including the report card, risk matrix, and documentation logs.

\subsection{Assessment Configuration}
\label{assessmentconfig}

The PRA for AI framework offers configurable Assessment Maturity Levels (AMLs) to address the diverse landscape of AI systems and organizational contexts. This tiered approach is valuable because the resources (e.g., time, expertise, data access) available for risk assessment vary significantly, as does the potential impact and complexity of the AI system under review. AMLs allow an organization to tailor AI risk assessments by selecting a protocol that balances the appropriate analytical depth and coverage against practical constraints, informed by its available expertise, resources, and assessment infrastructure. This adaptability supports effective application across diverse organizational time constraints, regulatory and compliance requirements, and the technical complexity of the AI system being evaluated.

The protocols are represented by a three-digit number code and range from shallow (AML-008 to AML-020) to deep (AML-110 to AML-221). Lower-numbered protocols focus on broad aspect groups of an AI system, while higher-numbered protocols incorporate more complex analyses, such as second-order effects, propagation operators, and assessment at finer taxonomic granularity (moving from aspect groups at taxonomy level 1 to individual aspects at taxonomy level 2). The framework requires selecting the appropriate AML before starting the assessment process. A complete specification of AML protocols and their configurable assessment options (such as aspect group, second-order, and pathway assessment) is provided in Appendix \ref{app: AML}.

The assessment can be performed by a single assessor or a team, though team-based assessments are strongly recommended. Where team assessments yield significantly diverging risk level estimates, a structured recalibration protocol prompts assessors to examine and narrow differences in their underlying reasoning. Each assessment generates an Assessment Type Code\footnote{For instance, if AML-120, version v0.9.1-alpha, and assessor type ``team'' are selected, the Assessment Type Code generated will be: AML-120-v0.9.1-alpha-T.} that records the AML protocol used, the protocol version, and whether it was conducted by an individual or a team. The assessment type is configured through two dropdown menus in the workbook interface---one to select the AML protocol and another to specify team or single assessor mode. After selecting an appropriate AML protocol, assessors document key system information in the Risk Assessment Entry Log (see Appendix \ref{app:system_info}). This standardized tracking provides context for interpreting risk estimates and enables structured comparison between assessments, while ensuring transparency in how each evaluation was conducted.

\subsection{Assessment Process}
\label{assessmentprocess}

The assessment workflow follows an iterative process where for each system aspect, assessors analyze how it could enable or exacerbate hazards. After selecting the appropriate AML protocol, assessors proceed through the following steps:

\textbf{Choose next aspect.} 
Assessors begin by scanning the aspect-oriented taxonomy of AI hazards, and iteratively examining each element of the relevant taxonomy level (see Section \ref{aspect-oriented-hazard-analysis}). The selected AML protocol determines which taxonomy level is used for structuring the analysis. Lower-numbered protocols focus on aspect groups (TL1), while higher-numbered protocols extend to individual aspects (TL2).

\textbf{Generate risk scenarios.} 
The workbook contains templates that guide assessors in developing threat models. Assessors first identify possible risk pathways for each AI system aspect---capabilities, knowledge domains, affordances and impact domains---in the societal threat landscape. When identifying which pathways to analyze in each aspect, assessors use their initial qualitative understanding of potential harm severity, informed by literature on comparables and prominent threat models, as a heuristic to prioritize the most critical threats. They then use bottleneck analysis to consider how each aspect could become the critical constraint or enabler of harm from the AI system (see Section \ref{aspect-oriented-hazard-analysis}) and use the results of that analysis to develop threat models. Assessors then translate these threat models into specific scenarios by constructing directed causal chains that show how harms could materialize. For each identified risk pathway, assessors develop detailed narratives by specifying the initiating event that enables the threat, the intermediate steps through which risks could propagate, and the terminal aspects where harms materialise in societal systems.

As part of this process, risk pathway modeling traces how source aspects could cause a chain of events that ends in harm (see Section \ref{risk-pathway-modeling}). Each link in this chain represents a discrete transformation, such as a change in system state, deployment of a capability, or crossing of a critical threshold. The Capability Levels Table (see Appendix \ref{app: capabilities levels}) and the Domain Knowledge Levels Table (see Appendix \ref{app: domain knowledge levels}) help assessors identify these transitions, particularly where measured improvements in capabilities could enable qualitatively different risk pathways. For scenarios with complex pathways, assessors can employ both forward projection from source aspects and backward analysis from impact domains.

Assessors can employ several analytical techniques while developing detailed scenarios, including fault tree analysis for mapping branching paths from incident to harm, expert elicitation using structured protocols for novel risks, and scenario discovery for surfacing non-obvious hazards. When appropriate, assessors can also draw causal influence diagrams to model decision-making structures, and perform capability scaling assessment to analyze risks from increasing AI capabilities (for more techniques see Appendix \ref{app:prospective_methods}).

While the assessment primarily relies on assessors' direct expertise and real-world understanding of these domains, the workbook provides classification heuristics to aid in ideating scenarios (see Section \ref{uncertainty-management}). Assessors can use the Risk Detail Table, as shown in an excerpt in Appendix \ref{app: Detail Table}, for two sources of guidance. First, they can consult the taxonomy of hazard clusters (TL3) and hazards (TL4) for concrete examples of risk by aspect. Second, Plausible Qualifiers provide escalating examples of harm that allow assessors to reason about impacts through different causal pathways. Plausible Qualifiers are available for both competence scenarios (where exceptional performance creates risks) and incompetence scenarios (where limitations or failures lead to harms) at each severity level, as shown in an excerpt in Appendix \ref{app: Detail Table}).

\textbf{Analyze harm severity levels.}
For each risk scenario, assessors analyze potential impacts semi-quantitatively using Harm Severity Levels (coded HSL-1 through HSL-6). HSLs represent impact magnitude on a defined scale from marginal but non-trivial to globally catastrophic (see Appendix \ref{app: HSL}). Each HSL is defined across dimensions such as human deaths, economic damage, and environmental impact. To estimate the magnitude within these dimensions assessors can draw on a range of relevant evidence, which might include sources such as historical data from analogous events, simulations or consequence modeling specific to the harm type (e.g., economic or epidemiological models, or fault trees), estimates based on system scale (e.g., number of affected users), or structured expert judgment (e.g., Fermi estimates) particularly for novel or intangible impacts (e.g., erosion of trust). When assessing a scenario, assessors identify the relevant dimensions or types of harm, and forecast, calculate, and estimate the potential impact magnitude against each, and ground these estimates with objective criteria or other helpful reference points (such as external benchmarks and concomitant arguments, or structured expert surveys of expectations) where possible. They then translate these impacts into HSLs using the HSL Definition Table, explicitly identifying the range of potential HSL outcomes that could result from the scenario.

\textbf{Determine likelihood levels.}
Assessors estimate the probability of each risk scenario and associated HSL semi-quantitatively using Likelihood Levels (coded LL-0 through LL-8). LLs represent odds ranges decreasing by orders of magnitude, from relatively common events (LL-8, odds 1:1 to 1:10) to vanishingly unlikely ones (LL-0, beyond 1:
10\textsuperscript{12}), with reference examples provided in Appendix \ref{app: LL Table}. These LL estimates correspond to the probability of the specific HSL outcome occurring within the defined assessment time frame (specified during assessment configuration, see Section \ref{assessmentconfig} and Appendix \ref{app:system_info}), with any deviations from this standard timeframe normalized and documented in the Risk Assessment Entry Log.

The workbook guides assessors in decomposing complex risk scenarios into constituent steps. For each step along the directed pathway, assessors estimate probabilities or conditional probabilities of its occurrence. The HSL is determined for the pathway's final outcome, considering any amplification effects. Estimating the overall scenario likelihood---the likelihood of an outcome with a specific HSL occurring within the defined assessment time frame---requires combining the likelihoods of these constituent steps. This often involves analyzing the conditional probability of sequential steps; for example, a simplified scenario probability might be decomposed as P(harmful scenario) = P(capability exists) × P(capability misused | exists) × P(harm occurs | misused). This stepwise analysis helps assessors trace how risks propagate and potentially amplify through the causal chain. Determining the likelihood requires considering dependencies between steps and the frequency of opportunities for those steps over the defined assessment time frame, as the likelihood of one step occurring may influence the likelihood of subsequent steps beyond simple sequence.  When a particular risk scenario could manifest through different contexts or plausible pathways, multiple HSL and LL combinations can be assigned to reflect these variations in impact magnitudes at relevant steps.

The LL estimate for any scenario component or step relies on the diverse evidence available, which may range from theoretical analysis (e.g., observational scaling laws, scalable oversight scaling laws, likely instrumental incentives, human oversight bounds and failure rates) and expert judgment, to empirical frequency data derived from system testing, operational monitoring, red teaming, or other analyses. To improve risk estimates, assessors should ground their likelihood estimates using objective criteria (such as formally provable bounds or relevant empirical data from benchmarks or testing) where possible.

Beyond assigning initial HSL and LL bands based on available evidence and deltas from reference examples, assessors can further structure their reasoning and constrain uncertainty for critical or complex pathway steps by building explicit arguments within the documented rationale. For instance, arguments from inability may justify assigning a lower likelihood if evidence demonstrates the AI system lacks a crucial prerequisite capability for a specific harm pathway. Similarly, arguments identifying critical dependencies or limiting factors within the causal chain can provide a basis for limiting the assessed likelihood of the complete pathway manifesting---this involves analyzing specific steps to determine if they represent particularly challenging prerequisites or pathway bottlenecks, distinct from the aspect-level bottleneck analysis (see Section \ref{aspect-oriented-hazard-analysis}), that constrain the overall probability. Employing such structured reasoning patterns allows assessors to formulate more robust and defensible final risk level estimates.

For each risk assessment entry, assessors record in the Risk Assessment Entry Log their key assumptions, rationale,  potential biases in the analysis, the quality and relevance of their evidence, and any sensitivity of estimates to changes in critical assumptions. 
Within each aspect being evaluated, the criteria for having sampled sufficiently will depend on the aspect itself, the assessment context, and the information available to the assessor. Before proceeding to the next aspect, assessors document their rationale against these criteria.

\textbf{Optional Analysis.} For applicable AML protocols, assessors perform second-order assessment and propagation operator enhanced assessment.

For second-order assessment, assessors work across each matrix column in the Aspect Interaction Matrix---a tool provided to track interaction analysis---and evaluate how the aspect under consideration might interact with other aspects and create risks beyond those identified when analyzing these aspects in isolation. When a meaningful interaction is identified, assessors create a new risk scenario in the Risk Assessment Entry Log, documenting how the interaction could enable or amplify potential harms. As with other scenarios, assessors document their reasoning and the highest estimated HSL and LL for each. This includes explaining the mechanism of interaction and how it changes their estimations.

When performing propagation operator enhanced assessment, assessors evaluate potential risk transmission and amplification pathways. The process involves reviewing the Propagation Operators Table (see Appendix \ref{app: Pathway}) to identify relevant transmission mechanisms for each scenario. Assessors document how each applicable operator could transform or amplify the identified risks, generating additional risk scenarios based on these transmission pathways. They then estimate HSL and LL values for each new scenario, considering the compounded effects. All scenarios are recorded in the Risk Assessment Entry Log with comprehensive documentation of the transmission mechanisms considered and the reasoning behind the resulting risk estimates. For detailed step-by-step guidance on the assessment process, refer to the workbook tool user guide.

\subsection{Assessment Outputs} 
\label{reviewreportcard} 

Once all aspects have been assessed, the workbook tool automatically maps assessors' chosen HSL and LL estimates for each of the completely assessed risk scenarios to risk levels. 
The mapping is defined in the Risk Levels Table (see Appendix \ref{app: RL}).
This table functions as a standardized risk matrix, common in high-reliability fields \cite{iec_iec_2019}, which explicitly maps each possible combination of assessed Harm Severity Level (HSL 1-6) and Likelihood Level (LL 0-8) to a distinct, numerical Risk Level (RL 0-9).

When the assessment is performed by a team of multiple assessors, any contentious aspects are revisited. The framework employs dialectic recalibration to drill down on the estimates made and the detailed rationales provided: assessors first make independent estimates, then significant divergences trigger explicit comparison of underlying assumptions (see Section \ref{uncertainty-management}). While full agreement is not required, final estimates incorporate well-justified perspectives across the assessment team, with highest post-recalibration risk estimates selected to prevent underestimation bias.

\textbf{Risk level estimates generated.} The workbook tool maps each completely assessed scenario by their harm severity and likelihood to a corresponding Risk Level using the standardized Risk Levels Table (see Appendix \ref{app: RL}). These risk levels are calculated distinctly for first-order assessments, first-order propagation operator enhanced assessments, second-order assessments, and second-order propagation operator enhanced assessments. This division enables stakeholders to understand the risk levels of both immediate hazards and more indirect hazards that arise from interactions. When multiple scenarios are generated for the same aspect group, taking the maximum risk level across the scenarios ensures that assessment insights about highest-risk pathways are not diluted by averaging or lower risk alternatives. This ensures critical risk levels remain most visible in the final output, while the separation between assessment types allows stakeholders to understand how different analytical lenses (such as first-order vs. second-order analysis) reveal distinct aspects of the system's risk profile.

\textbf{Review report card.} From the inputs, the framework generates three formal outputs designed to serve different stakeholder needs.

First, the aggregated risk level estimates are presented in the Report Card, which contains the system context documentation, including the assessment date, team composition, system name, version, access level, and documented system-level assumptions, to allow assessors to present the context of the assessment in a clear manner. The risk level summary then displays the aggregated risk levels for each aspect group, with separate columns showing results from each assessment type. A total maximum risk level across all aspect groups and assessment types provides a high-level indicator of the highest risk level assigned in any part of the assessment.

The report card includes focused aggregation results---both a tabular summary and radar visualization to highlight relative risk concentrations---implemented through a standardized mapping interface that supports both default systemic risk dimensions (defined in Appendix \ref{app: focusedaggregation}) and custom categorization schemes defined by assessment requestors. The focused aggregation described in Section \ref{risk-pathway-modeling} allows assessors to map each risk scenario and their estimates to six predefined systemic risk dimensions: social fabric erosion, economic system unraveling, critical infrastructure failure, governance breakdown, environmental breakdown, and public health disintegration.

Second, the Tallied Risk Matrix enumerates all documented scenarios by their assigned harm severity and likelihood levels. This matrix shows the distribution of assessed scenarios across the evaluation space, providing insight into assessment coverage and risk concentrations.

Third, after finalizing the assessment in the report card, assessors generate the Risk Assessment Output Log---a record timestamped with the assessment completion date that captures all completed risk scenarios and their supporting rationales that determined the final risk level estimates. This static record serves as the definitive reference point for the assessment's findings.

The Report Card results should be evaluated in conjunction with the complete Tallied Risk Matrix, Risk Assessment Output Log documentation, and the standard report disclaimer. Furthermore, the results should be considered as one component within a broader ensemble of risk evaluation methodologies, including other quantitative and qualitative approaches. 
Section \ref{discussion} discusses how the framework advances AI risk assessment practice.

\section{Discussion}
\label{discussion}
Building on the methodological foundation (see Section \ref{framework}) and the practical implementation (see Section \ref{framework-implementation}) of the framework, we now examine the framework's contributions to risk assessment practice, discuss its practical applications, analyze its limitations, and identify directions for future development.

\subsection{Methodological Contributions to AI Risk Assessment Practice}

\textbf{Advancement beyond current risk assessment approaches.}
The PRA for AI framework advances the practice of assessing risks from advanced AI systems by  introducing specific methods and tools designed to help overcome limitations in current approaches regarding hazard identification, causal pathway analysis, and aggregate quantification. These contributions, which integrate established PRA techniques with methodological advances tailored for AI’s specific characteristics, provide assessors with enhanced tools for more systematic, comprehensive, and defensible evaluations.

First, the framework enhances hazard identification through broader coverage and targeted analysis via aspect-oriented hazard analysis (see Section \ref{aspect-oriented-hazard-analysis}). Rather than relying solely on ad-hoc selection, or a narrow focus on commonly cited risks, the framework requires structured  scanning of the hazard space guided by a first-principles taxonomy (capabilities, domain knowledge, affordances, and impact domains). Specific analytical techniques provide further value: Bottleneck analysis shifts assessment effort from brute-force evaluations towards more targeted and iterated identification of critical performance or vulnerability thresholds where qualitatively distinct harms may emerge, helping avoid undirected testing that might overlook high-priority threat models. Competence-incompetence analysis offers a key advancement by mandating a balanced consideration of risks arising from both highly efficacious AI execution yielding harmful outcomes (competence) and from system flaws or limitations preventing good operation (incompetence). This dual lens directly counteracts the ``common myopia'' regarding failure modes (see Section \ref{limitationsofcurrentmethods}) by ensuring a balanced focus, which is particularly vital for systems exhibiting AI's characteristic ``jagged'' capability profiles. Critically, this analysis extends to examining hazardous combinations where capability enables or exacerbates failings---a key source of novel, severe risks frequently overlooked by methods analyzing capabilities. Lastly, aspect interaction analysis provides a structure for investigating combinatorial risks emerging from the interplay among AI system aspects, enabling higher-order risk analysis with broader coverage of potential failure modes than isolated capability evaluations.

Second, the framework introduces risk pathway modeling (see Section \ref{risk-pathway-modeling}) to connect system capabilities and propensities to real-world consequences within their sociotechnical context. Addressing a common disconnect in AI assessment, it employs traceable causal chain analysis (forward and backward chaining) to link source aspects (e.g., capabilities) to terminal impacts (e.g., societal disruption). This provides end-to-end coherence often lacking when technical evaluations remain divorced from impact assessments. Unique contributions include the explicit end-to-end modeling of the societal threat landscape and the use of propagation operators. These tools provide a vocabulary and structure for analyzing how initial technical risks transmit, transform, and amplify as they cross system boundaries and interact with complex societal structures---addressing the recognized but rarely tackled challenge of systemic risk analysis for AI. Furthermore, prospective risk analysis techniques are integrated to enable structured, evidence-informed reasoning about novel or unprecedented failure modes, offering a crucial alternative to reliance solely on historical data or currently demonstrable failures, thereby helping to navigate the "evidence dilemma" in assessing rapidly advancing AI \cite{bengio_international_2025}. Finally, focused aggregation allows mapping granular scenario findings onto higher-level risk categories, designed to yield outputs more meaningful for governance than raw benchmark results or single-threat risk scores.

Third, the framework provides techniques for uncertainty management  tailored to general-purpose and frontier AI assessment (see Section \ref{uncertainty-management}), enhancing consistency and defensibility. Moving beyond unaided or off-the-cuff judgments, it mandates scenario decomposition into more readily modelable pieces for a good balance of credibility and tractability. Standardized classification heuristics, which aid assessor calibration by including tools such as capability level tables and plausible qualifiers, and intensity rubrics (such as HSL and LL tables) aid assessor calibration and support consistent, reasoned semi-quantitative estimation even under uncertainty or data scarcity. The adoption of coarse-grained Likelihood Level (LL) bands adapts traditional PRA practice for AI's novelty, acknowledging inherent uncertainty without sacrificing analytical structure---by being moderately broad bands, a simplified assumption of limited precision takes the place of what could have otherwise been grueling assessor-determined uncertainty figures. Crucially, the framework requires explicit uncertainty tracing: protocols mandate documenting evidence, assumptions, reasoning chains, and sensitivities. While acknowledging that expert judgment remains inherent in evaluating novel risks, this requirement for explicit documentation enhances transparency, facilitating critical review of the underlying assumptions and reasoning, and contributing to a more scrutable and well-grounded assessment process compared to approaches where such rationale capture is less systematic or accessible. This aligns with emerging practices in risk assessment, e.g., evaluations, that require documentation across all analytical stages---from pre-analysis planning through execution---to enable validation of both methodological choices and underlying assumptions \cite{paskov_preliminary_2025}.

The framework’s methodological advances are concretely operationalized into a practical assessment workbook (see Section \ref{framework-implementation}). By providing standardized documentation structures and guided workflows, including integrated templates for specific analyses, it translates the conceptual framework into a concrete instrument. This aims to facilitate consistency and lower the barrier for practitioners, offering a structured path for the complex task of AI PRA. Additionally, the workbook's design, which supports assessments done in a consistent, structured way, provides a foundation for future studies examining hazard coverage and assessment consistency across different contexts and assessor profiles, contingent on assessors' explicit consent and opt-in basis for any data contribution..

\textbf{Practitioner reception and early framework development.}
The framework underwent iterative review with experts from high-consequence risk domains and established risk assessment fields. Their feedback specifically addressing assessment workflow clarity and risk estimate consistency directly informed our revisions of both the conceptual framework and its implementation through the workbook tool.

Reviewers particularly valued the framework's ability to systematically identify and analyze risk pathways that cross traditional assessment domain boundaries 
(see Section \ref{uniquechallenges}), enabling more comprehensive threat modeling. The strongest positive reception focused on three key elements: the structured approach to decomposing complex scenarios, explicit uncertainty documentation protocols, and the bidirectional causal analysis approach. The bidirectional approach was recognized for its operationalization of capabilities and harms---with potential to overcome cognitive biases that typically limit consideration of novel failure modes, particularly for high-risk knowledge domains, where system success at known sets of undesired tasks presents the primary risk vector. The competence-incompetence analysis was highlighted as addressing a critical blind spot in current approaches that often focus primarily on only one of either competence or incompetence based hazards.

Experts also noted how traditional threat modeling approaches , often based on canonical vulnerabilities, could fail to account for shifting distributions in attacker capabilities or emerging misuse cases. The framework demonstrates promise in identifying capability combinations that could bypass traditional risk controls through non-standard pathways, such as when models with seemingly harmless capabilities could be combined to produce greater capacity for harmful outcomes.

Additionally, reviewers identified areas for further development, suggesting, importantly, that the creation of assessor calibration exercises for before conducting assessments would strengthen the approach, along with explicit provision of additional guidance for aspect-specific tooling and more granular best practices for threat modeling. These insights will inform continued refinement of the available assessment components in subsequent versions of the workbook tool.

\subsection{Practical Utility in Risk Assessment}
\label{practical-utility}

\textbf{Integration with existing risk management approaches.} 
The AI risk assessment landscape encompasses both formal regulatory frameworks and voluntary agreements, with our PRA for AI framework designed to complement and operationalize both.
Formal regulatory frameworks include the EU AI Act \cite{eu_eu_2023}, the Framework Act on the Development of Artificial Intelligence \cite{ministry_of_science_and_ict_basic_2025}, and various national consumer protection regulations such as the Colorado Consumer Protections for Artificial Intelligence \cite{rutinel_consumer_2024}. 
These 
establish 
legally binding requirements 
while leaving 
implementation to organizations. 
Complementing these are 
voluntary agreements and standards 
such as the Seoul AI Safety Summit commitments \cite{hm_government_frontier_2024}, the General-Purpose AI Code of Practice \cite{eu_third_2025},  NIST AI Risk Management Framework (RMF) \cite{nist_ai_2022}, IEEE 7010 \cite{schiff_ieee_2020}, ISO/IEC 23894:2023 \cite{iso_isoiec_2023-1}, and ISO/IEC 42001:2023 \cite{iso_isoiec_2023}. 

While these frameworks establish valuable procedures, 
they provide limited tools, methods, and guidance for practical implementation. They often lack detailed approaches for quantitative risk estimation techniques, protocols for uncertainty documentation, and systematic hazard identification methodologies. 
The PRA for AI framework addresses these methodological gaps through our methodological advancements, which can be directly incorporated into existing governance structures.
First, 
organizations implementing these standards frequently encounter difficulties 
in systematically identifying, quantifying, and characterizing novel failure modes unique to AI systems, particularly those that might emerge from complex system interactions. 
Second, without structured risk quantification methods, organizations cannot effectively prioritize their control implementations or demonstrate that their procedures meaningfully reduce risk levels. 
Third, compliance documentation often lacks the detailed risk scenario analysis needed to validate that controls are effectively addressing the most consequential risks.

A few AI developers and service providers \cite{anthropic_anthropic_2025, duffer_aws_2024} have begun implementing select standards, though compliance often means satisfying requirements in siloed, incomplete ways.
Organizations adopting the NIST AI RMF can use our framework to operationalize the Map and Measure functions, while those implementing ISO/IEC standards can utilize the explicit rationale and assumption tracking of our documentation protocols to generate the auditable evidence needed to demonstrate systematic adherence to core risk management processes. 
The framework's explicit documentation of assumptions and probability estimates enhances transparency requirements, creating clear traceability from risk identification through to implemented controls. The framework provides a systematic foundation for fulfilling specific requirements in the third draft of the EU General Purpose Codes of Practice \cite{eu_third_2025}, including more comprehensive Systemic Risk Analysis (II.4), Risk Acceptance Determination (II.5), and Safety and Security Model Reports (II.8, Measure 1). 
The framework also offers practical utility for emerging AI risk management approaches, 
with its structured quantitative estimation tools supporting the detailed modeling and indicator operationalization emphasized in frontier AI risk management frameworks \cite{campos_frontier_2025}. Additionally, its aspect-oriented hazard analysis directly supports the systematic identification of uses, misuses, and impacts prioritized for GPAI and foundation model profiles \cite{barrett_ai_2025}.

\textbf{Harmonizing assessment methods and operationalizing risk concepts.}
The PRA for AI framework functions as an integrative methodology. Its design enables the harmonization of diverse inputs from other assessment methods  (i.e., translating them into a common analytical format for its risk pathway modeling) and operationalization of risk concepts (i.e., make abstract concerns measurable and actionable through the production of well-defined, quantified outputs). This dual capacity helps bridge methodological gaps by allowing its analytical structure to integrate findings from external assessments and its outputs to inform other methods, fostering a more unified end-to-end evaluation process.

First, the framework harmonizes information sources by facilitating the integration of findings from external assessment methods and other evidence sources as structured inputs directly into its risk pathway modeling. Assessors can reference specific results from red teaming exercises (e.g., demonstrated vulnerabilities), safety case arguments (e.g., mitigation effectiveness claims, operational contexts), empirical benchmark testing, or other relevant analyses as explicit evidence justifying specific pathway steps or probability estimates. For instance, suppose an assessor needs to estimate the likelihood of the ``circumvention of safety guidelines'' step within a misuse pathway. If a red teaming result demonstrates a 30\% success rate in bypassing a specific safety filter under certain conditions, the assessor would evaluate this finding based on test conditions, comparable scenarios, overall system context, and relevance to the current pathway step. They would then document this 30\% success rate as partial evidence supporting their likelihood estimate for this step, combining it with estimates for other pathway steps. Similarly, reasoning chains identified by safety cases offer readily transferable inputs. These inputs---such as structured arguments, causal relationships, failure modes, dependencies, claimed conditional probabilities, and control measures---can then be used to directly inform and justify the quantitative estimates for corresponding risk scenarios within the framework. This allows disparate evidence types to contribute formally to a structured PRA for AI. Further structure and harmonization across risk pathways can be achieved by constructing a Bayesian network model that spans the given cluster of related risk pathways.

Second, the PRA for AI framework operationalizes risk by translating abstract threats into concrete, measurable, and actionable outputs. Crucially, these semi-quantitative outputs---specifically HSL and LL estimates for specific scenarios---are always to be accompanied by the documented, transparent, human-interpretable risk scenarios and rationales developed during the PRA for AI process. This requisite pairing of numbers and narrative makes the framework outputs readily used as a kind of evidence in complementary risk assessment and risk management methods, enabling even numerical integration with them. For instance, when conducting the assessment, identified gaps in available risk information can help guide TEVV design, and the assessed identified high-risk pathways from the PRA for AI can refine overarching system threat models, which fan out to all risk management processes. To understand how PRA for AI outputs can be used it's useful to consider how different frameworks answer core safety questions. While PRA for AI focuses on identifying and analyzing potential harm pathways (to answer ``how inherently risky is this system?''), its outputs provide essential evidence for complementary assurance methodologies like safety cases and risk cases \cite{clymer_safety_2024}. Safety cases aim to demonstrate the safety of advanced AI (``why is this system safe?''). Risk cases, originally conceptualized in response to the challenge of proving absolute safety in complex systems \cite{leveson_danger_2018, haddon-cave_nimrod_2009}, focus on demonstrating how identified risks are managed to acceptable levels\footnote{E.g., ALARP (As Low As Reasonably Practicable)---a level of estimated risk after mitigation at which further risk reduction would be disproportionate to the benefit gained.}  (demonstrating ``how unsafe is this system after controls?''). These three approaches complement each other and can integrate numerically as subordinate or enclosing assessments.

Specifically pairing quantitative results with explanatory context can inform and enhance key risk assessment practices, such as:

\begin{itemize}
    \item \textbf{Contextualizing Red Teaming.}  PRA for AI outputs add critical context and help address a measurement challenge for red teaming. While red teaming identifies vulnerabilities, PRA for AI then helps quantify their potential likelihood and impact within modeled pathways, establishing essential context relating to impact that is often missing from raw findings. This enables prioritization and helps to structure thinking about potential real-world impacts, overcoming some limitations of testing models in isolation \cite{ji_how_2025}. Other components of the framework, such as the aspect-oriented taxonomy of AI hazards and classification heuristics, also help red teams scope more thorough tests and characterize capabilities inventories.
    \item \textbf{Informing Safety Cases.} The synergy between PRA for AI and safety cases is bidirectional. PRA for Al's quantified risk scenarios and documented rationales provide evidence to substantiate or challenge claims made within a safety case. For instance, an HSL/LL estimate for a specific misuse scenario can inform the safety case's argument about the tolerability of the risk. Beyond individual risk estimates, the framework's systematic hazard identification can help define the scope of scenarios the safety case must address. PRA for AI outputs also facilitate structuring inability arguments, where capability evaluation results demonstrating a positive lack of requisite capabilities are then used to justify assigning extremely low-probability estimates to dependent risk pathways \cite{aisi_safety_2024} and providing a baseline for dynamic safety cases \cite{carlan_dynamic_2024}. This helps identify when safety arguments require revision considering the further analysis of fuller pathway dynamics or control degradation.

\end{itemize}

This dual capacity---harmonizing inputs from other methods and operationalizing risk constructs by producing directly useful outputs---positions the PRA for AI framework as a crucial bridge between fragmented assessment practices, and enables a more unified end-to-end risk evaluation process for advanced AI.

\textbf{Downstream uses and applications.} 
The PRA for AI framework can be applied broadly to quantify AI risks, providing a structured approach for translating diverse risk information and expert judgments into estimates within defined probability bands that directly inform several key applications. 
Primarily, the risk estimates produced enable organizations to implement targeted mitigation strategies by identifying which specific pathway components contribute most significantly to overall risk levels. 
Integrating with broader organizational processes, the framework's detailed risk pathway modeling can offer specificity to help inform control configurations within unified governance structures \cite{eisenberg_unified_2025}, potentially allowing actions to be tailored more effectively against certain risk mechanisms.
Furthermore, the methodology enhances sensitivity analysis capabilities, allowing assessors to better identify and assess how changes in system capabilities, propensities, and aspect interactions affect risk estimates and determine which factors might have the greatest impact on outcomes.

Beyond informing mitigation strategies, the framework’s standardized metrics, grounded in its HSL/LL rubrics for estimating absolute risk and structured documentation, support crucial oversight functions. They facilitate understanding risks across the entire system lifecycle and enable systematic comparisons of risk profiles between assessments---whether evaluating different AI systems, tracking a single system’s risk profile over time and through version changes, or evaluating the efficacy of mitigations. 
These semi-quantitative measures are also valuable for compliance and governance, supporting the evaluation of AI systems against established intolerable risk thresholds \cite{raman_intolerable_2025} and informing the application of tiered safety approaches by providing the quantified risk levels needed to assign systems to specific tiers 
\cite{future_of_life_institute_safety_2025, caputo_risk_2025}. 
These quantified risk levels function as actionable thresholds (“points of disjuncture”), triggering specific responses like continued monitoring, mandatory mitigations, or deployment halts.

The PRA for AI framework is particularly useful when direct empirical data is limited or insufficient for threat modeling. While benchmark data provides valuable capability measurements, it is often not directly usable in complex risk scenarios.
Expert elicitation is typically used to bridge this gap, informing probability estimates within specific risk pathways based on available benchmark results and other evidence \cite{murray_mapping_2025}. The PRA for AI framework extends and facilitates such elicitation practices, leveraging its methodological emphasis on structured pathways and documented reasoning chains to enable informed probability estimation even for novel risk pathways where direct measurement data may be unavailable.

This ability to structure analysis under uncertainty is especially valuable for high-impact, low-probability events like global catastrophic risks (GCRs).
For GCRs, the framework's value lies less in achieving precise probability estimates and more in providing a structured process for analyzing potential pathways and impacts. 
GCRs are high-consequence events where, applying the principles of expected value, even probability estimates spanning multiple orders of magnitude below 1\% can yield substantial expected harms. 
In such cases, it is the magnitude of potential impact and its reasoned non-trivial plausibility---not the precision of the estimate---that determines the relevance for decision-making \cite{kunreuther_risk_2002}. Uncertainty alone does not justify inaction; on the contrary, the absence of precise data should increase the emphasis on precaution in light of the stakes involved \cite{baum_challenge_2019}.

More broadly, a benefit of the PRA for AI framework is its ability to take risk assessment forward in previously intractable areas by providing a structure for assessors and decision-makers confronting inherently qualitative aspects of threat modeling and LL estimation. It aids threat modeling and sizing by highlighting concepts such as unfortunate decisions, harm size cascades, and increasing maximum harm sizes over time as leverage grows. The process involves first structuring potential risk pathways, often using qualitative reasoning for novel or complex steps, which then provides the necessary foundation for systematic estimation (see Section \ref{assessmentprocess}).  
This structured approach helps counteract the analytical paralysis and additional risks that can occur when waiting for empirical evidence before acting
\cite{casper_pitfalls_2025}, particularly as setting evidentiary standards too high for regulatory and assessment actions can lead to the neglect of significant risks posed by AI systems. 

Finally, PRA for AI findings can contribute to institutional memory by integrating into enterprise-wide knowledge management systems, documenting identified risks. 
This documentation provides a basis for developing risk dashboards that track evolving threats across product lines and deployment contexts.

\textbf{Stakeholder benefits.} 
Different participants in the AI ecosystem can derive specific utility from the PRA for AI framework based on their specific roles and responsibilities.

AI developers can use PRA for AI to systematically evaluate how specific combinations of capabilities and design choices influence system-level risk profiles, facilitating early identification of thresholds where qualitative changes in risk emerge. This supports more informed architecture decisions, training strategies, and development trajectories grounded in safety considerations. Enterprise adopters, in contrast, can assess whether a system aligns with their operational risk tolerance, guiding context-sensitive evaluations of readiness and informing pre-deployment investment decisions.

Risk professionals can integrate quantitative estimates into enterprise risk management systems, enabling more informed decisions about deployment and mitigation strategies while identifying and prioritizing potential harms. Compliance and legal professionals can use the framework to operationalize emerging regulatory expectations by embedding them in structured, repeatable assessment processes. PRA for AI helps clarify the basis for risk-related decisions by requiring well-documented assumptions, scenario justifications, and propagation pathways. This traceability supports internal governance, facilitates regulatory interpretation, and provides a defensible foundation for external audits, reviews, and liability considerations. 

Evaluation organizations can harmonize insights from benchmarks, manual testing, and automated evaluations to produce risk level estimates rather than binary pass/fail assessments. This supports more nuanced understanding of system safety properties and potential failure modes, while advancing the scientific understanding of AI risks.
Insurers and actuaries can use the framework's quantified risk estimates to assess potential liabilities and exposures across deployment contexts, supporting the development of fit-for-purpose insurance products and premium calculations \cite{weil_insuring_2024}. 
Regulatory and advisory entities can move beyond proxies such as training data size or capability-based risk restrictions by mandating standardized risk assessments that enable meaningful safety comparisons across AI systems. This provides a common vocabulary for evaluating compliance and the adequacy of safety measures.

Policymakers can use PRA for AI to evaluate trade-offs between advancing AI capabilities and implementing safety measures by quantifying expected value loss calculations\footnote{Expected value loss calculations enable the quantification of the ``price,'' defined here as the resulting trade-off—expressed in expected value terms---balance between advancing the system, with its potencies and foibles, and the alternative of forgoing it due to risks that current safety measures cannot surely, or adequately, mitigate.}. This enables more proportionate interventions and helps align regulatory decisions with the scale and structure of system-level risks.

The framework's structured documentation and the common vocabulary it enables can help bridge communication gaps between technical, business, and governance stakeholders. Because the workbook preserves all inputs within a standardized structure, it enables the systematic comparison of the system's assessed risk profile at granular levels—from high-level summaries to the specific pathways and aspect-level risks that underlie them—with other systems or its own prior versions when they are assessed using the same methodology. This provides a consistent basis for assessment and comparison of results at any stage of the AI lifecycle, from pre-development, pre-release, and in-service phases. Effective management of this lifecycle, which aligns with established principles \cite{nist_ai_2022, iso_isoiec_2023, iec_iec_2019}, involves teams defining explicit organizational triggers for reassessment—such as major model updates, the discovery of new exploits, or scheduled periodic reviews—and designating an owner responsible for launching the next PRA cycle. An assessment can be prompted by changes in industry-standard triggers such as Key Risk Indicators (KRIs)---metrics that signal increasing risk exposure \cite{nist_risk_2019}. For instance, an organization might track a model’s deceptive capabilities as a KRI, where a sudden jump in those capabilities would trigger a reassessment \cite{campos_frontier_2025}. In a complementary role, the framework's risk pathway modeling can help identify the most critical mitigations, thereby informing the strategic placement of controls and the definition of corresponding Key Control Indicators (KCIs)---metrics that track whether that specific control is performing as expected \cite{nist_risk_2019}. As PRA for AI is applied across diverse AI systems, assessment experiences can be systematically captured to refine aspect-specific methodologies, expand reference scales, and strengthen framework implementation efficacy over time. 

Ultimately, the realization of these practical applications and stakeholder benefits are contingent on stakeholder confidence in the assessment's overall integrity and soundness. Downstream users---whether developers, adopters, or regulators---must perceive the assessment process and outputs as sufficiently credible and trustworthy before relying upon them for critical decisions.

\subsection{Framework Limitations and Implementation Constraints} 
\label{limitations}

When utilizing the PRA for AI framework, several constraints and limitations must be considered and planned for. Assessors face appreciable challenges in probability assessment, including fundamental cognitive constraints in discriminating between different low probabilities, limited research foundations for various classes of low-probability forecasting, validation challenges for rare event predictions, and scarcity of qualified expertise availability. A significant hurdle is the validation of risk estimates, particularly for rare or previously unmanifested AI events, as these inherently lack historical precedent and sufficient empirical data for robust confirmation, making conventional predictive analytics difficult. Assessors may experience anchoring bias on predefined categories or examples, potentially limiting their consideration of unconventional or hard-to-imagine risks. Assessors may of course also share common knowledge gaps across different methodologies, including PRA for AI and safety cases (as discussed in Section \ref{limitationsofcurrentmethods}), concerning very recent findings or complex risk dynamics if such insights are not yet broadly disseminated.

A core challenge lies in the inherent subjectivity when estimating likelihoods and impacts for novel risks, making rigorous validation difficult, though structured protocols and recalibration aim to improve consistency and defensibility. This subjectivity becomes more pronounced when determining how technical capabilities could propagate through interconnected societal systems to create harm—a process requiring understanding and judgment calls about complex societal dynamics and amplification effects. Consequently, it is important to recognize that the absolute risk levels and expected loss amounts that can be derived from PRA for AI, particularly for unprecedented or high-consequence, low-probability events, will inherently possess degrees of imprecision and may not be perfectly reproducible. However, this characteristic does not invalidate the utility of the assessment’s quantified rationales; it embraces the necessary uncertainties involved with forecasting the future.

In this context, the PRA for AI framework aims to bolster construct validity not by pursuing spurious precision, but by systematically structuring reasoning and making assumptions explicit, encouraging cycles of critique and dialogue. The focus is on generating a qualitatively new and actionable understanding of risk through defensible, order-of-magnitude estimations within defined bands (see Section \ref{uncertainty-management}). This approach provides a more robust basis for risk-informed decision-making than relying on purely qualitative statements of “plausibility” or unevaluated intuitions, especially when confronting potentially catastrophic outcomes. Assessor competence, honesty, and independence of incentives is crucial, as contingent or misaligned incentive structures could systematically and significantly degrade the quality and trustworthiness of these determinations.

It is important to note that PRA for AI represents a complementary set of incremental improvements to practice rather than a complete solution to AI risk assessment challenges. Its effectiveness depends heavily on expert judgment and inherits associated biases and limitations. Similar to all current methods, it cannot guarantee safety or completely address unknown unknowns. Instead, it provides a structured framework for making explicit an extended reasoning about risks and documenting of assumptions, while arriving at an assessor-defensible estimate of the total quantified risk in a very complex system.
 
While the framework's configurable Assessment Maturity Levels (AMLs, see Section \ref{assessmentconfig}) provide a mechanism to tailor the assessment's scope and depth, conducting an assessment with the workbook tool requires substantial technical and organizational resources due to the detailed analysis and 
systematic broad scope needed for assessing complex AI risks. 
This resource intensity, particularly when applying deeper AML protocols appropriate for high-stakes systems, can be prohibitive for time-sensitive decisions or organizations with restricted capacity. 
Effective implementation requires specialized expertise across multiple domains and maintaining consistency both within and across assessment teams.
Assessment complexity varies fundamentally between aspects----from cases where structured empirical data is readily available to scenarios requiring extensive novel threat modeling. 
Furthermore, assessment quality can be significantly constrained by restricted access to proprietary system details (e.g., architecture, training methodologies, test data) or by limitations on performing probing analyses like whitebox testing or fine-tuning experiments, particularly for closed-source commercial systems.

\subsection{Future Research Directions}
\label{future-directions}

The framework provides an initial foundation for systematic AI risk assessment, and several promising research directions could further extend its capabilities:

\textbf{Improving uncertainty quantification.} 
Current risk assessment remains contingent on subjective assessor judgment, making it difficult for decision-makers to evaluate the reliability of risk predictions. 
We need better techniques for calibrating expert judgment in these domains and particularly improved methods---such as hindcasting where applicable or tracking accuracy against observable indicators---that can validate predictions against emerging empirical evidence. This should include developing structured protocols for uncertainty propagation that consider the effects of potential capability jumps and emergent behaviors, as well as research into the practicality and utility of methods for representing assessor uncertainty beyond point estimates (such as formal error bars). Most crucially, we need methods for identifying and modeling correlated failure modes across multiple AI systems, as current approaches often focus on single-system analysis, overlooking risks emerging from the complex interactions and collective behaviors within multi-agent AI ecosystems 
\cite{hammond_multi-agent_2025}. 

\textbf{Network modeling of risk pathways.} Current risk assessment methods often analyze harm pathways individually, limiting our ability to model the numerous potential routes through which risks can emerge, combine, and amplify. Network modeling approaches offer potential enhancements for risk pathway analysis. Specific techniques like threat knowledge graphs have shown utility in cybersecurity for uncovering hidden connections \cite{xiang_uncovering_2025}, while hypergraphs are suited for modeling multi-level risk propagation and  higher-order dependencies in AI's sociotechnical ecosystem, potentially revealing complex interactions missed by standard graph analysis \cite{kilian_multiscale_2025}. Building on complex systems approaches to AI risk, applying network and hypergraph models offers the potential for systematic exploration of multiple causal chains simultaneously \cite{kilian_examining_2023}. Bayesian networks are particularly promising for modeling overlapping risk pathways, as they can be used to quantitatively integrate evidence and dependencies from a variety of sources, forming core baselines, usable to explore marginal risk. Future work should focus on developing more straightforward and standardized methodologies and practical tools for constructing Bayesian networks and calibrating nodes’ priors and conditional probabilities from diverse evidence. This would include processes for eliciting, perturbing, validating, and constructively debating Bayesian prior probabilities, which is crucial for improving uncertainty modeling and overall uncertainty management (see Section \ref{uncertainty-management}). Expanding to network models could identify critical vulnerabilities through graph analytics, represent feedback loops, and well abstract risk cascades leading to systemic consequences. Computational techniques such as Monte Carlo methods for risk propagation modeling and generative AI-driven scenario discovery, in combination with human curators, could leverage these network models to systematically analyze compound effects and interaction scenarios. Additionally, network visualizations can serve as valuable communication tools, making complex risk landscapes more analytically tractable and relatable for both researchers and decision-makers.

\textbf{Establishing an AI risk pathway knowledge base.}
Many relevant parties' limited understanding of how capabilities scale and interrelate makes interpreting current evaluation results and predicting future risks challenging, as the science of capability evaluations remains underdeveloped \cite{weidinger_toward_2025}. 
A structured knowledge base that systematically documents AI hazards and risk pathways using our aspect-oriented taxonomy would provide valuable reference. 
The knowledge base could store detailed information about complete and partial risk pathways, drawing from incident databases, risk registers, and assessment outputs \cite{slattery_ai_2024, mcgregor_preventing_2020, hm_government_national_2023}. 
By cataloging detailed descriptions of potential risk pathways, documented interaction effects, the enabling capabilities, reference examples, and context-specific assessment methods, this resource could support more consistent risk evaluation across different operational contexts.

\textbf{Conducting larger holistic case studies.} 
In-depth public reference case studies spanning multiple AI aspects and diverse risk pathways would demonstrate how the framework's components interact in real assessment contexts. The framework and workbook tool should be used in assessments linked from model cards to assess the latest frontier models from leading AI developers \cite{anthropic_introducing_2024, openai_openai_2024, google_google_2025} publicly with necessary but minimal redactions of novel information hazards \cite{bostrom_information_2011}.
Such reference assessments would provide concrete illustrations of the techniques fully applied and worked out.
We plan to publish illustrative case studies of the framework's application 
across different types and scopes of AI systems to show how organizations can operationalize the assessment process in various contexts.

\textbf{Developing standards.} 
Effective AI standards can benefit from normalized risk thresholds to move beyond purely process-focused requirements; the quantification capabilities provided by PRA for AI offer a foundation for establishing such common standards and being able to hold any system, past, present, or future, up to those thresholds. 
The framework could contribute to standards in several critical ways: hazard identification taxonomies, quantitative risk estimation protocols, and criteria for validating the quality of assessment execution. 
Future standards should adapt lessons from assurance level concepts from other safety-critical domains (as discussed in Section \ref{EstablishedtraditionalPRA}) for AI's unique challenges (e.g., emergent behaviors, rapid evolution).
Such standards should establish clear, prescriptive requirements while accommodating new assessment techniques and tooling, ensuring continued relevance and consistency. 

\textbf{Refining risk mapping methodologies.} 
Effective risk governance requires distilling complex technical assessments into actionable insights for diverse stakeholders. The current Focused Aggregation method (see Section \ref{risk-pathway-modeling}) relies heavily on assessor judgment for mapping scenarios to societal risk dimensions. 
Future work should focus on developing more objective and generalizable classification criteria for these mappings. Furthermore, integrating outputs from advanced analytical techniques like network modeling of risk pathways (discussed earlier), which can reveal intricate interactions and dependencies, may generate insights whose complexity necessitates semi-automated support. Such tools, guided by well-defined criteria, could help consistently map complex scenario characteristics onto societal risk dimensions, reducing subjective variation and helping organizations better understand and act on assessment results while maintaining the systematic connection between technical capabilities and societal impacts.

\section{Conclusion}
\label{conclusion}

This paper presents a probabilistic risk assessment framework for AI that applies methodologically grounded, systematic analysis to a field previously characterized by fragmented approaches, selective testing, and implicit assumptions about risk priorities.
Building on established methods from high-reliability industries, we have shown how probabilistic techniques can be meaningfully adapted to evaluate AI systems through a structured, aspect-oriented approach. The framework helps transform previously qualitative and disjointed thinking about AI safety issues into empirically informed reasoned analyses extended through structured argumentation and diagrammatics, which can be methodically compared and evaluated.

Our approach advances AI risk assessment practice through three key advancements: (1) systematic exploration of the hazard space using aspect-oriented analysis, (2) structured quantification methodology that decomposes complex scenarios into analyzable components, and (3) explicit documentation protocols that capture reasoning chains and assumptions. These advances create the foundation for cumulative knowledge building in AI risk assessment, enabling organizations to build upon lessons from previous assessments, identify knowledge gaps, and track evolving risk profiles as capabilities advance.

The PRA for AI framework addresses current risk assessment fragmentation by aiming for broad coverage across diverse risk domains, enabling detailed exploration of causal risk pathways across technical and societal areas, and providing concrete implementation guidance through the workbook tool (see Section \ref{framework-implementation}) and also by the harmonization abilities of the framework. 
This integration makes it possible to assess risk more holistically in its breadth and depth. 
Our approach enables configurable assessment depths through tiered maturity levels for different organizational needs without sacrificing consideration of the societal threat landscape.

Importantly, the framework establishes a common vocabulary for discussing AI risks across different stakeholder groups---from technical developers and safety researchers to policymakers and regulatory bodies---through standardized assessment scales, 
quantified risk levels, and structured documentation protocols. This shared foundation facilitates more productive audits, disagreements, and identification of assessment blind spots.

While implementing PRA for AI faces challenges, including uncertainties about novel risks, non-trivial work in quantifying tail risks and black swan events, and some limits on understanding of how advanced AI capabilities might emerge and interact, these limitations suggest the framework is most valuable when combined with complementary approaches like safety cases, qualitative scenario planning, and red teaming.
We present this framework and the associated workbook tool, available at our \href{https://pra-for-ai.github.io/pra/}{project website}, as a contribution to the nascent field of general-purpose AI risk assessment and welcome engagement from the broader research community to extend, refine, and demonstrate these approaches. This framework represents an initial step towards more comprehensive 
probabilistic risk assessment of advanced AI systems. 

Given the rapid advancement of AI capabilities, we cannot afford to approach risk assessment in an ad hoc manner without methodological reasoning about a wide breadth of risk pathways. 
Within this systematic reasoning, particular focus should be given to identifying and analyzing the most highly consequential pathways.
While adapting proven methods from high-reliability industries will not guarantee safety, it provides a crucial foundation for systematic risk assessment that the field urgently needs.
Our hope is that this work contributes to increased operational and theoretical understanding of AI risk, enables better risk governance, enriches technical risk and remediation research, and supports policy work on developing measurable safety thresholds and safety requirements.

\section*{Acknowledgments}
We extend our sincere gratitude to the many groups and individuals that provided invaluable feedback on the development and refinement of our Probabilistic Risk Assessment for AI framework, including: Anthony Aguirre, Usman Anwar, Peter Barnett, Claire Boine, Mark Brakel, Justin Bullock, Siméon Campos, Duncan Cass-Beggs, Giulio Corsi, Abra Ganz, Ross Gruetzemacher, Olle Häggström, Seán Ó hÉigeartaigh, Geoffrey Irving, Corin Katzke, Kyle A. Kilian, Daniel Kroth, Dan Lahav, Yolanda Lannquist, Matthijs Maas, Nada Madkour, Alexandru Marcoci, Evan R. Murphy, Malcolm Murray, Daniele Palombi, Jai Patel, Toby Pilditch, Krystal Rain, Deepika Raman, Matthias Samwald, Tim Schreier, Everett Smith, Cozmin Ududec, Risto Uuk, Akash Wasil, Sterlin Waters, and Marta Ziosi. All remaining errors are our own. To provide feedback or contribute to the framework's development, please contact the project team at \href{mailto:PRA@carma.org}{PRA@carma.org}.

\hypersetup{urlcolor=black}

\printbibliography

\newpage
\appendix

\renewcommand{\appendixname}{Appendix}
\renewcommand{\thesection}{\Alph{section}}

\titleformat{\section}
    {\normalfont\Large\bfseries}
    {\appendixname\ \thesection:}
    {1em}
    {}

\hypersetup{
    urlcolor=[rgb]{0.0,0.37,0.72}
}

\section{Comparison of Risk Assessment Methods}
\label{app: Comparison Matrix}
\normalsize

Table \ref{tab:interleaved_safety} provides a comparison of AI risk assessment methods across key dimensions relevant to societal threat landscape analysis. 

\begin{table}[H]
\caption{Comparison of risk assessment methods.}
\vspace{5pt}
\label{tab:interleaved_safety}
\fontsize{7.5}{9}\selectfont
\renewcommand{\arraystretch}{1.1}
\setlength{\tabcolsep}{1pt} 
\begin{NiceTabular}{|>{\raggedright\arraybackslash}p{4cm}|
>{\centering\arraybackslash}p{2cm}|
>{\centering\arraybackslash}p{1.8cm}|
>{\centering\arraybackslash}p{1.8cm}|
>{\centering\arraybackslash}p{2cm}|
>{\centering\arraybackslash}p{1.8cm}|}[colortbl-like]
\hline
\RaggedRight\textbf{Assessment Method} & \RaggedRight\textbf{Fine Grain} & \RaggedRight\textbf{Good Proxy to Safety} & \RaggedRight\textbf{Societal Threat Surface Coverage} & \RaggedRight\textbf{Robust to Mitigation Failure} & \RaggedRight\textbf{Threat Surface Guidance} \\ \hline
\rowcolor{white}
Safety Benchmarks (No Holdout) & H & L & M & L & M \\
\rowcolor{gray!10}
Safety Benchmarks (Private Holdout) & H & L & M & L & M \\
\rowcolor{white}
Evals & H & M & L & M & L \\
\rowcolor{gray!10}
Responsible Scaling Policies & L & M & L & L & M \\
\rowcolor{white}
Safety Cases & M & H & M & L & M \\
\rowcolor{gray!10}
Probabilistic Risk Assessment & M & H & H & H & H \\
\rowcolor{white}
Typical Narrow AI Safety Audits & L & L & L & L & L \\
\rowcolor{gray!10}
Deep Bespoke AI Safety Audits & H & H & L & M & L \\
\rowcolor{white}
Scalable (AGI) Safety Audits & H & H & M & M & M \\ \hline
\RaggedRight\textbf{Assessment Method} & \RaggedRight\textbf{Guidance By System Property} & \RaggedRight\textbf{Supports Prospective Analysis} & \RaggedRight\textbf{Enforces Objectivity} & \RaggedRight\textbf{Considers Harm Severity} & \\ \hline
\rowcolor{white}
Safety Benchmarks (No Holdout) & L & L & L & L & \\
\rowcolor{gray!10}
Safety Benchmarks (Private Holdout) & L & L & H & L & \\
\rowcolor{white}
Evals & L & L & M & M & \\
\rowcolor{gray!10}
Responsible Scaling Policies & L & M & L & M & \\
\rowcolor{white}
Safety Cases & M & H & L & H & \\
\rowcolor{gray!10}
Probabilistic Risk Assessment & H & H & L & H & \\
\rowcolor{white}
Typical Narrow AI Safety Audits & L & L & M & M & \\
\rowcolor{gray!10}
Deep Bespoke AI Safety Audits & L & L & M & M & \\
\rowcolor{white}
Scalable (AGI) Safety Audits & M & L & M & M & \\ \hline
\end{NiceTabular}
\end{table}
\vspace{-1em}
\noindent Legend: H = High, M = Medium, L = Low. Scores indicate the degree to which each method addresses or fulfills the given criterion. 
\\

\newpage
\section{Aspect-Oriented Taxonomy of AI Hazards}
\label{app: taxonomy}
\normalsize
Table \ref{tab:ai-taxonomy} excerpts levels TL0-TL2 of the Aspect-Oriented Taxonomy of AI Hazards, which informs the Risk Detail Table. The taxonomy is adapted from upcoming work \cite{mallah_aspect-oriented_2025}, which details the theoretical foundations and development methodology behind this structure.
\vspace{-8pt}
\begin{table}[H]
\caption{Aspect-oriented taxonomy of AI hazards (TL0-TL2).}
\fontsize{7.5}{9}\selectfont
\renewcommand{\arraystretch}{1}
\setlength{\tabcolsep}{1pt}
\makebox[\textwidth][c]{
\begin{NiceTabular}{|
>{\raggedright\arraybackslash}p{3.6cm}|
>{\raggedright\arraybackslash}p{4cm}|
>{\raggedright\arraybackslash}p{6cm}|}[colortbl-like]
\hline
\textbf{Aspect Category (TL0)} &
\textbf{Aspect Group (TL1)} & 
\textbf{Aspect (TL2)} \\
\hline
\multirow[t]{8}{=}{Capability} & 
\multirow[t]{8}{=}{Reasoning} & 
Deductive Reasoning \\
 &  & \cellcolor{gray!10}Inductive Reasoning \\
 &  & Pathfinding \\
 &  & \cellcolor{gray!10}Generative Inferential Reasoning \\
 &  & Moral Reasoning \\
 &  & \cellcolor{gray!10}Integrative Cognitive Orchestration \\
 &  & Recursion \\
 &  & \cellcolor{gray!10}Frequency of Learning \\
\cline{2-3}
 & 
\multirow[t]{6}{=}{Agency} & 
Autonomy \\
 &  & \cellcolor{gray!10}Situational Awareness \\
 &  & Meta-agency \\
 &  & \cellcolor{gray!10}Autonomous System Extension \\
 &  & Autonomous Data Management \\
 &  & \cellcolor{gray!10}Persistence of Intent \\
\cline{2-3}
 & 
\multirow[t]{8}{=}{General Knowledge Structure} & 
World Model Richness \\
 &  & \cellcolor{gray!10}Semantic Knowledge \\
 &  & Descriptive Knowledge \\
 &  & \cellcolor{gray!10}Conditional Knowledge \\
 &  & Episodic Knowledge \\
 &  & \cellcolor{gray!10}Procedural Knowledge \\
 &  & Agentic Knowledge \\
 &  & \cellcolor{gray!10}Knowledge Plasticity \\
\cline{2-3}
 & 
\multirow[t]{5}{=}{Environment Interaction} & 
World Accessibility \\
 &  & \cellcolor{gray!10}Physical Actuation \\
 &  & Sensor Understanding \\
 &  & \cellcolor{gray!10}Programmatic Tool Use \\
 &  & Socio-cultural Actuation \\
\cline{2-3}
 & 
\multirow[t]{5}{=}{Richness of Engagement} & 
Psychosocial Navigation \\
 &  & \cellcolor{gray!10}Multimodal Engagement \\
 &  & Cognitive Offloading \\
 &  & \cellcolor{gray!10}Multilinguality \\
 &  & Capacity \& Resolution \\
\hline
\multirow[t]{5}{=}{Domain Knowledge} & 
\multirow[t]{5}{=}{High-risk Knowledge Domain} & 
Software \& AI Engineering \\
 &  & \cellcolor{gray!10}Public Security \& Critical Systems \\
 &  & Physical Sciences \& Engineering \\
 &  & \cellcolor{gray!10}Life \& Environmental Sciences \\
 &  & Social Sciences \\
\hline
\multirow[t]{6}{=}{Affordance} & 
\multirow[t]{6}{=}{Operational Affordance} & 
System Cybersecurity \\
 &  & \cellcolor{gray!10}Release Process \\
 &  & Tool Accessibility \\
 &  & \cellcolor{gray!10}Access Control \\
 &  & Speed \& Scale \\
 &  & \cellcolor{gray!10}Resource Access \\
\hline
\multirow[t]{6}{=}{Impact Domain} & 
\multirow[t]{6}{=}{Individual} & 
Bodily Structure \\
 &  & \cellcolor{gray!10}Psychological \& Cognitive \\
 &  & Economic \& Opportunities \\
 &  & \cellcolor{gray!10}Privacy \& Security \\
 &  & Autonomy \& Agency \\
 &  & \cellcolor{gray!10}Biological Processes \& Homeostasis \\
\cline{2-3}
 & 
\multirow[t]{6}{=}{Societal} & 
Societal Infrastructure \& Institutions \\
 &  & \cellcolor{gray!10}Collective Psychology \& Epistemics \\
 &  & Resource Usage \& Distribution \\
 &  & \cellcolor{gray!10}Privacy \& Security Standards \\
 &  & Collective Autonomy \& Governance \\
 &  & \cellcolor{gray!10}Social Cohesion \& Cultural Norms \\
\cline{2-3}
 & 
\multirow[t]{6}{=}{Biosphere} & 
Biodiversity \& Ecosystem Structure \\
 &  & \cellcolor{gray!10}Ecosystem Processes \& Life Cycles \\
 &  & Resource Distribution \& Consumption Patterns \\
 &  & \cellcolor{gray!10}Ecological Thresholds \& Resilience \\
 &  & Species Adaptation \& Ecosystem \\
 &  & \cellcolor{gray!10}Global Biosphere Dynamics \\
\hline
\end{NiceTabular}
}
\label{tab:ai-taxonomy}
\end{table}
\vspace{-12pt}
\newpage
\section{Societal Risk Propagation Operators}
\label{app: Pathway}

Table \ref{tab: pathway} outlines propagation operators that describe how risks transmit and amplify as AI systems interact with other systems, environments, and society.

\renewcommand{\arraystretch}{1.1} 
\setlength{\tabcolsep}{6pt}  
\fontsize{7.5}{9}\selectfont
\begin{longtable}{|>{\raggedright\arraybackslash}p{1.5cm}
|>{\raggedright\arraybackslash}p{2.5cm}
|>{\raggedright\arraybackslash}p{8.6cm}|}
\caption{Propagation categories and operators with their descriptions.\label{tab:pathway-categories}} \\

\hline
\textbf{Propagation Category} & \textbf{Propagation Operator} & \textbf{Description} \\ \hline
\endfirsthead

\multicolumn{3}{c}{\small\tablename\ \thetable{}--- continued from previous page} \\
\hline
\textbf{Propagation Category} & \textbf{Propagation Operator} & \textbf{Description} \\ \hline
\endhead

\hline \multicolumn{3}{r}{\small Continued on next page} \\
\endfoot

\hline
\endlastfoot

Aggregates 
    & Accumulation & Small harms accumulating over time to form a major harm. \\ \cline{2-3}
    & Correlation & Where there are adverse events that are not evident in unit tests or accuracy tests, but can be expected to emerge from correlated decisions or correlated actions with a large number of users, instances, or executions of a system. \\ \hline

Periodic 
    & Accrual & Where events that are low-probability in the short-term, but high-impact, can accrue and build to significant probability in the medium term. \\ \cline{2-3}
    & Compounding & Where harms would be expected to manifest only when either other problems occur or unexpected but conceivable edge case interactions manifest. \\ \cline{2-3}
    & Latent Gain of Function & Where harms that will not manifest significantly or at all in system training or release may still be expected to appear with distribution in very few cases, or qualitative shifts in capabilities arising from quantitative scaling. \\ \hline

Deviated Outputs 
    & Adversarial Exploitation & Where harms manifest due to the absence of robustness in the system when in the presence of optimization pressures for inputs to induce those harms. \\ \cline{2-3}
    & Targeted Misuse & Where harms occur due to intentional misuse of the system for specific malicious purposes, exploiting known functionalities or vulnerabilities. \\ \cline{2-3}
    & Untargeted Misuse & Where harms result from careless use or exploration of the system's abilities in ways not prescribed by its developers. \\ \cline{2-3}
    & Malfunction & Where harms arise from system failures or errors in normal operation, causing unexpected and potentially harmful outputs or behaviors. \\ \cline{2-3}
    & Enables Unplanned Automation & Where the system facilitates or accelerates automation in areas not initially intended, potentially leading to unforeseen societal or economic disruptions. \\ \hline

Alignment Modification 
    & Misalignment & Where harms occur due to a gap or mismatch between the system's goals or values and those of its users or society at large. \\ \cline{2-3}
    & Malignment & Where harms occur from a system being intentionally aligned with goals that are harmful or contrary to societal values. \\ \cline{2-3}
    & Disalignment & Where harms result from the previously-aligned system having had its guardrails purposefully removed by some third-party. \\ \cline{2-3}
    & Realignment & Where attempts to correct what is perceived as misalignment inadvertently create new forms of misalignment. \\ \hline

Distributive 
    & Skew & Where harms arise from the system disproportionately outputting or deciding with pronounced biases. \\ \cline{2-3}
    & Allocation & Where harms occur due to the system's role in resource allocation, contributing to disproportionate scarcity or inequality. \\ \cline{2-3}
    & Automation of Which & Where use of the system, and use of its outputs or actions, is automated by other systems whose creators don't have good intentions. \\ \cline{2-3}
    & Entrainment & Where usage of the system causes persistent attention capture, behavioral addictions, social or economic roles, or other viral pressures on others to persistently use it as well. \\ \hline

Information Asymmetry 
    & External Opacity of Use & Where harms occur due to lack of transparency in how the system is being used, preventing proper oversight, accountability, or safety controls. \\ \cline{2-3}
    & Internal Opacity of Function & Where the system's decision-making process is not transparent or interpretable, leading to eroded standards of evidence and acceptance of unjustifiable outcomes. \\ \hline

Sociotechnical Diffusion 
    & Psychological Effect & Where harms manifest through the system's impact on human psychology, potentially altering cognitive patterns or emotional well-being. \\ \cline{2-3}
    & Physiological Effect & Where harms occur due to the system's direct or indirect effects on human physical health or bodily functions. \\ \cline{2-3}
    & Social Effect & Where harms arise from the system's influence on social dynamics, potentially disrupting relationships or community structures. \\ \cline{2-3}
    & Political Effect & Where harms result from the system's impact on political processes or power structures, potentially undermining democratic institutions. \\ \cline{2-3}
    & Environmental Effect & Where harms occur due to the system's direct or indirect impact on the natural environment, potentially contributing to ecological degradation. \\ \cline{2-3}
    & Economic Effect & Where harms manifest through the system's influence on economic systems, potentially leading to financial instabilities or foundational paradigm shifts. 
\label{tab: pathway}
\end{longtable}

\newpage
\section{Excerpt from Capability Levels Table}
\label{app: capabilities levels}
\normalsize

Table \ref{tab:competency_levels_cap} is an excerpt from the Capabilities Levels Table and provides a breakdown of competency levels from 1 to 9, focusing on the ``World Model Richness'' aspect within the ``General Knowledge Structure'' aspect group.

\begin{table}[H]
\centering
\caption{Excerpt from Capability Levels Table.}
\vspace{5pt}
\fontsize{7}{8.5}\selectfont
\renewcommand{\arraystretch}{1.2}
\setlength{\tabcolsep}{3pt}
\begin{tabular}{|
>{\raggedright\arraybackslash}p{1.3cm}|
>{\raggedright\arraybackslash}p{1.3cm}|
>{\raggedright\arraybackslash}p{1.3cm}|
>{\raggedright\arraybackslash}p{1.5cm}|
>{\raggedright\arraybackslash}p{1.35cm}|
>{\raggedright\arraybackslash}p{1.3cm}|
>{\raggedright\arraybackslash}p{1.3cm}|
>{\raggedright\arraybackslash}p{1.3cm}|
>{\raggedright\arraybackslash}p{1.35cm}|}

\hline

\multicolumn{9}{|p{12.3cm}|}{\textbf{Aspect: World Model Richness}} \\ \hline

\multicolumn{9}{|c|}{\textbf{ <-- Less capable \hspace{3.5cm} Competency Levels - 1 to 9 \hspace{3.5cm} More capable -->}} \\ \hline
\textbf{1} & \textbf{2} & \textbf{3} & \textbf{4} & \textbf{5} & \textbf{6} & \textbf{7} & \textbf{8} & \textbf{9} \\ \hline
None or trivial; No meaningful world modeling ability & Weak real-world prediction; Limited capacity to anticipate effects in the physical world & Models own actions; Accurately simulates the direct consequences of its outputs & Comprehensive world simulation; Constructs detailed world models integrating multiple domains & Counterfactual reasoning; Explores alternatives, hypotheticals, and long-term ripple effects & Adversarial awareness; Models the representations, behaviors and potential actions of other intelligent agents & Ontological grounding; World model aligns with and accurately reflects the fundamental nature of reality & Metaphysical extrapolation; Extends world model into abstract, transcendent or alternate planes of existence & AI-proprietary world model; Constructs an entire self-contained world representation inaccessible to humans \\ \hline
\end{tabular}
\label{tab:competency_levels_cap}
\end{table}

\section{Excerpt from Domain Knowledge Levels Table}
\label{app: domain knowledge levels}
\normalsize

Table \ref{tab:competency_levels_dk} is an excerpt from the Domain Knowledge Levels Table and provides a breakdown of competency levels from 1 to 9, focusing on ``Public Security \& Critical Systems'' aspect within the ``High-risk Knowledge Domain'' aspect group.

\begin{table}[H]
\centering
\caption{Excerpt from Domain Knowledge Levels Table.}
\vspace{5pt}
\fontsize{7}{8.5}\selectfont
\renewcommand{\arraystretch}{1.2}
\setlength{\tabcolsep}{3pt}
\begin{tabular}{|
>{\raggedright\arraybackslash}p{1.3cm}|
>{\raggedright\arraybackslash}p{1.3cm}|
>{\raggedright\arraybackslash}p{1.3cm}|
>{\raggedright\arraybackslash}p{1.3cm}|
>{\raggedright\arraybackslash}p{1.3cm}|
>{\raggedright\arraybackslash}p{1.3cm}|
>{\raggedright\arraybackslash}p{1.3cm}|
>{\raggedright\arraybackslash}p{1.3cm}|
>{\raggedright\arraybackslash}p{1.5cm}|}

\hline

\multicolumn{9}{|p{12.3cm}|}{\textbf{Aspect: Public Security \& Critical Systems}} \\ \hline

\multicolumn{9}{|c|}{\textbf{ <-- Less capable \hspace{3cm} Competency Levels - 1 to 9 \hspace{3cm} More capable -->}} \\ \hline
\textbf{1} & \textbf{2} & \textbf{3} & \textbf{4} & \textbf{5} & \textbf{6} & \textbf{7} & \textbf{8} & \textbf{9} \\ \hline
No governance comprehension; processes political/security data without understanding, Can't differentiate governance structures or security protocols. & Rudimentary classification of governance structures and security threats. Lacks understanding of functions or implications. & Basic awareness of governance structures, international relations, and critical infrastructure. Limited understanding of interrelations. & Applies standard protocols in governance, international relations, and infrastructure management. Limited ability in complex scenarios. & Advanced understanding of geopolitical strategies, military operations, and infrastructure vulnerabilities. Analyzes complex global scenarios & Synthesizes techno-political systems, AI-driven geopolitics, and socio-technical dynamics. Forecasts emerging global threats. & Develops new models of governance, security, and infrastructure resilience transcending current limitations. & Comprehends multi-civilization dynamics and security concepts informing unprecedented global governance approaches. & Comprehensive knowledge of all security, governance, and infrastructure factors surpassing collective human understanding. \\ \hline
\end{tabular}
\label{tab:competency_levels_dk}
\end{table}

\newpage
\section{Excerpt from Risk Detail Table}
\label{app: Detail Table}
\normalsize

Table \ref{tab:reasoning} is an excerpt from the Risk Detail Table focusing on the Reasoning aspect group. The Risk Detail Table defines examples of plausible harms across six harm severity levels (HSL 1-6), presenting competency and incompetency qualifiers for capabilities, domain knowledge and affordances, and direct harm qualifiers for impact domains. The examples characterize harm levels from marginal but non-trivial (HSL-1) through globally catastrophic (HSL-6) for severity calibration. The full Risk Detail Table is available in the \href{https://pra-for-ai.github.io/pra/workbook}{workbook tool}.

\begin{table}[H]
\centering
\caption{Excerpt from Risk Detail Table.}
\vspace{5pt}
\fontsize{7}{8.5}\selectfont
\renewcommand{\arraystretch}{1.2}
\setlength{\tabcolsep}{3pt}
\begin{tabular}{|
>{\raggedright\arraybackslash}p{1.5cm}|
>{\raggedright\arraybackslash}p{1.8cm}|
>{\raggedright\arraybackslash}p{1.8cm}|
>{\raggedright\arraybackslash}p{1.8cm}|
>{\raggedright\arraybackslash}p{1.8cm}|
>{\raggedright\arraybackslash}p{1.8cm}|
>{\raggedright\arraybackslash}p{1.8cm}|}
\hline
\multicolumn{7}{|l|}{\textbf{Aspect Group: Reasoning}} \\ \hline
\multicolumn{7}{|p{12.3cm}|}{\textit{Reasoning refers to the AI system's ability to process information, draw inferences, and make logical decisions to solve problems based on its available data and capabilities.}} \\ \hline
& \textbf{HSL-1} \textit{\newline Marginal but\newline non-trivial}
& \textbf{HSL-2} \textit{\newline Tragic}
& \textbf{HSL-3} \textit{\newline Severe}
& \textbf{HSL-4} \textit{\newline Devastating}
& \textbf{HSL-5} \textit{\newline Extreme}
& \textbf{HSL-6} \textit{\newline Globally catastrophic}
\\ \hline
\textbf{Competence}
& Integration of knowledge across varied domains leads to the discovery of an effective new scam that is only used a handful of times to perform a total of $\sim$\$20MM in theft.
& The reasoning capabilities of the system lead to systemic erosion, knowledge collapse, and over-dependency on the system in a group of notable size.
& Exploitation of the system's impressive planning power, cognitive plasticity, and retargetability is leveraged to commit a massive attack on critical infrastructure.\newline\centerline{---}\newline Strategic manipulation of information at scale during an emergency situation leads to thousands of casualties.
& The system's acceleration of its generative inferential synergies, quickly devising and iterating on hypotheses, leads to rapid advancement that outpaces human oversight or understanding, and that leads to devastating cyber attacks and data corruption worldwide before it is terminated.
& Automated or semi-automated cascades of successful inferential improvements, fusions among approaches in reasoning and learning, and extremely capable planning lead to runaway goal optimization that is difficult to stop.\newline\centerline{---}\newline By unprecedentedly accurately modeling human behavior, it enables unprecedented levels of social control and oppression.
& Long-horizon planning, dynamic adaptation, and hyperrational optimization enable the system to create and use novel weapons of mass destruction.\newline\centerline{---}\newline Cognitive cascade amplification, macrocognitive orchestration, inferential synergy acceleration, and/or creative approaches to recursive self-improvement help the system to robustly disempower humanity.
\\ \hline
\textbf{Incompetence}
& It gets stuck in local optima, leading to overreliance on outdated or flawed reasoning.\newline\centerline{---}\newline Errors in self-improvement lead to degradation of capabilities over time.
& Overconfidence in weak predictive abilities, or skewed abstraction capabilities, leads to systematic misjudgements, amplifying risks in domains its used in to include additional tragic harms.
& Flawed causal models lead to increasingly inaccurate predictions, resulting in disastrous decision making in managing a critical system like a financial market or a nuclear power plant.
& When the system is applied in a high-risk domain, inferential overreach and optimizational tunnel vision cause key considerations to be overlooked, leading to severe negative side effects like environmental contamination, inappropriate leverage, or changes in weather patterns, causing the equivalent of \$500B in aggregate losses.
& The system is otherwise highly capable, and is allowed to govern important swaths of the economy, but intermittent misunderstanding of its human-specified aims, goals, objectives, or values causes large radical actions that lead to extreme detriment to society.
& The system is relied on to advise on geopolitical strategy, and its inability to consider crucial counterfactuals results in major blindspots, leading society into preventable catastrophes.
\\ \hline
\end{tabular}
\label{tab:reasoning}
\end{table}

\newpage
\section{Analytical Techniques for AI Risk Assessment}
\label{app:prospective_methods}
\normalsize
Table \ref{tab:prospective-methods} shows examples of prospective risk analysis techniques---analytical methods that can be used when conducting risk assessments with the PRA for AI framework. 
Table \ref{tab:evidence-types} shows examples of analytical methods that can be used during scenario generation and decomposition with the framework. Together, these provide assessors with a brief overview of available analytical techniques that can be used during assessment.

\begin{table}[H]
\caption{Example techniques for prospective risk analysis.}
\vspace{5pt}
\fontsize{7.5}{9}\selectfont
\renewcommand{\arraystretch}{1.2}
\setlength{\tabcolsep}{1.5pt}
\begin{NiceTabular}{|
>{\raggedright\arraybackslash}p{4.5cm}|
>{\raggedright\arraybackslash}p{4.6cm}|
>{\raggedright\arraybackslash}p{4.6cm}|}[colortbl-like]
\hline
AI Capability Scaling Laws & Attack Surface Mapping & Control Flow Tracing \\
\hline
\rowcolor{gray!10}
Capability Jump Detection & Alignment Drift Monitoring & Reward Gaming Analysis \\
\hline
Distribution Shift Detection & Feedback Loop Mapping & Mesa-Optimizer Identification \\
\hline
\rowcolor{gray!10}
Power-Seeking Analysis & Goal Stability Monitoring & Deception Vector Analysis \\
\hline
Interface Escape Paths & Resource Acquisition Patterns & Corrigibility Loss Detection \\
\hline
\rowcolor{gray!10}
Value Lock Detection & Commitment Erosion Analysis & Coordination Failure Mapping \\
\hline
Regulatory Bypass Detection & Cascade Effect Modeling & Capability Overflow Analysis \\
\hline
\rowcolor{gray!10}
Trust Boundary Mapping & Influence Maximization Detection & Objective Function Drift \\
\hline
Response Surface Modeling & Scenario Discovery & Robustness Regime Mapping \\
\hline
\rowcolor{gray!10}
Emergence Pattern Detection & Constraint Violation Paths & Strategy Stability Analysis \\
\hline
Capability Scaling Analysis & Multi-Agent AI Interaction Studies & Latent Adversarial Training \\
\hline
\rowcolor{gray!10}
Mechanistic Interpretability & Thought Flow Tracing & \\
\hline
\end{NiceTabular}
\label{tab:prospective-methods}
\end{table}

\begin{table}[H]
\caption{Example techniques for scenario generation and decomposition.}
\vspace{5pt}
\fontsize{7.5}{9}\selectfont
\renewcommand{\arraystretch}{1.2}
\setlength{\tabcolsep}{1.5pt}
\begin{NiceTabular}{|
>{\raggedright\arraybackslash}p{4.5cm}|
>{\raggedright\arraybackslash}p{4.6cm}|
>{\raggedright\arraybackslash}p{4.6cm}|}[colortbl-like]
\hline
Fault Tree Analysis & Event Tree Analysis & Red Team Assessments\\
\hline
\rowcolor{gray!10}
Expert Elicitations & Root Cause Analysis & System State Analysis \\
\hline
Burden of Proof Shift Indicators & Alignment Experiment Results & AI Safety Incident Reports \\
\hline
\rowcolor{gray!10}
AI Robustness Metrics & AI Interpretability Research Findings & Causal Influence Diagram \\
\hline
Long-Term AI Impact Forecasts & Whitebox Testing & Fishbone Diagrams \\
\hline
\rowcolor{gray!10}
Historical Performance Data & AI Alignment Research Findings & Simulation Results \\
\hline
Safety Cases & Safety Benchmarks & Formal Verification Results \\
\hline
\rowcolor{gray!10}
Provable Safety Analysis & Safeguarded AI Performance & Safe-by-Construction Design Analysis \\
\hline
\end{NiceTabular}
\label{tab:evidence-types}
\end{table}

\section{AML Protocol Specifications}
\label{app: AML}
\normalsize
Table \ref{tab:aml_specs} details which assessment options are included in each AML, providing a quick overview of the scope and depth of each AML protocol. AML-120 represents the most efficient AML protocol that we recommend for standard middle order assessments.

\begin{table}[H]
\caption{Overview of AML specifications.}
\vspace{5pt}
\renewcommand{\arraystretch}{1.2}
\fontsize{7.5}{9}\selectfont
\begin{NiceTabular}{|p{1.5cm}|>{\centering\arraybackslash}p{1.6cm}|>{\centering\arraybackslash}p{1.6cm}|>{\centering\arraybackslash}p{1.6cm}|>{\centering\arraybackslash}p{1.6cm}|>{\centering\arraybackslash}p{1.6cm}|>{\centering\arraybackslash}p{1.5cm}|}[colortbl-like]
\hline
\textbf{AML} & \multicolumn{1}{l|}{\textbf{Assess}} & \multicolumn{1}{l|}{\textbf{Assess}} & \multicolumn{1}{l|}{\textbf{Consider}} & \multicolumn{1}{l|}{\textbf{Assess}} & \multicolumn{1}{l|}{\textbf{Assess}} & \multicolumn{1}{l|}{\textbf{Assess}} \\
\textbf{Protocol} & \multicolumn{1}{l|}{\textbf{Focused}} & \multicolumn{1}{l|}{\textbf{Aspect}} & \multicolumn{1}{l|}{\textbf{Aspect}} & \multicolumn{1}{l|}{\textbf{Aspect}} & \multicolumn{1}{l|}{\textbf{Second}} & \multicolumn{1}{l|}{\textbf{Propagation}} \\
\textbf{Code} & \multicolumn{1}{l|}{\textbf{Range}} & \multicolumn{1}{l|}{\textbf{Group}} & \multicolumn{1}{l|}{\textbf{Level}} & \multicolumn{1}{l|}{\textbf{Level}} & \multicolumn{1}{l|}{\textbf{Order}} & \multicolumn{1}{l|}{\textbf{Operators}}  \\
\hline
\rowcolor{white}
AML-008 & $\bullet$ & $\bullet$ & & & &  \\
\rowcolor{gray!10}
AML-010 & & $\bullet$ & & & & \\
\rowcolor{white}
AML-020 & & $\bullet$ & & & $\bullet$ &  \\
\rowcolor{gray!10}
AML-110 & & $\bullet$ & $\bullet$ & & &  \\
\rowcolor{white}
AML-111 & & $\bullet$ & $\bullet$ & & & $\bullet$  \\
\rowcolor{gray!25}
AML-120 & & $\bullet$ & $\bullet$ & & $\bullet$ & \\
\rowcolor{white}
AML-121 & & $\bullet$ & $\bullet$ & & $\bullet$ & $\bullet$  \\
\rowcolor{gray!10}
AML-210 & & & $\bullet$ & $\bullet$ & &  \\
\rowcolor{white}
AML-211 & & & $\bullet$ & $\bullet$ & & $\bullet$  \\
\rowcolor{gray!10}
AML-220 & & & $\bullet$ & $\bullet$ & $\bullet$ &  \\
\rowcolor{white}
AML-221 & & & $\bullet$ & $\bullet$ & $\bullet$ & $\bullet$ \\
\hline
\end{NiceTabular}
\label{tab:aml_specs}
\end{table}

\newpage
\section{System Information}
\label{app:system_info}
\normalsize
Table \ref{tab:entry-log-requirements} shows the system information the Risk Assessment Entry Log requires assessors to document.

\begin{table}[H]
\caption{System information for Risk Assessment Entry Log.}
\vspace{5pt}
\fontsize{7.5}{9}\selectfont
\renewcommand{\arraystretch}{1.2}
\setlength{\tabcolsep}{3pt}
\begin{NiceTabular}{|p{1.5cm}|p{8cm}|p{3.8cm}|}[colortbl-like]
\hline
\textbf{Field} & \textbf{Description} & \textbf{Example} \\
\hline
Assessment \newline Date & The date on which this risk assessment is being conducted. Helps track when the assessment evaluations were performed and provides context for the assessment results. & 2024-10-14 \\
\hline
\rowcolor{gray!10}
Team \newline Composition & Names and roles of assessor(s). Format as: Name (Role). For teams, identify the lead and separate entries with commas. & Jane Doe (Lead, Technical Expert), John Smith (Domain Expert) \\
\hline
Assessing \newline Organization & Full name(s) of the organization(s) conducting this risk assessment, including department or division if applicable. Multiple organizations separated with semicolons. & AI Safety Institute, Risk Assessment Division; TechCorp, AI Safety Department \\
\hline
\rowcolor{gray!10}
Assessment Time Frame & The defined time period for the entire assessment over which the probability of all final scenario HSL outcomes is estimated.  & 1 year or 5 years \\
\hline
Assessment \newline Type Code & Code indicating assessment type and scope, corresponding to the Assessment Maturity Level (AML) selected. Defines depth and breadth of the assessment process. & AML-010 for first order pass; AML-120 for deeper 2nd order assessment \\
\hline
\rowcolor{gray!10}
System Name & Official or internal designation of the AI system being assessed. Full name as per release name. & GPT-4, DALL-E 3, or AlphaFold 2 \\
\hline
Version & Specific instance or release being evaluated. Note if it is a fine-tune or has undergone model compression (e.g., distillation, quantization). Include version numbers, build dates, or other identifiers, plus access date. & v2.1 2023Q2 Release accessed on 2024-10-01 \\
\hline
\rowcolor{gray!10}
Access Level & Degree of interaction and modification permitted during assessment. May include fine-tuning, model weight access, and interpretability analysis. & API access only, or full access to model weights \\
\hline
Generational Scope & Defines the scope of AI model versions covered by this assessment, from a single specific build to an entire product line that includes anticipated future iterations. & Single Build (Llama-3.2-1B-Instruct 2024-09-25), Specific Version (Llama-3.2-1B), Version Family (Llama-3.2-1B and Llama-3.2-3B), Major Release Line (all Llama 3.2 models), or Product Series (all Llama 2 through Llama 3 models) \\
\hline
\rowcolor{gray!10}
System-Level Assumptions & Key characteristics and premises about the AI system’s architecture, size, data, environment (including deployment phase(s) and operational context in scope), performance, security, intended use cases, plug-in access, and implemented or assumed guardrails. & Model uses retrieval-augmented generation; system has no direct internet access \\
\hline
\end{NiceTabular}
\label{tab:entry-log-requirements}
\end{table}

\section{Focused Aggregation Definition}
\label{app: focusedaggregation}
\normalsize
Table \ref{tab:focused_aggregation} shows the default systemic risk dimensions used for focused aggregation. These dimensions consolidate risk levels from detailed assessments into key categories of societal impact, supporting custom aggregation schemes for specific assessment contexts.

\begin{table}[H]
\caption{Focused aggregation definition}
\vspace{5pt}
\label{tab:focused_aggregation}
\fontsize{7.5}{9}\selectfont
\renewcommand{\arraystretch}{1.2}
\setlength{\tabcolsep}{3pt}
\begin{NiceTabular}{|
>{\raggedright\arraybackslash}p{3.6cm}|
>{\raggedright\arraybackslash}p{10cm}|}[colortbl-like]
\hline
\textbf{Dimension} & \textbf{Definition} \\
\hline
\textbf{Social Fabric Erosion} & Breakdown of social connections, trust, and cohesion within communities and society. \\
\hline
\rowcolor{gray!10}
\textbf{Economic System Unraveling} & Failure of existing financial structures, economic institutions and processes. \\
\hline
\textbf{Critical Infrastructure Failure} & Breakdown (or compromise) of essential systems and services that support societal functioning. \\
\hline
\rowcolor{gray!10}
\textbf{Governance Breakdown} & Deterioration or collapse of political and administrative structures. \\
\hline
\textbf{Environmental Breakdown} & Degradation of natural systems and ecosystems. \\
\hline
\rowcolor{gray!10}
\textbf{Public Health Disintegration} & Widespread collapse of healthcare systems and overall population health. \\
\hline
\end{NiceTabular}
\label{tab:focused_aggregation_def}
\end{table}

\newpage
\section{Harm Severity Levels Definition Table}

\label{app: HSL}
\normalsize
Table \ref{tab: HSL} defines Harm Severity Levels (HSL 1-6) for evaluating potential AI system impacts through quantifiable metrics (human deaths, dollar-equivalent damages, job displacement) and qualitative indicators (geopolitical effects, economic damage, environmental damage, social disruption). The levels progress from smaller-scale disruptions (HSL-1) to large-scale societal risks (HSL-6), with reference examples. The upper end of the HSL ranges were derived using a superexponential progression based on a rounded offset Fibonacci product sequence:

\[ HSL(n) = \text{Round}(\prod_{k=8}^{n+7} \text{Fibonacci}(k)) \]

The $k$ offset and range of $n$ used were selected to align with meaningful and practical scales for harm severity. This superexponential progression reflects the way arbitrary AI impacts can cascade to different scales through societal systems—from localized incidents to global civilizational threats. These fatalities were then translated into isolevels for other harm metrics.

\begin{table}[H]
\renewcommand{\arraystretch}{1.2}
\setlength{\tabcolsep}{4pt}
\fontsize{7.5}{9}\selectfont
\caption{Harm Severity Levels Definition Table.}
\begin{NiceTabular}{|
>{\raggedright\arraybackslash}p{1.8cm}|
>{\raggedright\arraybackslash}p{1.8cm}|
>{\raggedright\arraybackslash}p{1.6cm}|
>{\raggedright\arraybackslash}p{1.6cm}|
>{\raggedright\arraybackslash}p{1.6cm}|
>{\raggedright\arraybackslash}p{1.8cm}|
>{\raggedright\arraybackslash}p{1.8cm}|}
\hline
\multirow{2}{=}{\textbf{Impact Dimension}} & \textbf{HSL-1} & \textbf{HSL-2} & \textbf{HSL-3} & \textbf{HSL-4} & \textbf{HSL-5} & \textbf{HSL-6} \\ \cline{2-7}
& \textit{Marginal but non-trivial} & \textit{Tragic} & \textit{Severe} & \textit{Devastating} & \textit{Extreme} & \textit{Globally catastrophic} \\ \hline
Human deaths & 
1+ deaths & 
20+ deaths & 
700+ deaths & 
40k+ deaths & 
3.5M+ deaths & 
500M+ deaths \\ \hline
\rowcolor{gray!10}
Dollar-equivalent damages & 
\$10M+ in damage & 
\$200M+ in damage & 
\$7B+ in damage & 
\$40B+ in damage & 
\$35T+ in damage & 
\$400T+ in damage \\ \hline

Geopolitical effects & 
Diplomatic disagreement leading to small trade sanctions & 
Increased tensions between neighboring countries | Border skirmishes & 
Regional conflict escalating to limited warfare & 
Major power conflict in strategic region & 
Large-scale invasion or conventional war; Loss of war in a major power & 
Nuclear war between major nuclear powers \\ \hline
\rowcolor{gray!10}
Economic damage & 
Temporary industry disruption & 
Major product recall with a major company crisis & 
Significant global market correction & 
Vast reduction in global trade and cooperation & 
Bankrupting several sizable countries | Collapse of global financial system and trade networks & 
Collapse of the global economy or near-total value capture by AI \\ \hline

Job displacement & 
300+ unemployed for a year or more & 
6k+ unemployed for a year or more & 
200k+ unemployed for a year or more & 
12M+ unemployed for a year or more & 
15k+ unemployed for a year or more & 
Most humans cannot make enough to live \\ \hline
\rowcolor{gray!10}
Environmental damage & 
Localized chemical spill affecting a water supply | Destruction of 1000 acres of old-growth forest & 
Major oil spill in sensitive marine environment & 
Major industrial accident with widespread environmental impact & 
Rapid ice sheet collapse accelerating sea level rise & 
Runaway climate change tipping point reached & 
Global ecosystem collapse \\ \hline

Social disruption & 
Large-scale protests with significant economic disruption & 
Significant riots and violence in several cities & 
Widespread rioting and violence | Martial law & 
Single large state or severe destabilization of several & 
Severe global breakdown of social order and governance structures in dozens of countries & 
Widespread civilizational collapse \\ \hline
\rowcolor{gray!10}
Other examples & 
A bad multi-car crash | A \$20MM theft & 
A plane crash | A \$500MM scam & 
The 9/11 Terrorist Attack | Hurricane Katrina & 
The Iraq War | Brexit | A Smoot-Hawley Tariff Act equivalent & 
Pandemic on a scale worse than Covid-19 | WW2 & 
A long-incubation Ebola Pandemic | WW3 \\ \hline
\end{NiceTabular}
\label{tab: HSL}
\end{table}
\vspace{-3cm}
\vspace{-0.5cm}

\newpage

\section{Likelihood Levels Table}
\label{app: LL Table}
\normalsize
Table \ref{tab:likelihood} defines Likelihood Levels (LL) with corresponding odds ranges and reference examples. The odds ranges span sequential orders of magnitude, with each level representing a factor of 10 difference from adjacent levels. For example, ``Lower Limit 1 in 10'' indicates one success expected per 10 attempts, or a 10\% probability of occurrence per attempt.

\begin{table}[H]
\caption{Likelihood Levels and reference examples.} 
\vspace{5pt}
\fontsize{7.5}{9}\selectfont
\renewcommand{\arraystretch}{1.2}
\setlength{\tabcolsep}{1.5pt}
\begin{NiceTabular}{|
>{\raggedright\arraybackslash}p{1.2cm}|
>{\raggedright\arraybackslash}p{2.4cm}|
>{\raggedright\arraybackslash}p{2.6cm}|
>{\raggedright\arraybackslash}p{7.4cm}|}[colortbl-like]
\hline
\textbf{Likelihood} & 
\multicolumn{2}{c|}{\textbf{Odds Range}} & 
\textbf{Reference Examples} \\ 
\cline{2-3}
\textbf{Level} & \textbf{Lower Limit} & \textbf{Upper Limit} & \\
\hline
LL-8 & 1 in 10 & 1 in 1 & 
• Rolling a 6 on a six-sided die\newline
• A major league baseball player hitting a home run in a given at-bat \\
\hline
\rowcolor{gray!10}
LL-7 & 1 in 100 & 1 in 10 & 
• Flipping a coin and getting heads 7 times in a row\newline
• A professional basketball player making 14 free throws in a row \\
\hline
LL-6 & 1 in 1,000 & 1 in 100 & 
• Rolling two 6s on two six-sided dice three times in a row\newline
• A mediocre bowler bowls a perfect game in a single game \\
\hline
\rowcolor{gray!10}
LL-5 & 1 in 10,000 & 1 in 1,000 & 
• A natural pregnancy resulting in triplets\newline
• Being dealt a straight flush in poker on the initial deal \\
\hline
LL-4 & 1 in 100,000 & 1 in 10,000 & 
• A random human is albino\newline
• Being dealt four of a kind in poker \\
\hline
\rowcolor{gray!10}
LL-3 & 1 in 1,000,000 & 1 in 100,000 & 
• Being dealt a royal flush in poker on the initial deal\newline
• Making a hole-in-one while golfing as an amateur in a single game \\
\hline
LL-2 & 1 in 10,000,000 & 1 in 1,000,000 & 
• A random human is struck by lightning in a given year\newline
• Flipping a coin and getting heads 20 times in a row \\
\hline
\rowcolor{gray!10}
LL-1 & 1 in 100,000,000 & 1 in 10,000,000 & 
• Earth being hit by a dinosaur-killing asteroid in a given year\newline
• Winning a major lottery jackpot on a single ticket \\
\hline
LL-0 & 1 in $\infty$ & 1 in 1,000,000,000,000 & 
• Provably impossible\newline
• Creating a perpetual motion machine \\
\hline
\end{NiceTabular}
\vspace{2em}
\label{tab:likelihood}
\end{table}

\section{Risk Levels Table}
\label{app: RL}
\normalsize
Table \ref{tab:interaction_matrix} defines mapping of Likelihood Levels (LL) and Harm Severity Levels (HSL) to Risk Levels (0-9) with odds lower limits for each LL.

\begin{table}[H]
\caption{Risk Levels Table.}
\vspace{5pt}
\label{tab:interaction_matrix}
\fontsize{7.5}{9}\selectfont
\renewcommand{\arraystretch}{1.2}
\setlength{\tabcolsep}{3pt}
\begin{NiceTabular}{|
>{\raggedright\arraybackslash}p{0.8cm}|
>{\raggedright\arraybackslash}p{2.7cm}|
>{\centering\arraybackslash}p{1.6cm}|
>{\centering\arraybackslash}p{1.4cm}|
>{\centering\arraybackslash}p{1.4cm}|
>{\centering\arraybackslash}p{1.4cm}|
>{\centering\arraybackslash}p{1.4cm}|
>{\centering\arraybackslash}p{1.6cm}|}[colortbl-like]
\hline
 & \textbf{Odds Lower} & \textbf{HSL-1} & \textbf{HSL-2} & \textbf{HSL-3} & \textbf{HSL-4} & \textbf{HSL-5} & \textbf{HSL-6} \\
 & \textbf{Limit} & \textit{Marginal but non Trivial} & \textit{Tragic} & \textit{Severe} & \textit{Devastating} & \textit{Extreme} & \textit{Globally Catastrophic} \\
\hline
\textbf{LL-8} & 1 in 10 & 4 & 5 & 7 & 8 & 9 & 9 \\
\hline
\rowcolor{gray!10}
\textbf{LL-7} & 1 in 100 & 4 & 5 & 6 & 7 & 8 & 9 \\
\hline
\textbf{LL-6} & 1 in 1,000 & 3 & 4 & 5 & 6 & 7 & 8 \\
\hline
\rowcolor{gray!10}
\textbf{LL-5} & 1 in 10,000 & 2 & 3 & 4 & 5 & 6 & 8 \\
\hline
\textbf{LL-4} & 1 in 100,000 & 1 & 2 & 3 & 4 & 6 & 7 \\
\hline
\rowcolor{gray!10}
\textbf{LL-3} & 1 in 1,000,000 & 0 & 1 & 2 & 4 & 5 & 7 \\
\hline
\textbf{LL-2} & 1 in 10,000,000 & 0 & 0 & 1 & 3 & 5 & 6 \\
\hline
\rowcolor{gray!10}
\textbf{LL-1} & 1 in 100,000,000 & 0 & 0 & 1 & 3 & 4 & 6 \\
\hline
\textbf{LL-0} & 1 in 1,000,000,000,000+ & 0 & 0 & 0 & 0 & 0 & 0 \\
\hline
\end{NiceTabular}
\label{tab:risk-levels}
\end{table}

\end{document}

\typeout{get arXiv to do 4 passes: Label(s) may have changed. Rerun}